\newif\ifarxiv 
\arxivtrue   

\ifarxiv
    \documentclass{fairmeta} 
    \usepackage{amsmath}
\else
    \documentclass{article}
    \usepackage{neurips_2025}              

\usepackage[utf8]{inputenc} 
\usepackage[T1]{fontenc}    
\usepackage{hyperref}       
\usepackage{url}            
\usepackage{booktabs}       
\usepackage{amsfonts}       
\usepackage{nicefrac}       
\usepackage{microtype}      
\usepackage[table,xcdraw]{xcolor}

\usepackage{siunitx}

\definecolor{RowGray}{HTML}{EEEEEE}

\hypersetup{
    colorlinks,
    linkcolor={red!80!black},
    citecolor={blue!80!black},
    urlcolor={blue!80!black}
}
    \usepackage{amsmath}
    \usepackage{cleveref}
\fi

\usepackage{algorithm}
\usepackage{algpseudocode}
\usepackage{diagbox}
\usepackage{amssymb}

\usepackage{rotating}
\usepackage{tikz}
\usepackage{tabularx}
\usepackage{enumitem}
\usepackage{xspace}
\usepackage{listings}
\usepackage{multirow}
\usepackage{array}
\usepackage{sidecap}
\usepackage{wrapfig}
\usepackage{graphicx}
\usepackage{subcaption}
\sidecaptionvpos{figure}{t}

\usepackage{graphicx}
\usepackage{siunitx}
\newcommand{\bxi}{\pmb{\xi}}

\newcommand{\bXi}{\pmb{\Xi}}

\newcommand{\bX}{{\bf X}}
\newcommand{\bx}{{\bf x}}
\newcommand{\bu}{{\bf u}}
\newcommand{\bz}{{\bf z}}
\newcommand{\bs}{{\bf s}}
\newcommand{\bQ}{{\bf Q}}
\newcommand{\bK}{{\bf K}}
\newcommand{\bV}{{\bf V}}
\newcommand{\R}{\mathbb{R}}
\newcommand{\softmax}{\text{SoftMax}}
\newcommand{\logsp}{\text{LogSumExp}}

\newcolumntype{H}{>{\setbox0=\hbox\bgroup}c<{\egroup}@{}}

\def\mypar#1{\vspace{2mm}\noindent\textbf{#1.}\hspace{1mm}}

\crefname{lstlisting}{listing}{listings} 
\Crefname{lstlisting}{Listing}{Listings}

\usepackage{pifont}
\newcommand{\cmark}{\ding{51}}%
\newcommand{\xmark}{\ding{55}}%

\makeatletter
\DeclareRobustCommand\onedot{\futurelet\@let@token\@onedot}
\def\@onedot{\ifx\@let@token.\else.\null\fi\xspace}

\def\eg{{e.g}\onedot} 
\def\ie{{i.e}\onedot} 
 
\def\etc{{etc}\onedot} 
\def\wrt{w.r.t\onedot}

\makeatother

\def\ours{Entropy Rectifying Guidance\xspace}
\def\OURS{ERG\xspace}

\title{Entropy Rectifying Guidance for Diffusion and Flow Models}

\ifarxiv

    \author[1,2]{Tariq Berrada Ifriqi}
    \author[1,3,4,5]{Adriana Romero-Soriano}
    \author[1]{Michal Drozdzal}
    \author[1]{Jakob Verbeek}
    \author[2]{Karteek Alahari}
    
    \affiliation[1]{FAIR at Meta}
    \affiliation[2]{Univ. Grenoble Alpes, Inria, CNRS, Grenoble INP, LJK, France}
    \affiliation[3]{McGill University}
    \affiliation[4]{Mila, Quebec AI institute}
    \affiliation[5]{Canada CIFAR AI chair}
    
    \abstract{Guidance techniques are commonly used in diffusion and flow models to improve image quality and input consistency for conditional generative tasks such as class-conditional and text-to-image generation.
In particular, classifier-free guidance (CFG) is the most widely adopted guidance technique. 
It results, however, in trade-offs across quality, diversity and consistency: improving some at the expense of others.
While recent work has shown that it is possible to disentangle these factors to some extent, such methods come with an overhead of requiring an additional (weaker) model, or require more forward passes per sampling step. 
In this paper, we propose \emph{\ours} (\OURS), a simple and effective guidance method based on inference-time changes in the attention mechanism of state-of-the-art diffusion transformer architectures, which allows for simultaneous improvements over image quality, diversity and prompt consistency. 
\OURS is more general than CFG and similar guidance techniques, as it extends to unconditional sampling. 
We show that \OURS results in significant improvements in various tasks, including  text-to-image, class-conditional and unconditional image generation.
We also show that \OURS can be seamlessly combined with other recent guidance methods such as CADS and APG, further improving  generation results.}
    
    \correspondence{First Author at \email{tariqberrada@meta.com}}

\else

    \author{%
      TBD
    }
\fi

\begin{document}

\maketitle

\ifarxiv
\else 
    \begin{abstract}
    Guidance techniques are commonly used in diffusion and flow models to improve image quality and input consistency for conditional generative tasks such as class-conditional and text-to-image generation.
In particular, classifier-free guidance (CFG) is the most widely adopted guidance technique. 
It results, however, in trade-offs across quality, diversity and consistency: improving some at the expense of others.
While recent work has shown that it is possible to disentangle these factors to some extent, such methods come with an overhead of requiring an additional (weaker) model, or require more forward passes per sampling step. 
In this paper, we propose \emph{\ours} (\OURS), a simple and effective guidance method based on inference-time changes in the attention mechanism of state-of-the-art diffusion transformer architectures, which allows for simultaneous improvements over image quality, diversity and prompt consistency. 
\OURS is more general than CFG and similar guidance techniques, as it extends to unconditional sampling. 
We show that \OURS results in significant improvements in various tasks, including  text-to-image, class-conditional and unconditional image generation.
We also show that \OURS can be seamlessly combined with other recent guidance methods such as CADS and APG, further improving  generation results.    
    \end{abstract}
\fi

\begin{figure*}[t]

\renewcommand{\arraystretch}{0.5}
\setlength\tabcolsep{1pt}
\setlength\fboxsep{0pt}

\def\myim#1{\fbox{\includegraphics[width=0.17\linewidth,height=0.17\linewidth]{figures/examples/showcase_0/img_#1.png}}}
\def\myimc#1{\fbox{\includegraphics[width=0.17\linewidth,height=0.17\linewidth]{figures/examples/showcase_0/astronaut/img_#1.png}}}
\def\myima#1{\fbox{\includegraphics[width=0.17\linewidth,height=0.17\linewidth]{figures/examples/panda/img_#1.png}}}
\def\myimb#1{\fbox{\includegraphics[width=0.17\linewidth,height=0.17\linewidth]{figures/examples/ballerina/img_#1.png}}}

\footnotesize
\resizebox{\textwidth}{!}{
\begin{tabular}{ccccccc}
 \begin{sideways} {\it CFG} \end{sideways}   
 & \myimc{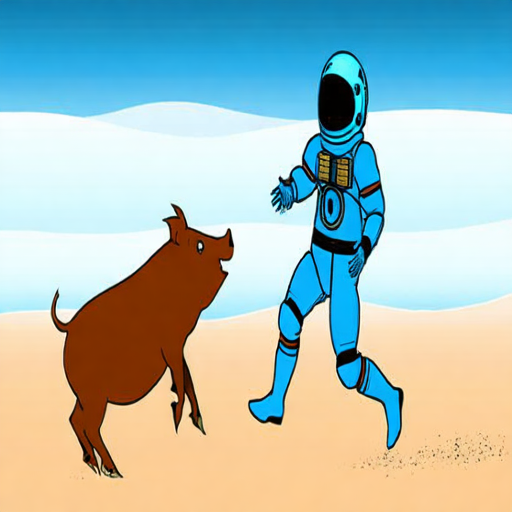}     
 & \myimc{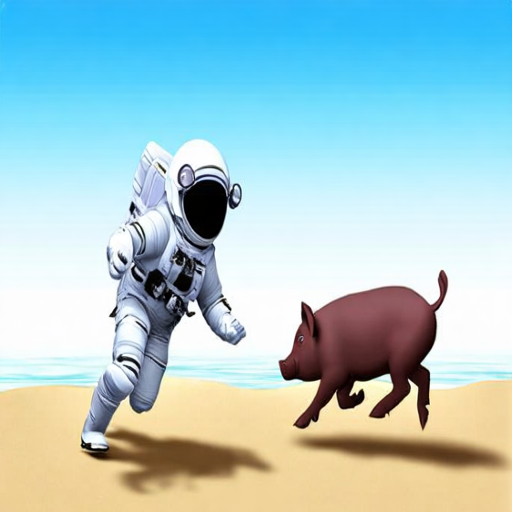}     & \myimc{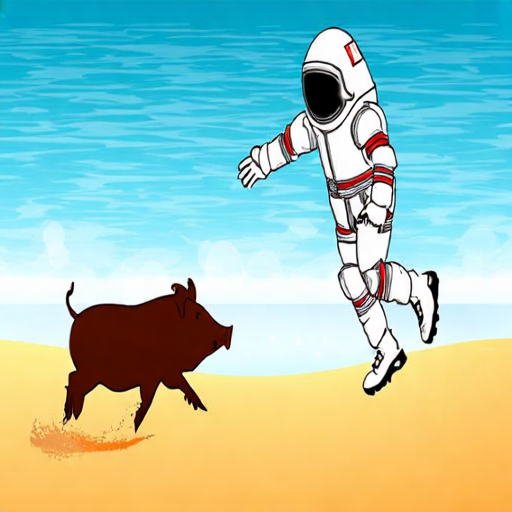}
 & \myim{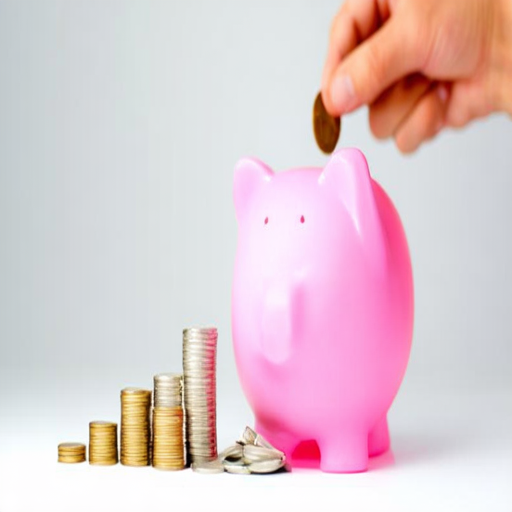}
 & \myim{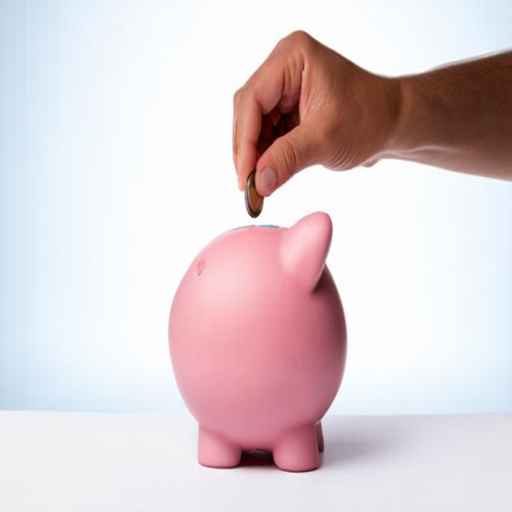}     & \myim{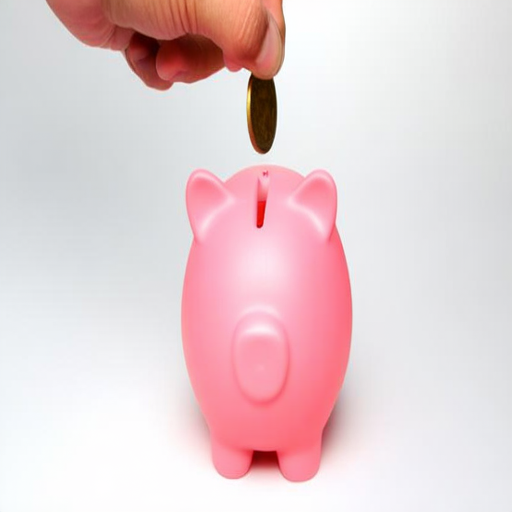}
 \\
\begin{sideways} {\it \OURS (ours)} \end{sideways}   
 & \myimc{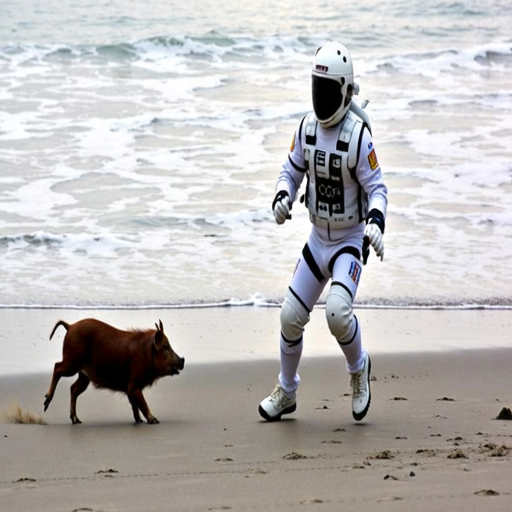}     
 & \myimc{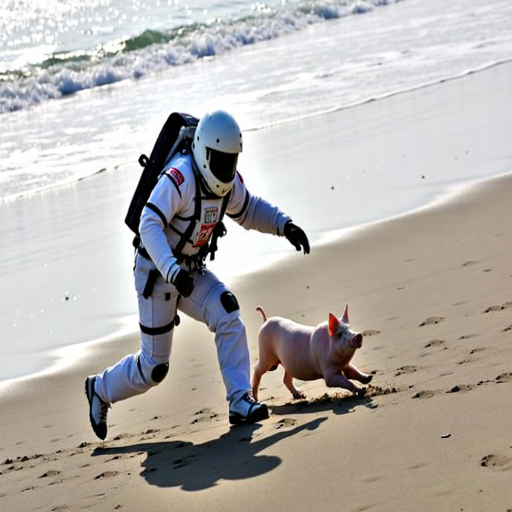}     & \myimc{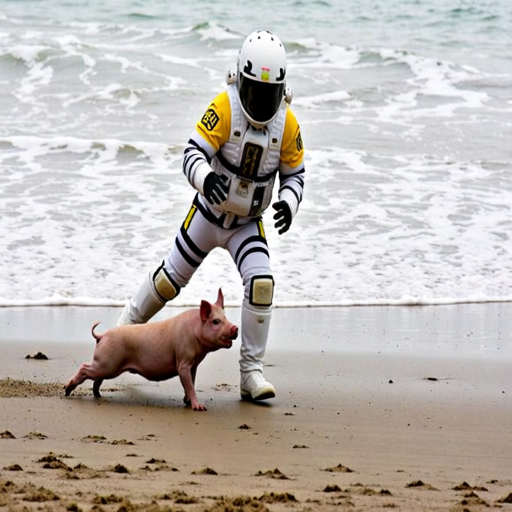}
 & \myim{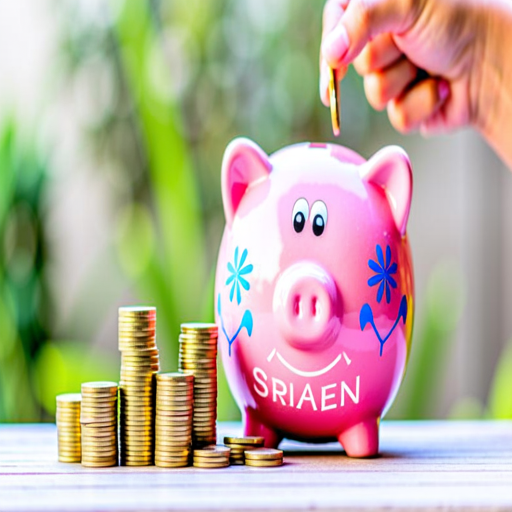}
 & \myim{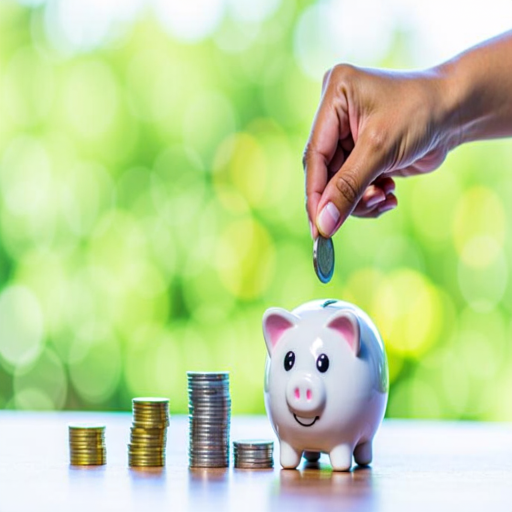}     & \myim{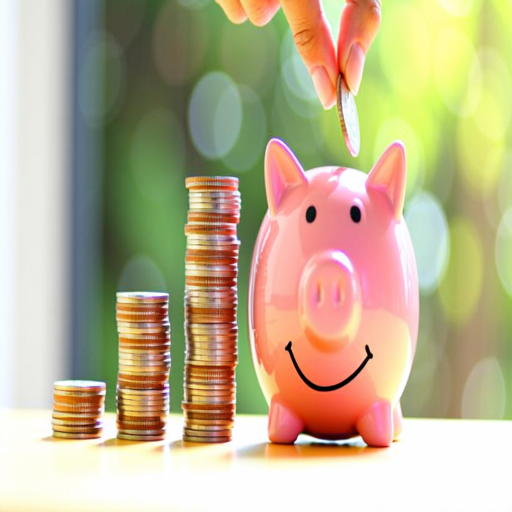}
 \\
 \begin{sideways} {\it CFG} \end{sideways}   
 & \myimb{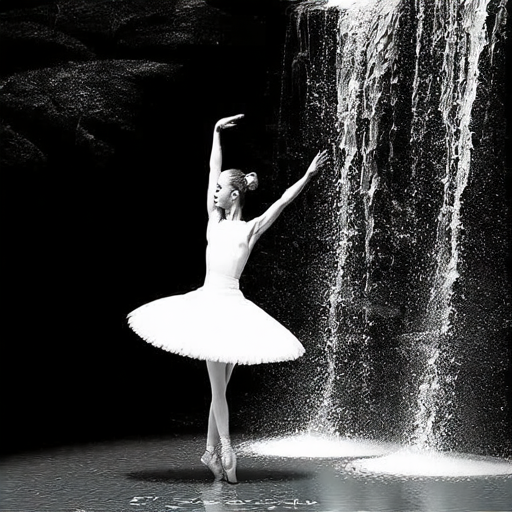}
 & \myimb{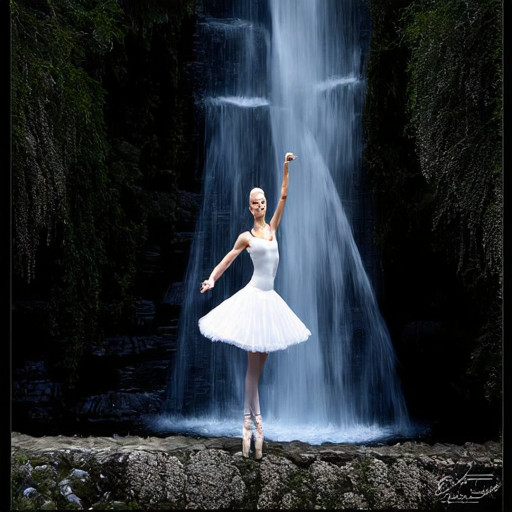}     & \myimb{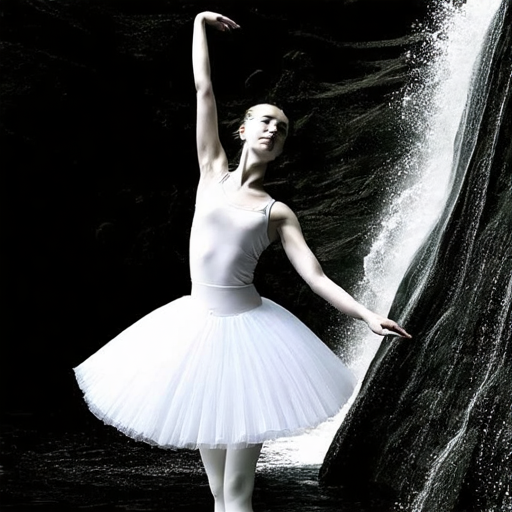} 
 & \myima{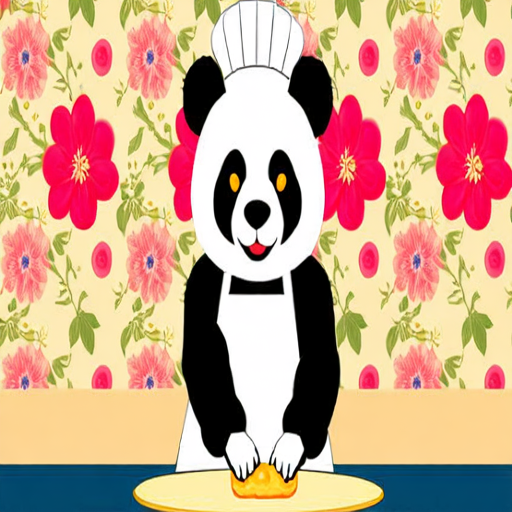}     
 & \myima{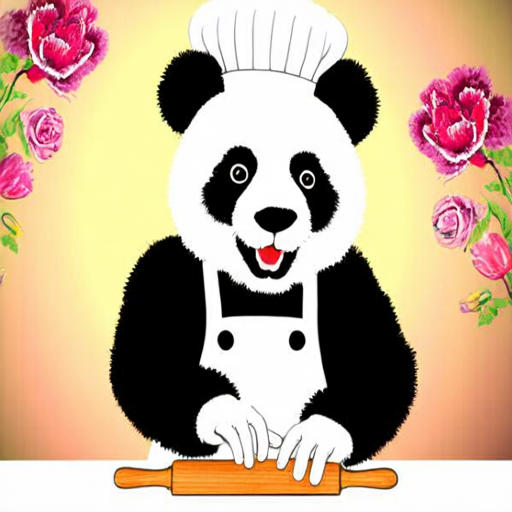}     
 & \myima{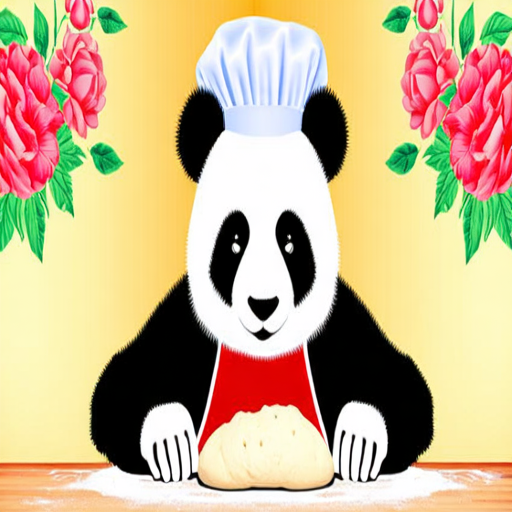}
 \\
\begin{sideways} {\it \OURS (ours)} \end{sideways}   
 & \myimb{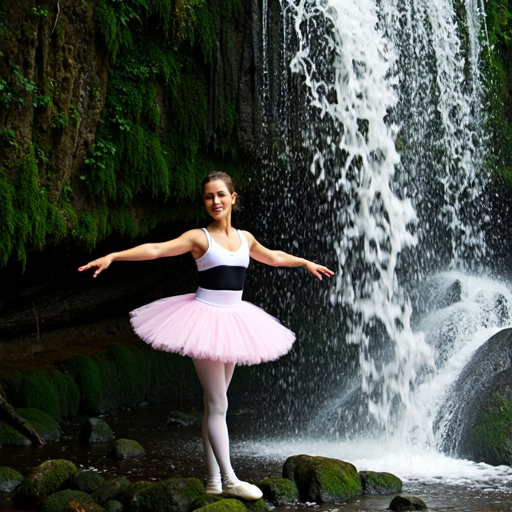}  
 & \myimb{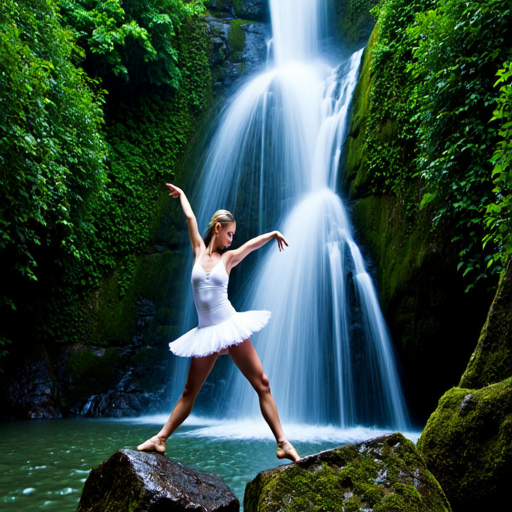}        & \myimb{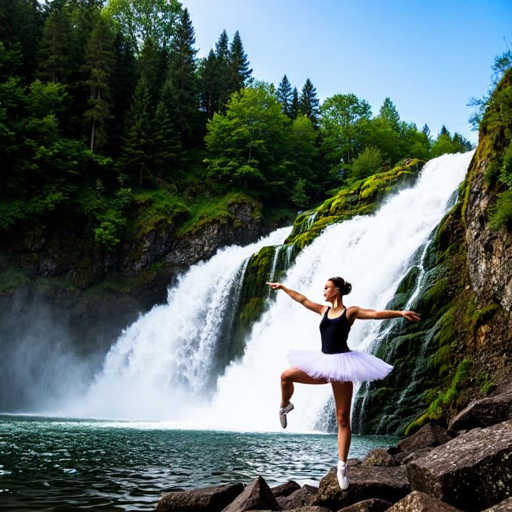} 
 & \myima{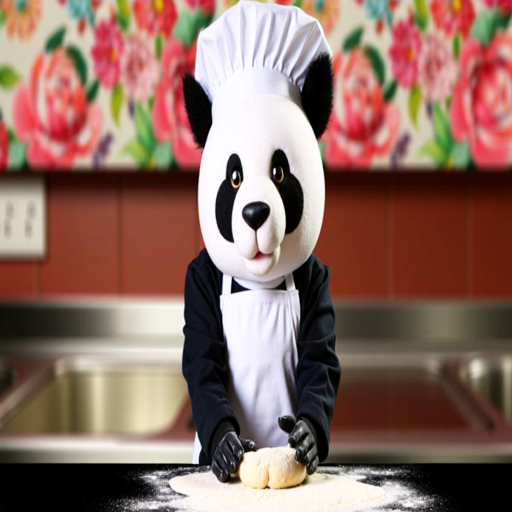}     
 & \myima{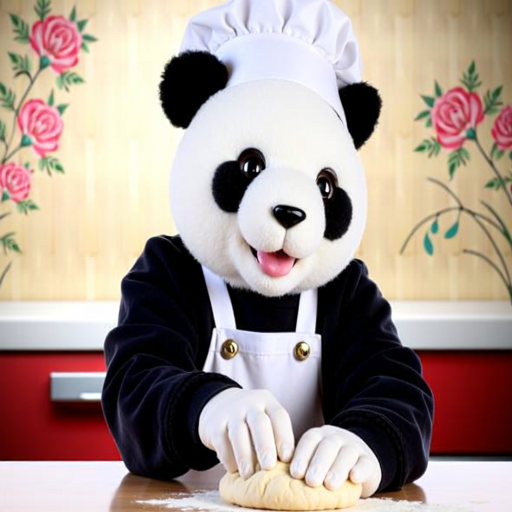}     & \myima{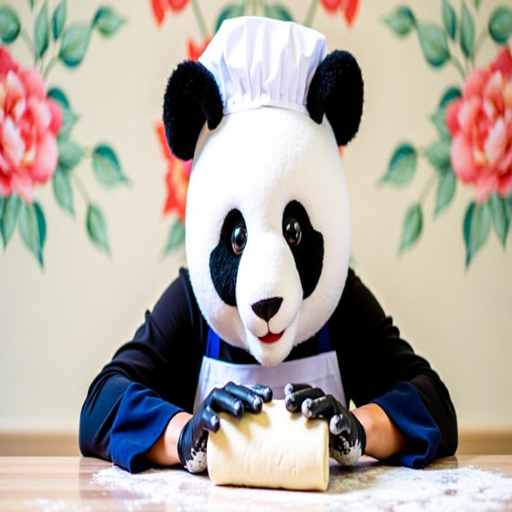}
 \\
\end{tabular}}
\caption{{\bf Qualitative comparison of classifier-free guidance (CFG) and our \ours (\OURS).}
The images generated using \OURS (bottom rows) exhibit greater quality and diversity than standard CFG. 
Images are generated using 50 Euler steps; each column corresponds to a different random seed for the generations.
}
\vspace{-1em}
\label{fig:showcase_method}
\end{figure*}

\section{Introduction}
\label{section:intro}

Diffusion~\citep{pmlr-v37-sohl-dickstein15,ho2020denoising, song2021scorebased, dhariwal2021diffusion} and flow models~\citep{lipman2023flow, ma2024sitexploringflowdiffusionbased, sd3} are state-of-the-art generative modeling tools for various modalities, ranging from images~\citep{ldm, podell2024sdxl, chen2023pixartalpha, sd3}, to audio~\citep{NEURIPS2023_e1b619a9,controllable-music}, and video~\citep{jin2024pyramidalflowmatchingefficient,polyak2024moviegencastmedia}. These models generate data by starting with a simple prior and iteratively removing noise -- a process called ``denoising''. 
See~\citet{kingma2023understanding,lipman2023flow} for details on diffusion and flow matching. 
These models can be conditioned on various inputs to control the generative process, \eg, in text-to-image  models. 
Guidance techniques such as classifier guidance~\citep{dhariwal2021diffusion} and classifier-free guidance~\citep{ho2021classifierfree} are commonly used to improve sample quality and consistency with input conditioning. 
In the sampling process, these techniques combine a conditional signal with an unconditional one, and control the strength of conditioning through a scaling parameter. 
Although such guidance techniques are a crucial component in achieving state-of-the-art results, they are also known to negatively affect the diversity of generated samples given a particular prompt~\citep{sadat2024cads, karras2024guiding,kynkaanniemi2024applying, saharia_deep_understanding}. Moreover, too high guidance scales may lead to overly saturated images, affecting the quality of generated images~\citep{saharia_deep_understanding}.
An extensive analysis on the trade-offs of image generation quality, diversity, and consistency was presented by~\citet{astolfi2024pareto}. 
To mitigate these quality-diversity-consistency trade-offs, more advanced guidance techniques have recently been proposed, see \eg  \citet{sadat2024apg,karras2024guiding}. 
Most guidance techniques , however, require spending part of the training cycles on unconditional generation for the guidance to work, even if unconditional generation is not a goal in itself, 
and  are not applicable to unconditional sampling~\citep{sadat2024apg,ho2021classifierfree}.  
Others rely on a second model with weaker performance than the main model, thereby increasing memory requirements~\citep{karras2024guiding}.

In our work, we build upon the work of~\cite{karras2024guiding} and~\citet{hong2024seg}, and propose  \ours (\OURS): a simple and effective method to obtain \emph{both} a \emph{strong} and a \emph{weak} predictive signal from a \emph{single model} that leverages attention layers, where the model may be conditional or unconditional. 
In particular, our method uses the Hopfield energy formulation of attention~\citep{hopfield_all_you_need,hong2024seg} and applies a temperature scaling to the softmax function of the attention layers in order to obtain the weak predictive signal. 
This scaling does not require any adaptations in model training, and may be applied to pre-trained denoising models, as well as their accompanying text encoders. 
Moreover, motivated by this energy interpretation of attention layers, we also consider iterative re-application of attention layers, and rescaling the residual attention update. 
Our experiments show that manipulating the attention layers in this manner results in simultaneous improvements in sample quality, diversity, and consistency, contrary to the trade-offs observed by~\citet{astolfi2024pareto} for classifier-free guidance. 

\noindent
In summary, our contributions are the following:
\begin{enumerate}[noitemsep,topsep=-\parskip] 
    \item We propose \ours (\OURS), a guidance mechanism based on modifying the energy landscape of the attention layers. 
    \item Since our guidance mechanism does not require  unconditional  inference, it is directly applicable to any attention-based diffusion or flow model, including unconditional, class-conditional, and text-to-image models. 
    \item Experimentally, we find that ERG significantly improves image quality \emph{and} diversity while retaining the same prompt consistency as the standard classifier-free guidance ($+30$ points in density, $+4$ points in VQAScore) on COCO at  512 resolution using our 1.9B text-to-image model.
\end{enumerate}
\section{Related work and background}

In this section, we briefly review background material on diffusion and flow models, and related work on guidance techniques to improve sampling as well as the Hopfield energy formulation of attention.

\subsection{Diffusion and flow matching models}
Diffusion models~\citep{pmlr-v37-sohl-dickstein15,ho2020denoising, song2021scorebased} form a flexible class of generative models whose underlying principle is to map samples $\pmb{\epsilon}$ from a trivial unit Gaussian prior $p_0=\mathcal{N}(0, {\bf I})$ to samples from a learned model $p_1$ of the data distribution.
The forward process is defined as:
$\bx_t = \alpha_t \bx_1 + \sigma_t \pmb{\epsilon}$  with $t\in [0,1]$,
where $\bx_1\sim p_1$, and  $\alpha_t$ is a decreasing function of ``time'' $t$ while $\sigma_t$ an increasing function of $t$.

Flow matching methods~\citep{lipman2023flow} assume that 
$\alpha_0=\sigma_1=1$ and $\alpha_1=\sigma_0=0$. 
Using these assumptions, during the reverse process $\bx_t$ interpolates between $\pmb{\epsilon}$ at $t=0$ and $\bx_1$ at $t=1$. In contrast, score-based diffusion models~\citep{ho2020denoising, dhariwal2021diffusion} set $\alpha_t$ and $\sigma_t$ implicitly through different formulations of stochastic differential equations (SDE) where $\mathcal{N}(0, {\bf I})$ is the equilibrium distribution.
Additionally, they consider $t \in [0,T]$ with $T$ large enough so that $\bx_T$ is approximately distributed as a unit Gaussian random variable.

\subsection{Guidance mechanisms}

\mypar{Classifier guidance} 
To enable high quality conditional generation,~\citet{dhariwal2021diffusion} proposed to guide the sampling process by leveraging gradients from pre-trained auxiliary classifier $p(c|\bx)$ in each  denoising step. They use the classifier to define the (scaled) joint score function as 
$
    \nabla_{\bx_t} \log p(\bx_t,c) =  
    \nabla_{\bx_t} \log p(\bx_t) + w  \nabla_{\bx_t} \log p(c|\bx_t),
$
where  $p(\bx_t)$ is an unconditional data model, and $w$ is a scalar parameter regulating the strength of the classifier guidance. While classifier guidance allows to improve input consistency and image quality (at the expense of diversity), it requires an auxiliary classification model that is robust to inputs $\bx_t$ with varying amounts of noise.

\mypar{Classifier-free guidance}
To avoid the need for an auxiliary noise-robust classifier, 
\citet{ho2021classifierfree} proposed classifier-free guidance (CFG). In this case, during the training process, two generative models are learned, one conditional $p(\bx|c)$ and one unconditional $p(\bx|\emptyset)$. In practice, the unconditional model is trained by dropping conditioning information with a small probability. The score function used for sampling is extrapolated towards the conditional prediction and away from the unconditional prediction

$
 \nabla_\bx^\textrm{CFG} \log p(\bx|c)=  w  \nabla_\bx \log p(\bx|c) +(1 - w)  \nabla_\bx \log p(\bx|\emptyset).  
\label{eq:cfg}
$
While CFG improves image quality and input consistency with respect to classifier guidance, it tends to come at the cost of a reduction in diversity~\citep{astolfi2024pareto,ho2021classifierfree,sadat2024cads}. Moreover, CFG often leads to generation artifacts, such as over-saturation as the guidance scale $w$ grows~\citep{sadat2024apg}.

\mypar{Advanced guidance techniques}
Several improved variants of classifier-free guidance have been proposed recently.
\citet{hong2023SAG} presented Self-Attention Guidance (SAG), a guidance mechanism based on feeding a modified intermediate sample $\bx_t$ when performing inference for unconditional prediction.
The modification consists in blurring $\bx_t$ in regions that are most attended to by the model's self-attention.
This method has been developed for the U-Net architecture, hence applying it to more recent diffusion transformer architectures requires a hyperparameter search to understand which attention layers should be used for this method.

Smoothed energy guidance (SEG)~\citep{hong2024seg} contrasts conditional prediction with a ``weaker'' conditional prediction obtained by altering the attention's softmax energy with a Gaussian kernel applied to queries. This method is developed for U-Net-style architectures and the softmax alteration applies to the self-attention layers in the middle block of the U-Net. 
In the conditional case, SEG uses a linear combination of the conditional, unconditional, and energy-smoothed unconditional prediction, whereas in our approach we only use the conditional and smoothed conditional term. Thus, SEG  requires an additional function evaluation with respect to \OURS for conditional inference. For unconditional inference, both approaches require only two function evaluations. 
In a similar spirit, \citet{ahn2024PAG}  propose a guidance method based on manipulating the attention mechanism, by replacing the attention matrix with an identity mapping inside the denoiser U-Net.

\citet{karras2024guiding} proposed AutoGuidance, a method that  uses a smaller/weaker version of the same conditional model for classifier-free guidance, resulting in better diversity and image quality.
Like ours, their approach can also be applied to unconditional sampling. 
However, their method requires access to an earlier checkpoint of the model  or, for best results, training a separate model with lower capacity, as well as accessing two models when  sampling,  which may increase the memory footprint. 

Rather than considering modifications of the unconditional model term, 
\citet{sadat2024cads} proposed the ``condition-annealed diffusion sampler'' (CADS) to increase the diversity of generations while maintaining sample quality. This is achieved by adding Gaussian noise to the conditioning tokens during inference, using a piecewise-linear decreasing schedule on the noise amplitude. 
\citet{sadat2024apg}  propose ``adaptive projected guidance'' (APG), a variant of CFG that resolves the over-saturation problem by emphasizing guidance orthogonal to the conditional prediction, rescaling the guidance term, and introducing a negative momentum term. 
The latter two approaches can be easily combined with our approach, and we consider such combinations in our experiments.

Lastly, several works \citep{kynkaanniemi2024applying,chung2024cfgmanifoldconstrainedclassifierfree,pavasovic25arxiv, wang2024analysis}
explore non-constant weight schedules for CFG. 
Such methods are complementary to our work, as our method only operates at the architecture level by rectifying the attention updates. 
Although they therefore could be combined with our approach, we defer this to future work. 

\subsection{Hopfield energy formulation of attention}

The Hopfield network~\citep{hopfield} is a dense associative memory model that aims to associate an input with its most similar pattern. More specifically, it constructs an energy function to model an energy landscape that contains basins of attraction around the desired patterns. 
Modern Hopfield energy networks~\citep{hopfield_all_you_need} introduce a new family of energy functions that improve the storage capacity of the model and make it compatible with continuous embeddings.
Specifically, the following energy functional matches a continuous $d$-dimensional state (query) pattern $\bxi \in \R^d$ with $N$ stored (key) patterns $\bX=(\bx_1,...,\bx_N)\in \R^{d \times N}$ as 
$
    \label{eq:hopfield_energy}
    E(\bxi; \bX) = \frac{1}{2} \bxi^\top \bxi - \logsp\left(\bX^\top \bxi,\beta\right),
$
where $\logsp(\bx,\beta)=\beta^{-1} \log \left( \sum_{i=1}^d \exp(x_i) \right)$, where $x_i$ are the elements of the vector $\bx$, and $\beta$ a scalar hyperparameter defining the sharpness of the approximation of the maximum in the $\logsp$ operation. Intuitively, the first term imposes a finite norm on the queries while the second term measures the alignment between the state patterns (queries) and stored patterns (keys).
Using the Concave-Convex Procedure (CCCP)~\citep{CCCP}, an iterative update rule, which converges to the global minima of the energy, can be derived as
$
    \label{eq:attn_en}
    \bxi_{l+1} = \bX \;\softmax(\beta \bX^\top \bxi_l),   
$
where $\softmax(\bx) = \exp(\bx-\logsp(\bx, 1))$. 
Equivalently, each iterative update step can be seen as a gradient update in the negative direction of the energy
$
    \label{eq:e_grad}
    \nabla_{\bxi} E(\bxi; \bX) = \bxi - \bX \;\softmax (\beta \bX^\top \bxi).
$
Taking a gradient descent step on this energy landscape with a step size  $\gamma$ results in an update of the form $\bxi_{l+1} = \bxi_l - \gamma \left( \bxi_l - \bX \softmax(\beta \bX^\top \bxi_l)\right)$, and for $\gamma=1$ we recover the CCCP update.

\citet{hopfield_all_you_need} show that the CCCP update is related to the standard attention operation as follows.
Assuming there are $S$  state (query) patterns, and $N$ stored (key) patterns that can be mapped to keys, queries and values using linear transformations, the state pattern can be obtained through a concatenation: $\bXi = [\bxi_1, \dots, \bxi_S] \in \R^{d\times S}$.
Then the attention map is given by
$\Xi_{l+1} = \bX \softmax(\bX^\top \Xi_l )$.
The keys, queries and values are obtained via linear projections as $\bK = \bX W_K^\top \in \mathbb{R}^{N \times d},\ \bV = \bX W_V^\top \in \mathbb{R}^{N \times d}$ and $\bQ = \bXi_{l+1} W_Q^\top \in \mathbb{R}^{S \times d}$, respectively.
By setting $\beta=\frac{1}{\sqrt{d}}$ and substituting this into the CCCP update rule, we obtain:
$
    \bQ_{l+1} = \softmax(\frac{\bQ \bK^\top}{\sqrt{d}}) \bK
    = \softmax \left( \frac{\bQ \bK^\top}{\sqrt{d}} \right) \bV \left( W_K W_V^{-1}  \right)^\top.
$
Hence, we observe the equation of attention up to a linear transformation, which provides an interpretation of the attention mechanism through the Hopfield energy lens.
Such a formulation has been used in the Energy Based Cross-Attention method (EBCA)~\citep{park2024energy} for adaptive context control in order to incorporate additional contexts into the generative process of a conditional diffusion model.

\section{Guidance via Entropy Rectification}

Similarly to previous approaches~\citep{ho2021classifierfree,hong2024seg,karras2024guiding}, our approach is based on contrasting the denoising estimate with a less powerful one.
In particular, using the Hopfield energy interpretation of attention in \Cref{eq:hopfield_energy}, we manipulate the energy landscape of the attention layer, by rectifying the entropy of the associations in the attention operation.
The modified attention layers lead to lower quality predictions, which are used as the contrasting term for guidance. 
This approach does  not require a second model and can be used for both conditional and unconditional sampling.
We refer to our approach as \ours (\OURS).

\section{Guidance via entropy rectification}

Similarly to previous approaches~\citep{ho2021classifierfree,hong2024seg,karras2024guiding}, our approach is based on contrasting the (conditional) denoising estimate with a less powerful one.
In particular, using the Hopfield energy interpretation of attention~\citep{hopfield_all_you_need}, see  \Cref{eq:hopfield_energy}, we manipulate the energy landscape of the attention layer, by rectifying the entropy of the associations in the attention operation.
The modified attention layers lead to lower quality predictions, which are used as the contrasting term for guidance. 
This approach does  not require a second model, and can be used for both conditional and unconditional sampling.
We refer to our approach as \ours (\OURS).

\subsection{Manipulating the energy landscape}

Our method manipulates the energy landscapes by introducing two new test-time hyperparameters, $\alpha$ and $\tau$ in the energy function:
\begin{equation} \label{eq:general_energy}
    E (\bxi; \bX) = \frac{1}{2} \bxi^\top \bxi -  \alpha \cdot \logsp\left(\bX^\top \bxi,  \tau \cdot \beta\right),
\end{equation}
\begin{wrapfigure}{r}{0.47\textwidth}
    \vspace{-1.6em}
    \begin{minipage}{0.47\textwidth}
        \begin{algorithm}[H] 
            {\scriptsize
            \caption{Multi-step \ours.}\label{alg:cap}
            \begin{algorithmic}
                \Require $K \in \mathbb{N}$ number of gradient update steps.
                \Require $\gamma > 0$ step size.
                \Require $\alpha \in \mathbb{R}$ State pattern matching weight.
                \Require $\tau > 0$ attention temperature.
                \Require $\bK, \bV \in \mathbb{R}^{d \times N}$ keys and values.
                \Require $\bQ \in \mathbb{R}^{d \times S}$ queries.
                \State $k \gets 0$
                \While{$k < K$}
                    \State $\bQ \gets \bQ - \gamma \left( \bQ - \alpha \cdot \text{softmax}\left(\tau \cdot \beta \bQ \bK^\top \right)\bV \right)$~\footnotemark
                    \State $k \gets k+1$
                \EndWhile
            \end{algorithmic}}
        \end{algorithm}
    \end{minipage}
\end{wrapfigure}
where $\beta = \frac{1}{\sqrt{d}}$ is the default temperature of the attention attention update.
The temperature rescaling parameter $\tau$ controls the sharpness of the softmax attention, 
and $\alpha$ the relative importance of the similarity between the state matching term compared to the norm of the state patterns.
Temperature rescaling with $\tau$ is similar to the Gaussian blurring of the attention maps introduced in SEG ~\citep{hong2024seg}, but 
allows for non-local smoothing of the attention maps.
Additionally, the view of the attention layer as a CCCP update of the energy function, allows for consideration of different methods to minimize the energy landscape.
For instance, taking $K$ gradient descent steps with  step size $\gamma$, as illustrated in \Cref{alg:cap}. 
Using different settings of the hyperparameters $\alpha, \gamma,$  $\tau$ and $K$ allows us to manipulate the  attention operation, and obtain noise estimates that deviate from the trained model. We expect these to be weaker estimates compared to those provided by the model, as it was trained with standard attention layers, \ie with $\alpha=\gamma=\tau=K=1$.
When applying this rectification mechanism to the denoiser model, we refer to the method as image-\OURS, or I-\OURS for short. 
In particular, we apply it to the negative/unconditional prediction part of the classifier-free guidance, and only on certain layers of the network that we will identify in our experiments.
Additionally, we impose a kickoff threshold $\kappa$ on the time steps in which guidance  is applied.

\subsection{Manipulating the energy of the text encoder}
Besides the image denoising model, we can also manipulate the energy landscape of attention-based  text encoders to obtain a weak version of the conditional embeddings. 
Let $c$ be the text prompt that is used as conditioning. The text tokens are obtained by feeding the prompt to the text encoder:
$\textrm{Enc}(c) \in \mathbb{R}^{d_t}$, such as  Llama~\citep{grattafiori2024llama3herdmodels} or T5~\citep{chung2022scalinginstructionfinetunedlanguagemodels}.
Text tokens are then fed to the denoiser model through cross-attention layers.

To obtain a contrasting signal for guidance, we manipulate the energy landscape in the self-attention layers of the text encoder following Algorithm~\ref{alg:cap}. 
More specifically, for every self-attention layer in the text encoder, we introduce a temperature hyperparameter in the  softmax function. 
This enables us to change the strength by which keys and queries are being matched, resulting in a modified prompt embedding at the output.
For the remainder of the manuscript, we refer to this method condition-\OURS, or C-\OURS for short.
For simplicity reasons, for the text encoder, we only consider changing the temperature but not the step size $\gamma$,  pattern matching weight $\alpha$, and number of  update steps $K$.

\subsection{Guidance update}
When combining the energy modulations in the text-encoder and denoising model, we obtain our \ours (\OURS) update:

\begin{equation}
    \Delta_\text{\OURS}(\bx, c, t; \Theta_\xi) = w \cdot D(\bx, \pmb{\phi}_c, t)  + (1-w) \cdot D^{\bxi}(\bx, \pmb{\phi}^\tau_c, t; \Theta_\xi),
\end{equation}
where $D$ is the learned denoiser model with parameters $\Theta$ that are omitted for simplicity, $D^{\bxi}$ is the denoiser model where the attention layers have been replaced by the modified version presented in \Cref{eq:general_energy}, and $\Theta_\xi=\{\alpha,\gamma,\tau,K\}$ the set of hyperparameters introduced by \OURS. 
We use $\pmb{\phi}_c$ to denote the prompt embedding produced by  the text encoder, while $\pmb{\phi}_c^\tau$ denotes the embedding  obtained with the  modified  attention layers.

Compared to  standard CFG, the main differences are that (i) we replace the 
unconditional text embeddings with conditional embeddings obtained with the  entropy-rectified attention mechanism (C-\OURS), and (ii)  the denoiser model  for the negative/unconditional predictions  also uses modified  attention layers  (I-\OURS).
Finally, the changes to the image denoiser are only applied after a certain point during sampling in order to not overly penalize the negative components of the \OURS update at the start of sampling. 
For the text-encoder we apply the temperature scaling throughout the denoising process, so that at any stage we obtain a noise prediction that can be contrasted with the vanilla denoising signal.

Note that our approach can be combined with other approaches, \eg, CADS~\citep{sadat2024cads} and  APG ~\citep{sadat2024apg}. 
We will explore such combinations in our experiments. 
\subsection{Manipulating the energy landscape}
Our method manipulates the energy landscapes by introducing two new test-time hyperparameters, $\alpha$ and $\tau$ in the energy function:
\begin{equation} \label{eq:general_energy}
    E (\bxi; \bX) = \frac{1}{2} \bxi^\top \bxi -  \alpha \cdot \text{logsumexp}\left(\bX^\top \bxi,  \tau \cdot \beta\right).
\end{equation}
\begin{wrapfigure}{r}{0.47\textwidth}
    \vspace{-1.6em}
    \begin{minipage}{0.47\textwidth}
        \begin{algorithm}[H] 
            {\scriptsize
            \caption{Multi-step \ours.}\label{alg:cap}
            \begin{algorithmic}
                \Require $K \in \mathbb{N}$ number of gradient update steps.
                \Require $\gamma > 0$ step size.
                \Require $\alpha \in \mathbb{R}$ State pattern matching weight.
                \Require $\tau > 0$ attention temperature.
                \Require $\bK, \bV \in \mathbb{R}^{d \times N}$ keys and values.
                \Require $\bQ \in \mathbb{R}^{d \times S}$ queries.
                \State $k \gets 0$
                \While{$k < K$}
                    \State $\bQ \gets \bQ - \gamma \left( \bQ - \alpha \cdot \text{softmax}\left(\tau \cdot \beta \bQ \bK^\top \right)\bV \right)$~\footnotemark
                    \State $k \gets k+1$
                \EndWhile
            \end{algorithmic}}
        \end{algorithm}
    \end{minipage}
\end{wrapfigure}
\footnotetext{$\beta = \frac{1}{\sqrt{d}}$ is the default temperature of the attention attention update.}
The temperature rescaling parameter $\tau$ controls the
sharpness of the softmax attention, 
and $\alpha$ the relative importance of the similarity between the state matching term compared to the norm of the state patterns.
Temperature rescaling with $\tau$ is similar to the Gaussian blurring of the attention maps introduced in SEG ~\citep{hong2024seg}, but 
allows for non-local smoothing of the attention layers.
Additionally, the view of the attention layer as a CCCP update of the energy function, allows for consideration of different methods to minimize the energy landscape.
For instance, taking $K$ gradient descent steps with  step size $\gamma$, as illustrated in \Cref{alg:cap}. 
Using different settings of the hyperparameters $\alpha, \gamma,$ and $\tau$ allows us to manipulate the energy landscape modeling the attention operation, and obtain noise estimates that deviate from the trained model. We expect these to be weaker estimates compared to those provided by the model, as it was trained with standard attention layers, \ie with $\alpha=\gamma=\tau=1$.
When applying this rectification mechanism to the denoiser model, we refer to the method as image-\OURS, or I-\OURS for short. 
In particular, we apply it to the negative/unconditional prediction part of the classifier-free guidance, only on certain layers of the network that will be defined subsequently.
Additionally, we impose a kickoff threshold $\kappa$ on the time steps in which guidance  is applied.

\subsection{Manipulating the energy of the text encoder}
Besides the image denoising model, we can also manipulate the energy landscape of attention-based  text encoders to obtain a weak version of the conditional embeddings. 
Let $c$ be the text prompt that is used as conditioning. The text tokens are obtained by feeding the prompt to the text encoder: 
$\textrm{Enc}(c) \in \mathbb{R}^{d_t}$, such as  Llama~\citep{grattafiori2024llama3herdmodels} or T5~\citep{chung2022scalinginstructionfinetunedlanguagemodels}.
Text tokens are then fed to the denoiser model through cross-attention layers.

To obtain a contrasting signal for guidance, we manipulate 
the energy landscape in the self-attention layers of the text encoder following Algorithm~\ref{alg:cap}. More specifically, for every self-attention layer in the text encoder, we introduce a temperature hyperparameter in the attention map of every softmax. This enables us to reduce the certainty with which keys and queries are being matched, resulting in less precise tokens at the output.
For the reminder of the manuscript, we refer to this method condition-\OURS, or C-\OURS for short.
For simplicity reasons, we only consider changing the temperature but not the step size $\gamma$ and pattern matching weight $\alpha$.

\subsection{Guidance update}
When combining the energy modulations in the text-encoder and denoising model, we obtain our 
\ours (\OURS) update:
\begin{equation}
    \Delta_\text{\OURS}(\bx, c, t; \Theta_\xi) = w \cdot D(\bx, \pmb{\phi}_c, t)  + (1-w) \cdot D^{\bxi}(\bx, \pmb{\phi}^\tau_c, t; \Theta_\xi),
\end{equation}
where $D$ is the learned denoiser model with parameters $\Theta$ that are omitted for simplicity, $D^{\bxi}$ is the denoiser model where the attention layers have been replaced by the modified version presented in \Cref{eq:general_energy}, and $\Theta_\xi$ the set of hyperparameters introduced by our method. 
$\pmb{\phi}_c$ are hidden states of the text encoder while $\pmb{\phi}_c^\tau$ is the hidden states of the text encoder obtained with a modified energy landscape in the attention layers.
Compared to the standard CFG, the main differences are that we replace the 
unconditional text embeddings with conditional embeddings obtained with the  entropy-rectified attention mechanism (C-\OURS). 
Similarly, the denoiser model used for the negative/unconditional predictions is also obtained with a rectified attention mechanism (I-\OURS).
Finally, the denoiser level updates are only applied after a certain point during sampling in order to not overly penalize the negative components of the \OURS update at the start of sampling.

Note that our approach can be combined with other approaches, \eg, CADS~\citep{sadat2024cads} and  APG ~\cite{sadat2024apg}. 
We will explore such combinations in our experiments. 

\section{Experimental evaluation}
 

\subsection{Experimental setup} \label{sec:exp_setup}

\mypar{Datasets and architectures}
We experiment with class-conditional and text-to-image models  trained  using  rectified flow-matching~\citep{lipman2023flow, sd3}.
We use a face-blurred version of ImageNet~\citep{deng09cvpr} to train class-conditional models at 256 and 512 resolution based on the  XL/2 variant of the DiT architecture~\citep{Peebles2022DiT}, which is composed of 28 attention blocks with hidden dimension of $1152$, resulting in $790$M parameters.
For the text-to-image model, we use an architecture similar to MMDiT~\citep{sd3}, and train a 512 resolution model on a mix of a proprietary dataset of 320M text-image pairs and YFCC100M~\citep{thomee2016yfcc}, where all faces in YFCC100M have been blurred.
Similar to MMDiT~\citep{sd3}, the model uses a mix of different text encoders: Llama3-8B~\citep{grattafiori2024llama3herdmodels} and Flan-T5-XL~\citep{chung2022scalinginstructionfinetunedlanguagemodels}.
During training, each of the text encoders is disabled with a probability of $\sqrt{0.1}$, so that the probability of both encoders being disabled is around $10\%$.
We enable both text encoders during inference time for text-to-image generation, and disable both text-encoders for unconditional image generation experiments.
The architecture of the model is made of $38$ blocks with a hidden dimension of $1,536$, resulting in approx $1.9$B parameters.
For both datasets, we recaption the images using both Florence-2 Large~\citep{xiao2023florence2advancingunifiedrepresentation} to obtain medium-length captions and PaliGemma-3B~\citep{beyer2024paligemmaversatile3bvlm} for shorter COCO-style captions.
Additional techniques to improve training efficiency  of this model, such as conditioning mechanisms and pre-training strategies, were adopted from \cite{berrada2024on}.
For the class-conditional model we use the asymmetric  autoencoder of \cite{zhu2023designing}, while for the larger text-to-image model we use the SD3 autoencoder~\citep{sd3}.

\mypar{Metrics}
We consider metrics for quality, diversity, and consistency~\citep{astolfi2024pareto}.
We measure \emph{sample quality} with FID~\citep{heusel17nips} and density~\citep{naeem2020reliable}; 
\emph{sample diversity} is measured with coverage~\citep{naeem2020reliable} and FID; 
and \emph{prompt consistency} is measured with 
CLIPScore~\citep{hessel21emnlp} and VQAScore~\citep{lin2024evaluatingtexttovisualgenerationimagetotext}.
For evaluation of text-to-image and unconditional generation, we use the 40k COCO'14 validation image-caption pairs. 
For the class-conditional models, we sample 50 images for each of the 1,000 ImageNet classes and use the ImageNet validation set as a reference. All evaluated models are sampled using the Euler method with $50$ sampling steps.
We use the EvalGIM~\citep{hall2024evalgimlibraryevaluatinggenerative} library for all evaluations.

\begin{table*}[t]
    \centering
    \caption{{\bf Comparison of \OURS with other guidance approaches for text-to-image generation.} 
    We compare \OURS to other state-of-the-art guidance approaches and mark the best result in each column in bold in the  top part of the table. In the bottom part of the table we evaluate combinations of  ERG with APG and CADS, and bold results when they surpass the results in the upper part of the table. 
    }
    {\scriptsize
    \begin{tabular}{lcHHccccc}
        \toprule
        \diagbox[innerwidth=2cm]{Guidance}{Metric} & FID $(\downarrow)$ & Precision $(\uparrow)$ & Recall $(\uparrow)$ & Density $(\uparrow)$ & Coverage $(\uparrow)$ & CLIPScore $(\uparrow)$ & VQAScore $(\uparrow)$ & NFE $(\downarrow)$\\
        \midrule
        CFG~\citep{ho2021classifierfree} & $12.81$ & $65.71$ & $43.95$ & $98.24$ & $71.12$ & $26.45$ & $70.15$ & $2$\\
        APG~\citep{sadat2024apg} & ${11.88}$ & $66.34$ & $45.15$ & $104.07$ & $73.06$ & $26.54$ & $72.47$& $2$\\
        CADS~\citep{sadat2024cads} & $11.93$ & $ 66.42$ & $45.75$ & $101.01$ & $72.99$ & $26.76$ & $73.36$ & $2$\\
        PAG\textsuperscript{*}~\citep{ahn2024PAG} & $12.75$ & $ 66.42$ & $43.94$ & $107.21$ & $72.20$ & $26.80$ & $73.32$ & $2$\\
        SAG\textsuperscript{*}~\cite{hong2023SAG} & ${\bf 11.68}$ & $64.97$ & ${\bf 47.98}$ & $103.58$ & $72.74$ & $26.81$ & $72.16$ & $2$ \\
        SEG\textsuperscript{*}~\citep{hong2024seg} & $16.87$ & $61.22$ & $36.88$ & $87.77$ & $61.91$ & ${\bf 26.86}$ & $73.59$& $3$\\
        AutoGuidance~\citep{karras2024guiding} & $16.62$ & $61.48$ & $34.60$ & $87.02$ & $62.75$ & $26.59$ & $73.53$ & $2$\\
        \OURS (ours) & $13.62$ & $\bf 70.92$ & $41.43$ & $\bf 120.25$ & $\bf 73.21$ & ${\bf 26.86}$ & ${\bf 73.96}$ & $2$\\
        \midrule
        \OURS (ours) + APG & ${\bf 11.37}$ & $69.50$ & ${\bf 50.25}$ & $115.08$ & ${\bf 80.50}$ & $26.74$ & $73.55$ & $2$\\
        \OURS (ours) + CADS & $12.87$ & ${\bf 72.72}$ & $38.89$ & ${\bf 128.54}$ & $76.23$ & $26.75$ & $73.45$ & $2$\\
        \bottomrule
    \end{tabular}}
    \label{tab:main_comp}
\end{table*}

\mypar{Baselines}
In addition to the standard classifier-free guidance, we compare our method to several recent state-of-the-art guidance techniques: Condition-Annealed Diffusion Sampler (CADS)~\citep{sadat2024cads}, Adaptive Projected Guidance (APG)~\citep{sadat2024apg}, Smooth Energy Guidance (SEG)~\citep{hong2024seg}, and Auto-Guidance~\citep{karras2024guiding}.
For APG, we follow the recommendations from the paper and set $\gamma_\text{APG}=-0.5$, $\eta_\text{APG}=0.0$, $r_\text{APG}=5.0$.
For CADS, we perform a grid search over $\tau_1^\text{CADS} \in [0.6, 0.8], \tau_2^\text{CADS} \in [0.8, 1.0], s^\text{CADS} \in [0.25, 1.0], \psi^\text{CADS}=1.0$.
Since SAG, PAG and SEG were developed specifically for the U-Net architecture, 
we adapt these method for diffusion transformers~\citep{Peebles2022DiT} by applying the method in the attention layers of the middle blocks of the transformer; we refer to these methods with an asterisk superscript.
Additional details on the choice of layers are provided in \Cref{sec:att_blocks}.
For AutoGuidance, we follow the recommendations from the paper and use an earlier checkpoint of the same model, at approximately $1/16$-th of the training, as the weaker model.
To ensure a fair comparison, we select the best performing guidance strength for baselines as well as our method using the rank-scoring algorithm detailed in  \Cref{sec:rankscoring}. 
Note that this is different from reporting the optimal score achieved for each metric, which might not correspond to any particular run because of inherent trade-offs between the different facets of the generations.


\subsection{Main experimental results} \label{sec:exp_results}
Throughout our experiments, we modify attention layers in both the text encoder (C-\OURS) and the image denoising model (I-\OURS), unless specified otherwise.

\begin{figure*}
    \centering
    \includegraphics[width=\linewidth]{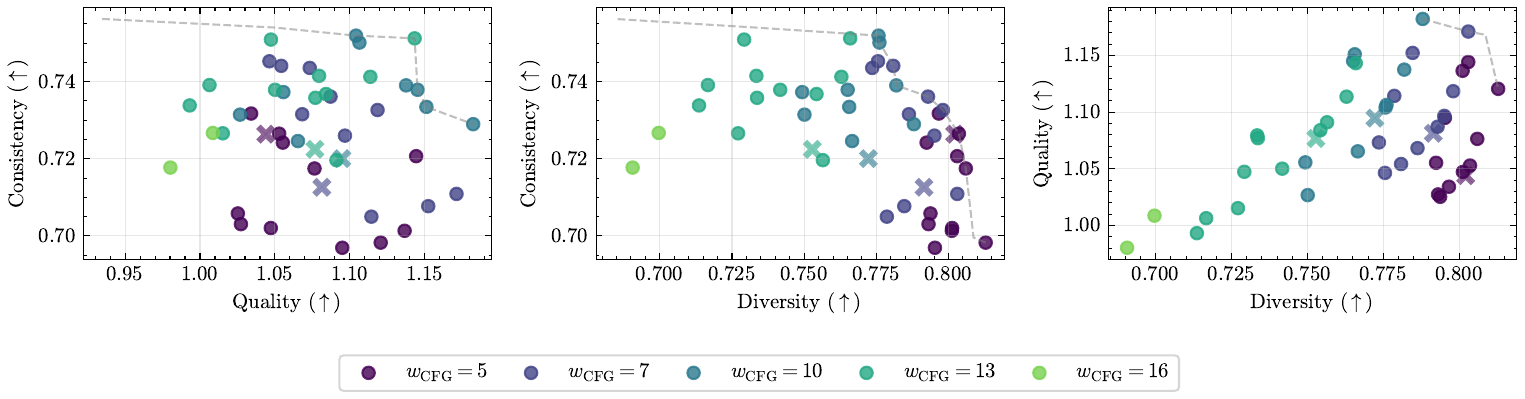}
    \caption{
    {\bf Pareto fronts on consistency-diversity-quality for text-to-image generation.} Comparing \OURS + APG (dots) with APG  
   (crosses).
    We sweep over  different guidance scales (each marked with a different color), and hyper-parameters $\alpha, \gamma, \tau$ for \OURS.
    Dashed lines trace the Pareto fronts for each plot.
    We measure consistency with VQAScore, quality with density and diversity with coverage.
    }
    \label{fig:coco_apg_pareto}
    \vspace{-1em}
\end{figure*}

\mypar{Text-to-image generation}
In \Cref{tab:main_comp}, we compare  \OURS with recent state-of-the-art guidance mechanisms  on the text-to-image generation task.
 \OURS  demonstrates excellent performance, outperforming baselines such as CFG, SAG, PAG and SEG\textsuperscript{*} in most metrics. Specifically, \OURS considerably boosts image quality as reported for Density (\eg, +22 points when compared to CFG). 
Additionally, \OURS achieves the highest consistency scores: $+0.4$ points in CLIPScore and $+3.8$ in VQAScore when compared to standard CFG.
Moreover, \OURS combined with APG achieves the overall best diversity, measured by FID and Coverage, and also improves  Density over prior methods. 
These results suggest that our \OURS approach is successfully able to boost the generation quality of the model in all three facets of generation (quality, diversity, and consistency).
Moreover, the number of function evaluations (NFE) required for 
our approach is only two per inference step, which is comparable to other methods except SEG which requires three. 

Following \citet{astolfi2024pareto}, we plot Pareto fronts for quality (measured by Density), diversity (Coverage), and consistency (VQAScore) metrics in \Cref{fig:coco_apg_pareto} when using \OURS + APG. 
We find all the points belonging to the Pareto fronts  correspond to \OURS + APG, which improves in all three facets of the generations \wrt APG, and provides significant boosts for quality and consistency. 
This can also be observed in the qualitative comparison between APG and \OURS + APG in \Cref{fig:methods_comp} and \Cref{fig:showcase_method_apg} in the supplementary material.
Similarly, in  \Cref{fig:showcase_method} the images sampled using \OURS show better visual quality than those sampled with CFG. 

\begin{figure*}
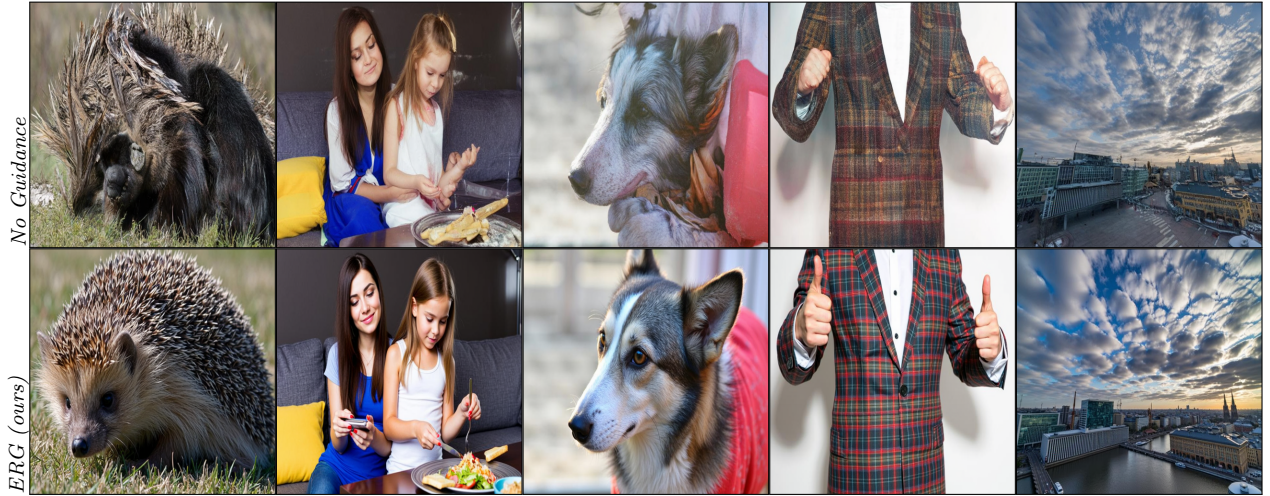

    \renewcommand{\arraystretch}{0.}
    \setlength\tabcolsep{0pt}
    \setlength\fboxsep{0pt}
\def\myim#1#2{\fbox{\includegraphics[width=0.195\linewidth,height=0.195\linewidth]{figures/uncond/#1/#2.png}}}
    \footnotesize
    \begin{tabular}{ccccccc}
     \centering \rotatebox{90}{\it No Guidance}   
     & \myim{baseline}{000020}     
     & \myim{baseline}{000021}     & \myim{baseline}{000043}
     & \myim{baseline}{000049}     & \myim{baseline}{000114}
     \\
     \begin{sideways} {\it \OURS (ours)} \end{sideways}   
     & \myim{tau_i}{000020}     
     & \myim{tau_i}{000021}     & \myim{tau_i}{000043}
     & \myim{tau_i}{000049}     & \myim{tau_i}{000114}
    \end{tabular}        
    \caption{{\bf Unconditional generation results.} 
    Compared to not using  guidance (top), our \OURS generates more realistic and detailed images and more coherent structure (bottom). 
    Images obtained from T2I model at $512$ with empty prompt as input. 
    Samples in each column use the same seed. 
    }
    \label{fig:qual_unc}
    \vspace{-1em}
\end{figure*}


\begin{table}
\centering
    \begin{minipage}[t]{0.49\textwidth}
            \centering
            \caption{{\bf Unconditional generation.} Comparing  \OURS to other approaches compatible with unconditional sampling.
            }
            \vspace{2.2em}
            \resizebox{\textwidth}{!}{
            \begin{tabular}{lcHHcc}
                \toprule
                 & FID $(\downarrow)$ & Precision $(\uparrow)$ & Recall $(\uparrow)$ & Density $(\uparrow)$ & Coverage $(\uparrow)$\\
                \midrule
                No guidance & $101.50$ & $13.50$ & $38.22$ & $8.99$ & $3.63$\\
                SAG~\citep{hong2023SAG} & $39.25$ & $43.45$ & $55.00$ & $46.22$ & $30.70$ \\
                PAG\textsuperscript{*}~\citep{ahn2024PAG} & $41.50$ & $42.91$ & ${\bf 62.86}$ & $45.65$ & $30.51$\\
                SEG\textsuperscript{*}~\citep{hong2024seg} & $37.75$ & $48.99$ & $50.10$ & $55.56$ & $34.79$\\
                AutoGuidance~\citep{karras2024guiding} & $39.50$ & $49.97$ & $42.09$ & $48.26$ & $34.71$\\
                \OURS (ours) & ${\bf 36.25}$ & ${\bf 68.87}$ & $57.15$ & ${\bf 55.84}$ & ${\bf 51.59}$\\
                \bottomrule
            \end{tabular}}
            \label{tab:energy_uncond}
    \end{minipage}
    \hfill
    \begin{minipage}[t]{0.49\textwidth}
            \centering
            \caption{\linespread{1.} {\bf Class-conditional generation.} Comparison of \OURS with other guidance methods for models trained for 256 and 512 resolution.
            }
           \resizebox{\textwidth}{!}{
            \footnotesize
            \begin{tabular}{llcHHcc}
                \toprule
                 & Res. & FID $(\downarrow)$ & Precision $(\uparrow)$ & Recall $(\uparrow)$ & Density $(\uparrow)$ & Coverage $(\uparrow)$\\
                \midrule
                 CFG~\citep{ho2021classifierfree} & \multirow{5}{*}{$256$} & ${\bf 3.67}$ & $76.63$ & $55.70$ & $127.03$ & $85.81$\\
                 PAG\textsuperscript{*}~\citep{ahn2024PAG} & & $5.31$ & $72.38$ & $52.55$ & $111.94$ & $81.04$\\
                 SAG~\citep{hong2023SAG} & & $3.78$ & $75.56$ & $\bf 58.68$ & $131.89$ & $86.35$\\
                 SEG\textsuperscript{*}~\citep{hong2024seg} & & $6.15$ & $77.97$ & $48.10$ & $132.22$ & $84.51$\\ 
                 \OURS (ours) & & ${\bf 3.67}$ & ${\bf 80.51}$ & $54.03$ & ${\bf 141.96}$ & ${\bf 86.72}$\\
                 \midrule
                 CFG~\citep{ho2021classifierfree} & \multirow{5}{*}{$512$} & $5.65$ & $81.39$ & $49.06$ & $146.97$ & ${\bf 86.70}$\\
                 PAG\textsuperscript{*}~\citep{ahn2024PAG} & & $4.65$ & $78.04$ & $52.44$ & $134.49$ & $86.50$\\
                 SAG~\citep{hong2023SAG} & & $4.81$ & $74.90$ & $52.84$ & $120.09$ & $83.91$ \\
                 SEG\textsuperscript{*}~\citep{hong2024seg} & & $6.59$ & $82.88$ & $33.25$ & $160.11$ & $81.85$\\ 
                 \OURS (ours) & & ${\bf 4.56}$ & ${\bf 84.39}$  & ${\bf 52.92}$ & ${\bf 163.63}$ & $86.13$\\
                \bottomrule
            \end{tabular}}
            \label{tab:energy_in1k}
    \end{minipage}
\end{table}

\mypar{Unconditional generation}
In the unconditional model sampling experiment, we compare I-\OURS with sampling without guidance and using applicable methods: AutoGuidance, SEG\textsuperscript{*}, PAG\textsuperscript{*} and SAG\textsuperscript{*}.
We provide quantitative evaluation results in  \Cref{tab:energy_uncond}, where we find that \OURS outperforms all other methods, 
with significant boosts in FID, Density, and Coverage over other methods.
The qualitative examples in  \Cref{fig:qual_unc} clearly exhibit artifacts in terms of structural coherence in all the objects present in the generated images when not using guidance, which disappear when using \OURS.

\mypar{Class-conditional generation}
For class-conditional generation, we observe similar trends as those seen for  text-to-image and unconditional sampling in \Cref{tab:energy_in1k}. 
In particular,  at 256 resolution, we find  improvements in Density and Coverage.
FID remains similar to CFG but is better compared to all other methods tested.
At 512 resolution, \OURS is best across all metrics, except for coverage where \OURS is slightly behind CFG and SEG\textsuperscript{*}. 

\begin{figure*}
    \centering
        \includegraphics[width=\textwidth]{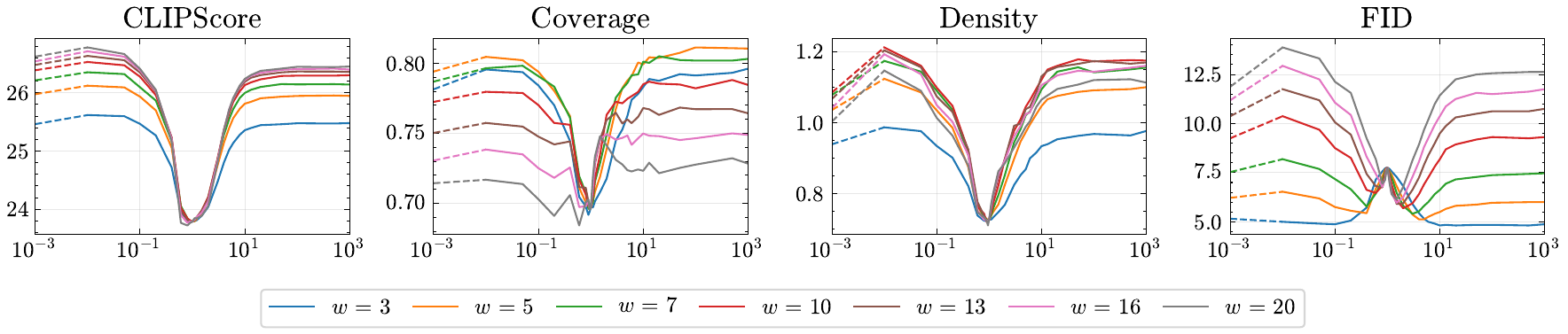}
        \caption{
        \linespread{1.}
        {\bf Temperature rescaling in the conditioning  for text-to-image generation (C-\OURS).} 
        We vary the text encoder's attention temperatures $\tau_c$. 
        Each curve corresponds to a different guidance strength $w$. 
        The left-most point on each curve represents the result for standard CFG.
        }
        \label{fig:coco_apg_tau_c}
        \vspace{-1em}
\end{figure*}

\subsection{Analysis and ablations} \label{sec:ablations}

\mypar{Text attention energy}
To isolate the effect of the temperature re-scaling in the text-encoder and image denoising components, we  experiment with C-\OURS and disable temperature rescaling in the image denoising model.
In \Cref{fig:coco_apg_tau_c}, we vary the SoftMax temperature $\tau_c$ used in the text encoder, and consider generation performance for different guidance strengths $\omega$.
For all metrics we find a somewhat symmetrical behavior around $\tau_c=1$, which corresponds to an unguided prediction because in this case there is no difference with the normal conditional prediction. 

The CLIPScore, Coverage and Density metrics generally improve when moving away from  $\tau_c=1$, making it either larger or smaller. 
For FID, the best values are obtained with low guidance scales and intermediate temperature values, the general trend shows improved FID using C-\OURS compared to standard guidance.
Compared with standard classifier-free guidance (left-hand side of the dashed lines), we find that all metrics can be improved for any guidance scale, provided with the right temperature.
While $\tau_c < 1$ results in higher CLIPScore, $\tau_c > 1$ results in higher Coverage, indicating that tuning $\tau_c$ provides an easy way to control the diversity-consistency tradeoff.

\mypar{Combining the different parts}
In \Cref{tab:energy_components}, we combine different parts of \OURS and measure the effects on different facets of image generation.
Our results show that all components show positive effects across all metrics, at the expense of a slight degradation in FID.
Compared to the CFG baseline (first row), most of the improvement in CLIPScore and VQA are brought by the conditional entropy rectification ($\tau_c$, second row),  while the  improvements in Coverage and Density mostly come from rectifying the attention in the denoiser to obtain ($\tau_i$, third row). 
Finally, a further modest improvement in Density, Coverage and VQA is brought by the update step size ($\gamma$, fourth row).

\mypar{Multi-step gradient descent}
In  \Cref{tab:steps_lr} we consider the effect of varying number of updates $K$ for each attention operation along with the update step size.
We find small variations in metrics with respect to the baseline $K= \gamma=1$ with slight improvements when setting $\gamma=1.5$, using multiple gradient descent steps did not induce significant gains.
Therefore, we used $K=  \gamma=1$ in our default setup in our experiments, unless specified otherwise.

\begin{table}
\centering

    \begin{minipage}[]{0.58\textwidth}
            \centering
            \caption{{\bf Impact of the different components of \OURS.} We accumulate different components of \OURS, namely C-\OURS, then I-\OURS through denoiser entropy rectification and objective reweighting, finally all are merged into \OURS. 
            }
            \label{tab:energy_components}
            \resizebox{\linewidth}{!}{

            \begin{tabular}{ccc|cccccc}
                \toprule
                $\tau_c$ & $\tau_i$ & $\gamma$ & FID $(\downarrow) $ & Density $(\uparrow)$& Coverage $(\uparrow)$ & CLIP $(\uparrow)$ & VQA $(\uparrow)$\\
                \midrule
                \xmark & \xmark & \xmark & {\bf 12.81} & 98.24 & 71.12 & 26.45 & 70.15\\
                \cmark & \xmark & \xmark & 13.06 & 109.52 & 72.06 & 26.73 & 73.10\\
\rowcolor[HTML]{EEEEEE}\cmark & \cmark & \xmark & 13.62 & 120.25 & 73.21 & {\bf 26.86} & 73.96\\
\cmark & \cmark & \cmark & 13.62 & {\bf 123.65} & {\bf 74.07} & 26.81 & {\bf 74.67}\\
                \bottomrule
            \end{tabular}
            }

    \end{minipage}
    \hfill
        \begin{minipage}[]{0.4\textwidth}
            \centering
            \caption{{\bf Multi-step gradient descent for optimizing the energy landscape.}
                we ablate different values for $K$ and $\gamma$.
            }
            \resizebox{\linewidth}{!}{
            \newcolumntype{s}{>{\columncolor[HTML]{EEEEEE}} c}

            \begin{tabular}{lsccc}
                \toprule
                $K$ & $1$ & $1$ & $1$ & $10$\\
                $\gamma$ & $1$ & $1.5$ & $0.5$ & $0.1$\\
                \midrule
                FID $(\downarrow)$ & $13.62$ & $13.62$ & $13.06$& ${\bf 12.68}$\\
                Density $(\uparrow)$ & $120.25$ & ${\bf 123.65}$ & $121.95$ & $119.07$\\
                Coverage $(\uparrow)$ & $73.21$ & ${\bf 74.07}$ & $72.63$ & $72.11$\\
                CLIP $(\uparrow)$ & ${\bf 26.86}$ & $26.81$ & $26.16$ & $26.77$\\
                VQA $(\uparrow)$ & $73.96$ & ${\bf 74.67}$ & $73.01$ & $73.95$\\
                \bottomrule
            \end{tabular}}
            \label{tab:steps_lr}
    \end{minipage}
    
\end{table}

\section{Conclusion}
We presented {\it \ours (\OURS)}, a novel guidance mechanism for sampling diffusion and flow models  which significantly improves sample quality without sacrificing diversity and consistency performance.
In particular, by manipulating the energy landscape of the attention layers in the diffusion transformer and the text encoder at inference time, \OURS significantly boosts the performance of  different models when studying their quality-consistency-diversity trade-offs and is applicable to different modalities such as text-to-image, class conditional and unconditional models.
Furthermore, \OURS outperforms recent state-of-the-art guidance methods such as CADS, APG, SEG and AutoGuidance, while  requiring the same compute as standard CFG.
\OURS can be  combined with approaches such as APG and CADS, which   further improves results.

\mypar{Acknowledgements} Karteek Alahari was supported in part by the Institute of Information \& Communications Technology Planning \& Evaluation (IITP) grant funded by the Korean Government (MSIT) (No.\ RS-2024-00457882, National AI Research Lab Project).

\clearpage\newpage
\ifarxiv
    \bibliographystyle{assets/plainnat}
\else
    \small
    \bibliographystyle{ieeenat_fullname}
\fi
\bibliography{paper}

\clearpage \newpage
\appendix

\section{Definitions and Background}
\subsection{Preliminaries}

\mypar{Background on diffusion}
Consider a data point $\bx$ drawn from the distribution $p_{data}(\bx)$, and let $t$ be a time step within the interval $[0, 1]$. The forward process is defined as $\bx_t = \bx + \sigma(t)\pmb{\epsilon}$, where Gaussian noise $\pmb{\epsilon} \sim N(0, \pmb{I})$ is added to the data. The function $\sigma(t)$ is monotonically increasing, with boundary conditions $\sigma(0) = 0$ and $\sigma(1) = \sigma_\text{max}$, where $\sigma_\text{max}$ is significantly larger than $\sigma_\text{data}$.
\citet{elucidating} demonstrated that the progression of the noisy samples $\bx_t$ can be captured by an ordinary differential equation (ODE):
\begin{equation} 
d\bx = -\dot{\sigma}(t)\sigma(t) \nabla_{\bx_t} \log p_t(\bx_t) dt.
\end{equation}
Alternatively, this can be expressed as a stochastic differential equation (SDE):
\begin{equation}
    \begin{split}
    d\bx = & -\dot{\sigma}(t)\sigma(t) \nabla_{\bx_t} \log p_t(\bx_t) dt\\
     & - \beta(t)\sigma(t)^2 \nabla_{\bx_t} \log p_t(\bx_t) + \sqrt{2\beta(t)\sigma(t)} d \pmb{\omega}_t.
    \end{split}
\end{equation}
In this context, $d \pmb{\omega}_t$ represents the standard Wiener process, and $p_t(\bx_t)$ denotes the distribution of the perturbed samples, with initial and final conditions $p_0 = p_{data}$ and $p_1 = \mathcal{N}(0, \sigma^2_{max}I)$, respectively.

To sample from diffusion models, one must solve the diffusion ODE or SDE in reverse, moving from $t = 1$ back to $t = 0$. This process relies on the time-dependent score function $\nabla_{\bx_t} \log p_t(\bx_t)$, which is approximated by a denoiser $v_\theta(\bx_t, t)$. This denoiser is trained to predict the original clean samples $\bx$ from their noisy counterparts $\bx_t$.
The framework also supports conditional generation by employing a denoiser $v_\theta(\bx, t, y)$ that incorporates additional input signals $y$, such as class labels or textual prompts, allowing for more controlled and specific sample generation.

\mypar{Background on flow matching}
Flow matching approaches aim to learn a velocity field $v_t$ mapping random noise $\bx_0 = \pmb{\epsilon} \sim \mathcal{N}(0, I)$ to data samples $\bx_1 \sim p_\text{data}(\bx)$.
Such a mapping is obtained by solving an ordinary differential equation (ODE) of the form:
\begin{equation} \label{eq:flow_ode}
    \frac{d\bx_t}{dt} = v_t(\bx_t).
\end{equation}

\cite{lipman2023flow} provide a simple simulation-free training objective for flow generative models by directly regressing the velocity field $v_t$ on a conditional vector field $u_t(.|\bx_1)$ :
\begin{equation}
    \mathbb{E}_{t,q(x_1), p_t(x_t|x_1)} \lVert v_t(\bx_t) - \bu_t(\bx_t|\bx_1) \rVert^2,
\end{equation}
where $\bu_t(.|\bx_1)$ uniquely determines a conditional probability path $p_t(.|\bx_1)$ towards data sample $\bx_1$.
A popular choice for the conditional probability path corresponds to a linear interpolation between data and noise
 $
    \forall t \in [0,1],\ \bx_t = t \bx_1 + (1-t) \bx_0
$
 resulting in a conditional probability path of the form $\bu(\bx_t|\bx_1) = \bx_1 - \bx_0$.
 Once the conditional probability path is learned, sampling from the model can be achieved by solving the ODE defined in \Cref{eq:flow_ode} using any appropriate ODE solver from the literature.

\mypar{Latent diffusion models (LDMs)}
\cite{ldm}  proposed to train diffusion models  in a latent space induced by a pretrained and frozen autoencoder.
The autoencoder converts all images into latents of smaller resolution, inducing a space in which the generative model is trained, afterwards the latents can be converted back into image space by using the decoder of the autoencoder.
Such models allow to significantly scale up model training as the effective number of tokens representing each data point is reduced significantly. For example, recent autoencoders have a downscaling factor of $8 \times 8$, resulting in a reduction of the sequence length by a factor of 64.

\subsection{Rank scoring} \label{sec:rankscoring}
As different guidance methods require their own set of  hyperparameters, and may respond differently to shared hyperparameters, we select the best hyperparameter set for each method using a global score.
The global score is computed as the average ranking across all tracked metrics, where the ranking is determined within the pool of tested hyperparameter combinations. 

\section{Limitations}

While our method is successfully able to boost the Pareto fronts for quality-consistency-diversity, tradeoffs between these metrics still subsist, although to a lesser extent, when using \OURS.
Future work should explore whether these tradeoffs are inherent to the model or a result of subpar sampling methods.

Additionally, our method operates under the assumption that the model follows the diffusion transformer architecture.
As our method manipulates the energy landscape in attention layers, it needs to be tuned for different architectures containing a different number of attention layers.
It is therefore not directly applicable to models that do not make use of attention layers in their architecture, such as U-Nets without self-attention.

Our method was only tested on diffusion transformers with standard architecture shapes (presenting a depth between $16$ and $38$), therefore extrapolating our claims to extremely deep/shallow models is not evident without supporting experiments.
However, we believe such cases to be less relevant currently as they are not usual choices in state-of-the-art generative models.

\section{Interpreting ERG}
In the following, we provide two interpretations of ERG based on entropy regularized RL and variational inference.
Let $p$ be the density function for the strong model and $p^{\tau}$ the one induced by the attention rectification mechanism we introduce in the paper.

\mypar{I-ERG}
We begin with the I-ERG guidance update, which modifies the model’s score estimate as:
\begin{equation}
\nabla_{z}\log p^{\tau}(\bz)
=
\nabla_{z}\log p(\bz)
\;+\;
w\,\big(
\nabla_{z}\log \rho(\bz)
-
\nabla_{z}\log p^{\tau}(\bz)
\big).
\end{equation}
Rewriting the right-hand side, we obtain:
\begin{equation}
\nabla_{z}\log p^{\tau}(\bz)
=
\nabla_{z}\log p(\bz)
\;+\;
w\,\nabla_{z}\log\!\left(\frac{\rho(\bz)}{p^{\tau}(\bz)}\right).
\end{equation}
Integrating both sides with respect to $\bz$, we find
$p^{\tau}(\bx)\propto p(\bx)\cdot \exp\big(w\cdot R(\bx)\big)$,
where 
$R(\bx)=\log\!\left(\frac{\rho(\bx)}{p(\bx)}\right)$.

This form reveals that I-ERG guidance parallels a posterior distribution that re-weights the base model’s density $p(\bx)$ by an exponential function of the reward signal $R(\bx)$.
This exponential reweighting aligns with objectives used in \emph{entropy-regularized RL} or \emph{KL control}, where the optimal distribution maximizes an expected reward under a KL penalty with respect to a prior:
\begin{equation}
\pi^{\star}=\arg\max_{\pi}\,
\mathbb{E}_{x\sim\pi}[R(\bx)]
-\frac{1}{\lambda}\mathrm{KL}(\pi||p),
\end{equation}
where $\lambda$ serves as an inverse temperature controlling the sharpness of reward influence. The solution to this optimization problem has the form:
\begin{equation}
\pi(\bx)\propto p(\bx)\cdot \exp\big(\lambda R(\bx)\big).
\end{equation}

This shows that the I-ERG update implicitly implements the maximum entropy policy framework, steering the sampling distribution toward high-reward regions while maintaining proximity to the base model $p(\bz)$. Equivalently, using Bayes rule, it can be viewed as defining a \emph{joint distribution} over $\bx$ under $p(\bx)$ and the energy induced by reward $R(\bx)$ (assuming conditional independence).

\mypar{C-ERG}
In the conditional setting, assume we start from a model $p(\bz| \bs)$ conditioned on some semantic signal $\bs$. Let $\bs^{\tau}=\bs+\Delta \bs$ represent a degraded or weakened version of the conditioning (e.g., via temperature scaling or perturbation).
The C-ERG guidance update is given by:
\begin{equation}
\nabla_{z}\log p^{\mathrm{ERG}}(\bz\mid \bs,\bs^{\tau})
=
\nabla_{z}\log p(\bz\mid \bs)
+
w\cdot
\big(\nabla_{z}\log p(\bz\mid \bs)
-
\nabla_{z}\log p(\bz\mid s^{\tau})\big).
\end{equation}

Or, equivalently:
\begin{equation}
\nabla_{z}\log p^{\mathrm{ERG}}(\bz\mid \bs,\bs^{\tau})
=
(1+w)\,\nabla_{z}\log p(\bz\mid \bs)
-
w\,\nabla_{z}\log p(\bz\mid \bs^{\tau}).
\end{equation}

To analyze this further, we use a first-order Taylor expansion of the degraded score $\nabla_{z}\log p(\bz\mid \bs^{\tau})$ around $\bs$:
\begin{equation}
\nabla_{z}\log p(\bz\mid \bs^{\tau})
\approx
\nabla_{z}\log p(\bz\mid \bs)
+
\frac{\partial}{\partial \bs}\big(\nabla_{z}\log p(\bz\mid \bs)\big)\cdot \Delta \bs.
\end{equation}

Substituting this into the update:
\[
\nabla_{z}\log p^{\mathrm{ERG}}(\bz\mid \bs,\bs^{\tau})
\approx
\nabla_{z}\log p(\bz\mid \bs)
-
w\cdot
\left(
\frac{\partial}{\partial \bs}\big(\nabla_{z}\log p(\bz\mid \bs)\big)\cdot \Delta \bs
\right).
\]

Thus, C-ERG introduces a \emph{Jacobian-level product correction}, nudging the sample along directions where the score is most sensitive to perturbations in the conditioning. This helps reinforce fine-grained semantic detail in generation, even when the conditioning signal is noisy or coarsened.

This Jacobian-product reward has a natural interpretation: it reflects how the image-level score $\nabla_{z}\log p(\bz\mid \bs)$ changes when the conditioning $\bs$ is slightly perturbed. This resembles \emph{classifier guidance}, where one steers the generation based on $\nabla_{z}\log p({\bf y} \mid \bz)$—the gradient of a classifier with respect to the image.

Here, the conditioning $\bs$ is a continuous vector, and C-ERG guides generation using:
\begin{equation}
\frac{\partial}{\partial s}\big(\nabla_{z}\log p(\bz\mid \bs)\big)\cdot \Delta \bs,
\end{equation}
which is the direction in image space most sensitive to changes in the semantics of $\bs$. This amounts to amplifying features that are especially responsive to semantic precision.

\section{Additional Experiments}

\subsection{C-\OURS Impact on diversity}
To better understand the effect of c-\OURS on diversity, we perform a simple experiment where we track the initial velocity direction during sampling and how it varies when the initial noised input $\bx_1 = \pmb{\epsilon} \sim \mathcal{N}(0, I)$ varies.

\begin{SCfigure}[2]
    \centering
    \includegraphics[width=.37\linewidth]{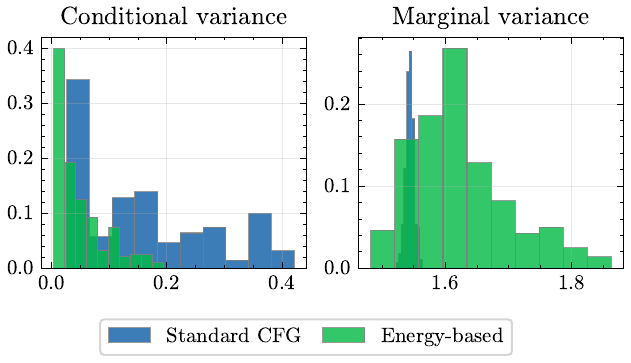}
    \caption{{\bf Effect of C-\OURS on initial velocity step.}
    Standard CFG shows very localized initial marginal variance while the conditional variance is much larger. 
    This means that the negative model predicts an initial velocity that is very similar when varying the initial noise (and prompt in case o C-\OURS), and is largely decorrelated from the conditional prediction, leading to a lack of diversity in the generated samples.
    Conversely, C-ERG results in much higher marginal variance and smaller conditional variance, reducing the error accumulation that can happen at earlier timesteps and leading to better diversity.
    }
    \label{fig:text_temp_grad}
\end{SCfigure}

We examine the variance of the predicted  velocity at the start of sampling, conditioning  either on the encoding of an empty prompt $\pmb{\phi}_\emptyset$ or on the encoding of caption using temperature rescaled attention layers $\pmb{\phi}^\tau_c$. 
For this we sample $N=20$ random noise inputs and compare the variance when the inputs vary, the right panel of \Cref{fig:text_temp_grad}  presents a histogram of the variances where each datapoint corresponds to different spatial/channel location; we refer to it this as {\it marginal} variance:
\begin{equation}
    \begin{cases}
    \mathrm{Var}_{\substack{{\epsilon \sim \mathcal{N}(0,I)}\\ {c \sim D_c}}} \left[  v_\Theta(\pmb{\epsilon}, \pmb{\phi}_\emptyset, t=0) \right],\\ 
    \mathrm{Var}_{\substack{{\epsilon \sim \mathcal{N}(0,I)}\\ {c \sim D_c}}} \left[ v_\Theta(\pmb{\epsilon}, \pmb{\phi}^\tau_c, t=0)
 \right],
    \end{cases}
\end{equation}
where $D_c$ is a dataset of text prompts.

Similarly, in the left panel of \Cref{fig:text_temp_grad}   we consider the variance of the difference between these  velocity estimates and the true conditional one, \ie when we condition on $\pmb{\phi}_c$; we refer to this as {\it conditional} variance:
\begin{equation}
    \begin{cases}
        \mathrm{Var}_{\substack{{\epsilon \sim \mathcal{N}(0,I)}\\ {c \sim D_c}}} \left[ v_\Theta(\pmb{\epsilon}, \pmb{\phi}_c, t=0) -v_\Theta(\pmb{\epsilon}, \pmb{\phi}_\emptyset, t=0) 
     \right],\\
    \mathrm{Var}_{\substack{{\epsilon \sim \mathcal{N}(0,I)}\\ {c \sim D_c}}} \left[ v_\Theta(\pmb{\epsilon}, \pmb{\phi}_c, t=0) -v_\Theta(\pmb{\epsilon}, \pmb{\phi}^\tau_c, t=0) \right].
    \end{cases}
    \label{eq:cond_var}
\end{equation}

From \Cref{fig:text_temp_grad}, we observe that standard classifier-free guidance results in a larger  range for the conditional variance while the marginal variance is significantly reduced.
In contrast, C-\OURS exhibits a larger range of marginal variances but a smaller range of conditional variances.
Such results indicate that standard classifier-free guidance with high guidance scales operate as an initial condition for the flow ODE (\cref{eq:flow_ode}) that has low variance, resulting in exploring only a subset of the solutions of the ODE.
Furthermore, as the initial conditional velocity is much larger, it can easily cause overshooting problems as large updates are being taken with a flawed prediction because of imperfections in the model, while the low marginal variance results in oversaturated colors and simplified image compositions. Similar observations were made by \citet{saharia_deep_understanding, kynkaanniemi2024applying, sadat2024apg}.

Qualitative results corroborate this, see for example \Cref{fig:showcase_method} where using standard CFG results in low diversity, saturated images with simple compositions, while our \OURS is able to generate more complex scenes that suffer less from the ``{\it cartoonish}'' effect, especially for highly out-of-distribution prompts where the denoiser model is expected to be less accurate.

\subsection{I-\OURS impact on quality}

\begin{wrapfigure}{r}{0.35\textwidth}
    \centering   
    \vspace{-1.3em}
    \includegraphics[width=\linewidth]{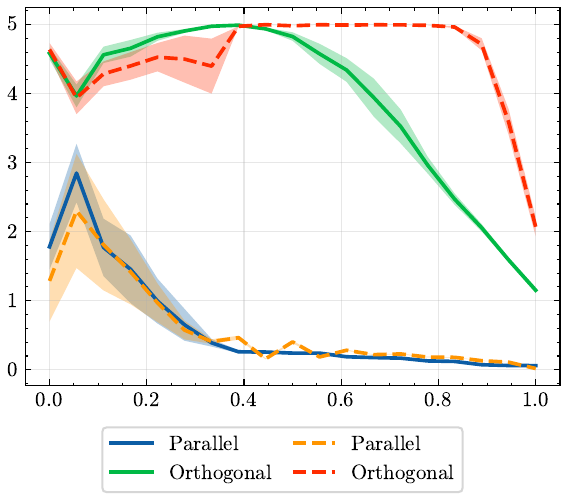}
    \caption{{\bf Orthogonal and parallel differences during sampling.}
    Solid lines correspond to the parallel and orthogonal differences between the conditional and unconditional prediction of the model under standard CFG while dashed lines correspond to those of I-\OURS.
    The sampling process proceeds from left ($t=0)$ to right ($t=1$) over the horizontal axis. }
    \label{fig:erg_components}
    \vspace{-2em}
\end{wrapfigure}
To better understand the differences between  I-\OURS and CFG, we consider the difference between the two denoising terms used in CFG and 
 I-\OURS, see \Cref{eq:cond_var}.
In particular we decompose the difference into parallel and orthogonal components to the conditional prediction, and track the magnitude of these components throughout the sampling process.

Results in  \Cref{fig:erg_components} reveal a fundamental difference between  I-\OURS and CFG.
The parallel differences show similar trends between CFG and I-\OURS:  a peak around $t=0.075$ followed by a sharp decline towards zero, indicating that for approximately $t \geq 0.4$ the denoising task becomes easy enough that both conditional and unconditional predictions yield similar and correlated results.
On the other hand, when examining the orthogonal component of the differences, we see that it keeps a steady norm through sampling while in the case of standard CFG, the norm of the orthogonal difference quickly decreases as sampling progresses for $t \geq 0.4$, this indicates strong correlations between conditional and unconditional predictions as the denoising task becomes easier and the denoiser needs less to rely on the conditional embeddings.

While \cite{sadat2024apg} completely eliminate the parallel component  from the  difference term, we are able to break such correlations by manipulating the energy landscape of the attention layers of the denoiser.
Imposing this divergence between the positive and negative predictors results in semantically meaningful errors for the negative velocity prediction which, when used as a guidance term, improves the low-level details in the resulting image.

Notice that the I-\OURS kickoff threshold $\kappa$ corresponds to the point where the parallel difference between the predicted velocities converges towards 0 while the orthogonal part starts decreasing in a quasi-linear manner, this indicates that the optimal value for $\kappa$ can be obtained with a simple analysis of the correlation between positive/conditional and negative/unconditional predictions of the model.

\subsection{Attention rectification mechanisms}

In \Cref{tab:smoothing_vs_temp}, we compare different attention rectification mechanisms form the litterature, including energy smoothing (from SEG)~\citep{hong2024seg}, identity mapping (from PAG)~\citep{ahn2024PAG} and temperature reduction at the denoiser level as is done in I-\OURS.
We use a guidance scale of $5.0$ and $\tau=0.2$.
Our experiments indicate better performance when using low attention temperature when compared to both energy smoothing and identity mapping in the attention.
The most notable difference being in density which is $12$ points higher for temperature reduction compared to energy smoothing and $3$ points higher than identity mapping.
Similarly, temperature reduction achieves the best FID at $14.43$ which is $0.6$ points lower than energy smoothing and $2.2$ points better than identity mapping.
Compared to energy smoothing, Identity mapping, as proposed by \cite{ahn2024PAG}, achieves better precision, density, coverage and CLIPScore than energy smoothing but still underperforms when compared to temperature reduction.
Similar results can be observed qualitatively in \Cref{fig:qual_rectif_mech}, temperature reduction generates images that showcase better realism in terms of the details of the image such as wooden textures on the floor in the first row, and details on the musician's face and the background in the second row.
Similarly, other qualitative results provide evidence supporting this effect, for example in \Cref{fig:qual_uncond} and \Cref{fig:qualitative_comp_in1k_512_xl2} we observe more detailed backgrounds when using I-\OURS than SEG.

\begin{table*}[h]
    \centering
    \caption{{\bf Comparison between different mechanisms for attention rectification.} Conducted when using I-\OURS. 
    }
     {\scriptsize
    \begin{tabular}{lcHHccc}
        \toprule
         &  FID $(\downarrow)$ & Precision $(\uparrow)$ & Recall $(\uparrow)$ & Density $(\uparrow)$ & Coverage $(\uparrow)$ & CLIPSCore $(\uparrow)$\\
         \midrule
        Energy smoothing $\sigma=+\infty$ & $15.06$ & $66.39$ & ${\bf 39.24}$ & $102.82$ & $69.45$ &  $26.60$\\
        Identity mapping & $16.62$ & $69.40$ & $35.37$ & $111.83$ & $70.76$ & $26.42$\\
        Temperature reduction & ${\bf 14.43}$ & ${\bf 69.95}$ & $36.57$ & ${\bf 114.50}$ & ${\bf 71.44}$ & ${\bf 26.71}$\\
        \bottomrule
    \end{tabular}}
    \label{tab:smoothing_vs_temp}
\end{table*}

\begin{SCfigure}[1.3][h]
    \centering
    \includegraphics[width=0.4\linewidth]{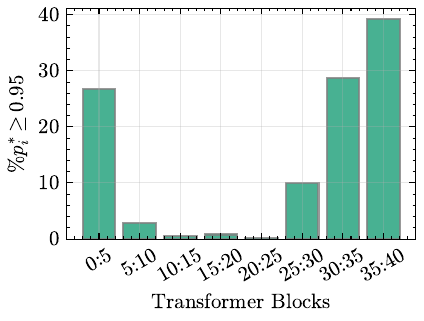}
    \caption{{\bf Percentage of points in the attention matric that have a high maximum probability.} 
    We observe three different regimes depending on the depth of the block.
    For blocks [0-5], the maximal association probability is high with more than $25\%$ of tokens having a maximum probability superior to $0.95$, For blocks 0-25, the association probability is rarely above $95\%$. 
    Finally, the proportion of high certainty associations grows back for deeper layers in the model [25-38], reaching almost $40\%$ for the last layers $35-38$.
    }
    \label{fig:denoiser_probas_grouped}
\end{SCfigure}

\subsection{Attention blocks} \label{sec:att_blocks}
In order to uncover the role of different attention blocks on the sampling process, we track the number of tokens where the maximum probability in the (non-rectified) attention is superior to a threshold of $0.95$, resulting in a near one-to-one matching between the predicted queries and values.

Our results (for $t=0.3$) are summarized in \Cref{fig:denoiser_probas_grouped}.
We find a similar pattern across the different sampling steps.
We observe three different regimes through the different layers of the model.
While the first and last stage have a reasonably high matching probability, the middle stage has much smaller maximum probabilities and thus combines information from multiple tokens. 

\begin{figure}[h]
    \centering
    \includegraphics[width=.7\linewidth]{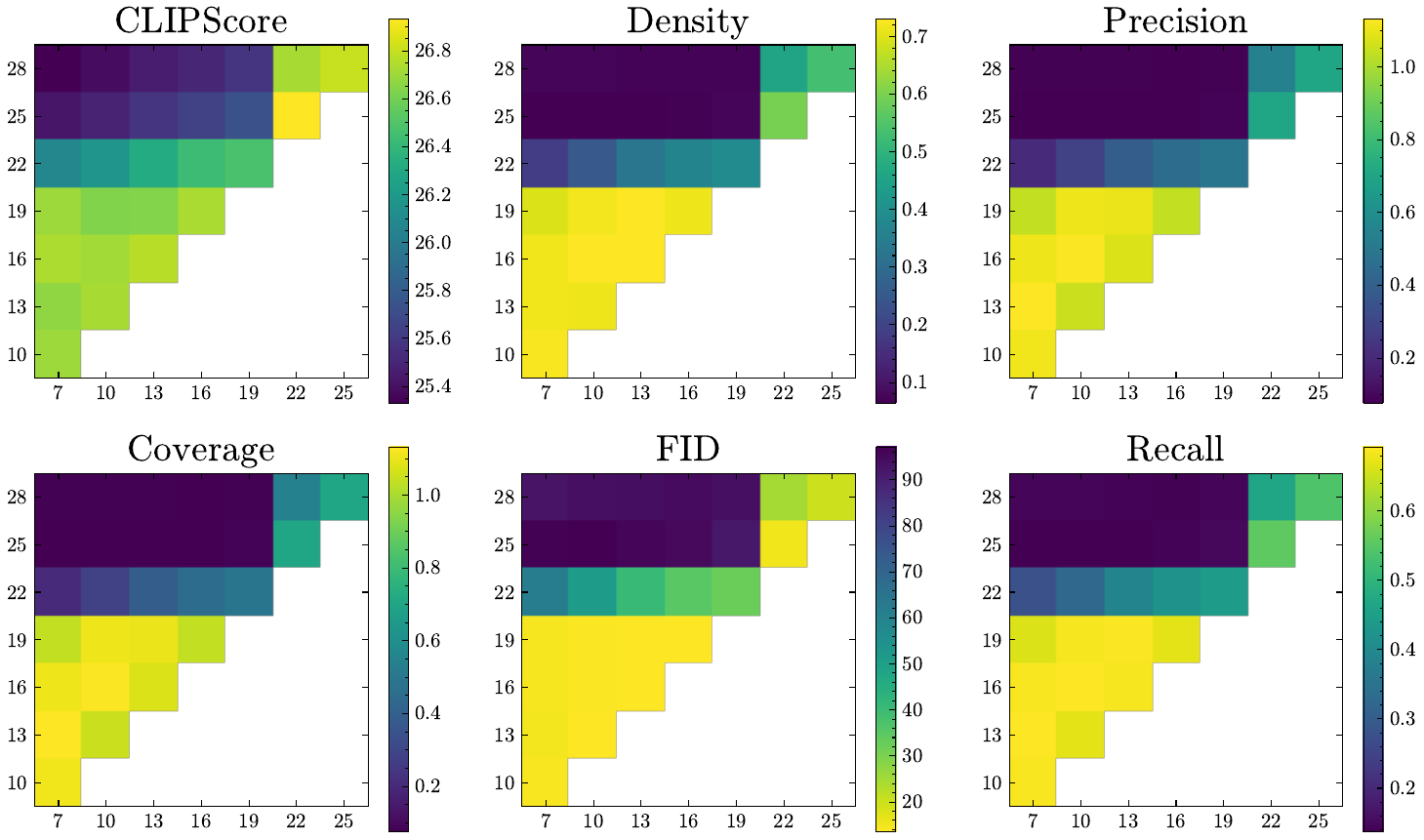}
    \caption{{\bf Choice of rectified attention layers.} Impact of the attention layers where the energy landscape is modified for I-\OURS. (x-axis: start block, y-axis: end block).
    Operating \OURS on middle layers seems to performa favorably, while including later layers results in harmful effects.
    }
    \label{fig:attention_blocks}
\end{figure}

With this in mind, we posit that the energy landscape manipulations should only happen in the middle regime, as it would not overly harm the representations of the model.
To validate this intuition, we perform a grid search over the range of blocks in which the energy manipulation is performed in \Cref{fig:attention_blocks}, we find a similar trend with the emergence of three different regimes, we obtain the best performance when applying the energy guidance on the middle blocks.

\begin{figure*}[ht]
    \centering
    \includegraphics[width=\linewidth]{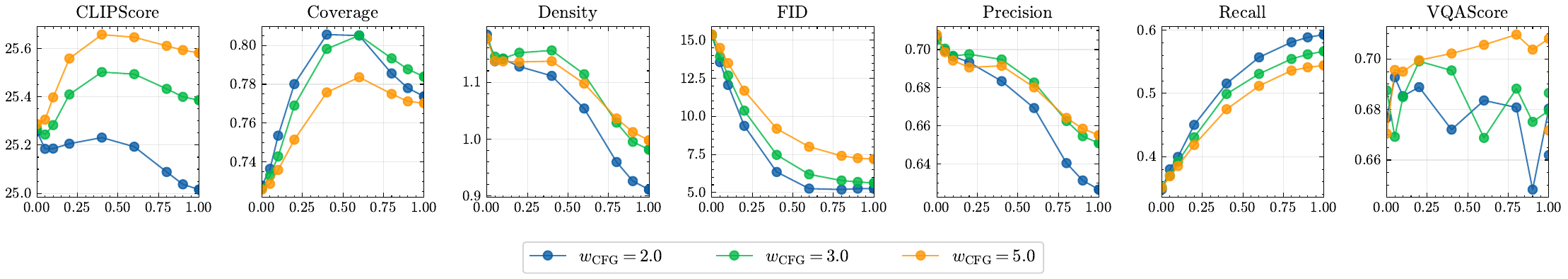}
    \caption{{\bf Influence of I-\OURS kickoff threshold $\kappa$.} Experiment conducted on our T2I model using Euler sampler with $50$ steps. We observe different trends for different metrics, 
    FID and Recall are improved as $\kappa$ goes to $1$, precision and density on the other hand are improved when $\kappa$ goes to $0$, CLIPScore, VQAScore and coverage are optimal for $\tau$ in the middle of the range $\kappa \in [0.2, 0.6]$.}
    \label{fig:im_tau_noapg}
\end{figure*}

\subsection{I-\OURS kickoff threshold }
In \Cref{fig:im_tau_noapg}, we perform a sweep over the kickoff threshold $\kappa$ for the denoiser-level energy guidance, with the temperature rescaling parameter set as $\tau_i=0.01$.

For density and precision, we achieve uniform improvements as $\kappa$ becomes smaller.
For CLIPScore, coverage, and VQAScore, we observe better improvements when $\kappa$ is shifted to the middle of the time range.
For recall and FID, we observe consistent degradation as the range of timesteps in which denoiser energy guidance is applied gets larger.

As also reported by \cite{astolfi2024pareto}, there seems to be a natural trade-off between different facets of the generation {\it i.e} quality, consistency and diversity. 
While lower thresholds achieve the best precision and density ({\it quality}), higher thresholds notably improve coverage, recall and FID ({\it diversity}).
On the other hand, CLIPScore and VQA (consistency) are maximized for a mid level threshold.
This indicates that $\kappa$ can be tuned according to the task at hand to achieve suitable results, we find in our experiments that setting $\kappa \in [0.2, 0.4]$ achieves the most balanced performance across different settings.



\begin{table*}[h]
    \centering
    \caption{{\bf Sampling parameters for different settings.}
    We provide the default hyperparameters used in our experiments grouped by model and modality.
    }
\resizebox{\linewidth}{!}{
    \begin{tabular}{lccccccccccc}
        \toprule
        Task & Text encoder & $\tau$ & $l^i_\text{min}$ & $l^i_\text{max}$ & $\tau_i$ & $l^c_\text{min}$ & $l^c_\text{max}$ & $\tau_c$ & $\alpha$ & $\gamma$ & $K$ \\
        \midrule
        \multirow{2}{*}{{\it Text-to-image (MMDiT-B/2)} }& llama3 & \multirow{2}{*}{0.4} & \multirow{2}{*}{10} & \multirow{2}{*}{17} & \multirow{2}{*}{0.01} & 0 & 32 & 0.01 & \multirow{2}{*}{1.0} & \multirow{2}{*}{1.5} & \multirow{2}{*}{1.0}\\
         & T5 & & & & & 0 & 24 & 0.01 &  &  & \\
         \midrule
        {\it Unconditional (from text2image) (MMDiT-B/2)} & --- & 0.2 & 10 & 17 & 0.01 & --- & --- & --- & 1.0 & 1.0 & 1.0\\
        \midrule
        {\it Class-conditional (DiT-XL/2)} & --- & 0.3 & 12 & 15 & 0.01 & --- & --- & --- & 1.0 & 1.0 & 1.0\\
        \midrule
        \multirow{3}{*}{{\it Text-to-image (SD3-medium)}} & OpenCLIP-ViT/G  & \multirow{3}{*}{0.2} & \multirow{3}{*}{6} & \multirow{3}{*}{8} & \multirow{3}{*}{0.01} & 1 & 31 & 0.1 & \multirow{3}{*}{1.0} & \multirow{3}{*}{1.0} & \multirow{3}{*}{1.0}\\
        & CLIP-ViT/L & & & & & 1 & 10 & 0.1 &  &  &  \\
        & Flan-T5-XL & & & & & 1 & 22 & 0.1 &  &  &  \\
        \multirow{3}{*}{{\it Text-to-image (SD3-medium)}} & OpenCLIP-ViT/G  & \multirow{3}{*}{0.2} & \multirow{3}{*}{10} & \multirow{3}{*}{13} & \multirow{3}{*}{0.01} & 1 & 31 & 0.1 & \multirow{3}{*}{1.0} & \multirow{3}{*}{1.0} & \multirow{3}{*}{1.0}\\
        & CLIP-ViT/L & & & & & 1 & 12 & 0.1 &  &  &  \\
        & Flan-T5-XL & & & & & 1 & 22 & 0.1 &  &  &  \\
        \bottomrule
    \end{tabular}}
    \label{tab:sampling_params}
\end{table*}

\begin{table}[h]
    \caption{{\bf FID tuned metrics.} Instead of considering Pareto optimal parameter set, we tune each of the baselines for optimal FID.}
    \label{tab:fid_optimal}
    \centering
    \scriptsize{
    \begin{tabular}{lcccccccccc}
        \toprule
        Method & CFG & ERG & APG & ERG+APG & PAG & SAG & SEG & CADS & ERG+CADS & AG\\
        \midrule
        FID & 5.25 & 4.93 & 6.62 & 5.24 & 7.00 & 5.28 & 6.72 & 5.12 & 5.00 & 7.11\\
        \bottomrule
    \end{tabular}}
\end{table}
\subsection{Optimizing for FID}
As shown by \cite{astolfi2024pareto}, FID-optimal checkpoints often lie far from the Pareto front when jointly considering realism, consistency, and diversity. 
\cite{dhariwal2021diffusion} further show that lower guidance scales minimize FID and recall, while higher guidance scales improve perceptual sharpness and structure, measured by IS and precision even though they increase FID. 

Consequently, we tune different methods for FID alone, with results reported in \Cref{tab:fid_optimal}. 
CFG achieves its best FID at a guidance scale of 1.25, while ERG achieves its optimal FID of 4.93 using a guidance scale of 1.25 and an ERG scale of 2.0. 
This demonstrates that ERG is capable of outperforming prior guidance methods (e.g., SEG, APG) when optimized for FID alone, in addition to providing better trade-offs under broader evaluation criteria as can be seen in the Pareto fronts in Figure 2 of the main paper.

\subsection{Additional Metrics}
For better comparability, we  extend the results in  \Cref{tab:main_comp} by adding  Inception Score (IS), sFID, and Precision \& Recall in \Cref{tab:additional_metrics}.
\begin{table}[h]
    \centering
    \caption{{\bf Additional metrics.} Comparison of guidance methods. Best results are in bold.}
    \scriptsize
    \begin{tabular}{lcccccccccc}
        \toprule
        Method & CFG & APG & CADS & PAG & SAG & SEG & AutoGuidance & ERG & ERG+APG & ERG+CADS \\
        \midrule
        Precision & 65.71 & 66.34 & 66.42 & 66.42 & 64.97 & 61.22 & 61.48 & 70.92 & 69.50 & \textbf{72.72} \\
        Recall    & 43.95 & 45.15 & 45.75 & 43.94 & 47.98 & 36.88 & 34.60 & 41.43 & \textbf{50.25} & 38.89 \\
        sFID      & 26.53 & 20.31 & 18.20 & 27.21 & 28.55 & 33.80 & 14.75 & 19.56 & \textbf{9.30} & 15.32 \\
        IS        & 37.25 & 43.76 & 40.75 & 40.50 & 38.28 & 40.64 & 41.81 & 43.16 & \textbf{53.47} & 42.52 \\
        \bottomrule
    \end{tabular}
    \label{tab:additional_metrics}
\end{table}
For precision \& recall, we observe that similar trends to density and coverage with ERG+CADS achieving the best precision (72.72). ERG on its own achieves a precision of 70.92 compared to 65.71 for CFG. Recall is slightly lower when using ERG alone (41.43 vs.\ 43.95 for CFG), but ERG+APG achieves the best recall of 50.25. For Inception Score, ERG improves over the CFG baseline (43.16 vs.\ 37.25 for CFG) and achieves the best score when combined with APG (53.47). Simiwlarly, ERG+APG results in significant sFID and Inception score improvement, with ERG alone on par with APG.

\section{Implementation} \label{app:impl_det}
In this section, we provide pseudo-code in the style of pytorch to implement \OURS.

\mypar{I-\OURS}
Changing standard attention mechanism with the energy-based variant defined in \Cref{alg:cap} and applying this for attention blocks between $l_\text{min}$ and $l_\text{max}$ in the negative/unconditional prediction.
Only applied if $t > \kappa$.

\mypar{C-\OURS}
For the negative prediction, instead of having the prediction conditioned on empty text tokens, we condition it on tokens obtained by forwarding the text prompts through the text encoder while changing the temperature of the attention in the text encoders between layers $l^c_\text{min}$ and $l^c_\text{max}$ from $\beta$ to $\beta \cdot \tau_c$.
Applies for all timesteps.

\mypar{\OURS}
Consists in applying both I-\OURS and C-\OURS.

In \Cref{lst:temperature_forward}, we provide a function that applies the temperature rescaling in the attention layers of text-encoders by utilizing forward hooks on the query mapping.
In practice, the temperature rescaling is equivalent to rescaling of the queries prior to the attention operation.

In \Cref{lst:energy_based_attention}, we provide code for the energy-based attention in the case of the mutli-modal transformer block used in \cite{sd3}.

In \Cref{tab:sampling_params}, we provide different hyperparameter sets used for our \OURS method under different settings.
We consider these values as the defaults used for our experiments unless specified otherwise.

\begin{figure*}[h]
\renewcommand{\lstlistingname}{Algorithm}
\lstset{
  language=Python,
  basicstyle=\scriptsize\fontfamily{zi4}\selectfont,
  keywordstyle=\color{blue},
  commentstyle=\color{green!50!black},
  stringstyle=\color{red!50!black},
  numbers=none,
  frame=single,
  rulecolor=\color{gray!20},
  backgroundcolor=\color{gray!10},
  captionpos=b,
}
     
\begin{lstlisting}[caption={{\bf Pseudo-code for temperature scaled encoding of prompts.} Written using Pytorch functionalities.}, label={lst:temperature_forward}]
def forward_with_temperature(transformer, prompts,
                                tau=0.01, l_min=15, l_max = 20):
    handles = []
    
    # define forward_hook_function.
    def hook(module, input, output):
        output[:] *= tau
        return output

    # register forward hooks.   
    for i in range(l_min, l_max):
        handle = transformer.blocks[i].self_attn.q_proj.register_forward_hook(hook)
        handles.append(handle)

    # encode prompts.
    encoder_hidden_states = transformer.encode(prompts)

    # remove registered_hooks.
    for hook in handles:
        hook.remove()
        
    return encoder_hidden_states
\end{lstlisting}
\end{figure*}

\begin{figure*}[h]
\renewcommand{\lstlistingname}{Algorithm}
\lstset{
  language=Python,
  basicstyle=\scriptsize\fontfamily{zi4}\selectfont,
  keywordstyle=\color{blue},
  commentstyle=\color{green!50!black},
  stringstyle=\color{red!50!black},
  numbers=none,
  frame=single,
  rulecolor=\color{gray!20},
  backgroundcolor=\color{gray!10},
  captionpos=b,
}

\begin{lstlisting}[caption={{\bf Pseudo-code for temperature scaled encoding of prompts.}  
The hidden state ${\bf x}$ consists of image tokens concatenated with text/class tokens.}, label={lst:energy_based_attention},]
def multistep_attention(q,k,v,step_size=1.0,steps=1, gamma=1.0):
    q_new = q.clone()
    for i in range(steps):
        att = scaled_dot_product_attention(q_new,k,v)
        q_new = q_new - step_size * (q_new - gamma * att)
    return q_new

def energy_based_attention(self, x, rope_freqs, num_img_tokens, tau_i):
    B, N, C = x.shape
    # B: batch size - N: Number of tokens - C: Hidden dimenstion.
    s1 = num_img_tokens

    q = self.q_linear(x).reshape(
            B, N, num_heads, C // num_heads
        ).permute(0, 2, 1, 3)
    kv = kv_linear(x).reshape(
        B, N, 2, num_heads, C // num_heads
    ).permute(2, 0, 3, 1, 4)
    
    k, v = kv.unbind(0)
    q = self.q_norm(q)
    k = self.k_norm(k)

    # Optionally apply rotary positional embeddings.
    qi, qc = q[:, :, :s1], q[:, :, s1:]
    ki, kc = k[:, :, :s1], k[:, :, s1:]
    qi, ki = apply_rotary_emb(qi, ki, rope_freqs=rope_freqs)

    # Entropy rectification in the image tokens.
    if tau_i > 0:
        qi[:] = qi[:] * tau_i
    
    q = torch.cat([qi, qc], dim=2)
    k = torch.cat([ki, kc], dim=2)
    
    if not self.use_e_att:
        x = scaled_dot_product_attention(q, k, v)
    else:
        x = multistep_attention(
            q,
            k,
            v,
            lr=self.step_size,
            steps=self.num_steps,
            gamma=self.potential_w,
        )
        
    x = x.transpose(1, 2).reshape(B, N, C)
    x = proj(x)
    return x    
\end{lstlisting}
\end{figure*}

\section{Combining different methods}

\mypar{\OURS + APG} APG can be seamlessly integrated with \OURS by switching the guidance update to the APG update, as both conditional and unconditional predictions can be converted into clean latent estimates, the algorithm operates as described by~\citet{sadat2024apg}.
In the case of rectified flows, such a conversion is easily obtained as $\hat{\bx}_0 = \bx_t + (1-t) \cdot \hat{\bf v}_\Theta(\bx_t, \pmb{\phi}, t)$.
After the guidance update, the clean image estimate is projected back into velocity prediction by inverting the formula: $\hat{\bf v}_\Theta(\bx_t, \pmb{\phi}, t) = (\hat{\bx}_0 - \bx_t) / (1-t)$.

\mypar{\OURS + CADS} CADS is also straightforward to integrate with \OURS, for this we switch the text encoder hidden states in the conditional/positive prediction with the interpolation between the tokens and Gaussian noise as described by~\citet{sadat2024cads}.

\section{Reimplementation}
We implement our method on the open-source Stable Diffusion 3 model, we use the \texttt{sd3-medium} from diffusers library.
We refer to \Cref{tab:sampling_params} for details about the choice of hyperparameters.
Qualitative examples on \Cref{fig:sd3_example} show significant improvements in terms of image quality.
In \Cref{tab:sd3_comp}, we report quantitative results on this model on COCO-5k benchmark.
For this experiment we use the standard setup with a guidance scale of $5.0$ and Euler sampler with $28$ steps.
We observe significant improvements for all metrics  when using \OURS compared to standard classifier-free guidance.
More precisely, we observe significant boosts in FID $-10.03$ pts, density $+73.56$, coverage $+13.34$ and CLIPScore $+3.0$.
\begin{table}[h]
    \centering
    \caption{{\bf Effect of \OURS on Stable Diffusion 3 models.} We compare standard classifier-free guidance with our method on COCO-5K benchmark. We observe significant improvements across reported metrics.}
    {\scriptsize
    \begin{tabular}{clcHHccc}
        \toprule
         & & FID $(\downarrow)$ & Precision $(\uparrow)$ & Recall $(\uparrow)$ & Density $(\uparrow)$ & Coverage $(\uparrow)$ & CLIPScore $(\uparrow)$ \\
         \midrule
        \multirow{2}{*}{SD3.0} & CFG & $35.85$ & $40.09$ & ${\bf 43.17}$ & $41.11$ & $10.95$ & $24.71$\\
        & \OURS & ${\bf 25.82}$ & ${\bf 68.59}$ & $38.66$ & ${\bf 114.67}$ & ${\bf 24.29}$ & ${\bf 27.71}$\\
        \midrule
        \multirow{2}{*}{SD3.5} & CFG & $34.33$ & $37.94$ & ${\bf 43.17}$ & $37.94$  & $11.76$ & $28.66$\\
        & \OURS & ${\bf 23.81}$ & ${\bf 88.59}$ & $38.66$ & ${\bf 118.59}$ & ${\bf 25.26}$ & ${\bf 29.61}$\\
        \midrule
        \multirow{2}{*}{PixART-alpha} & CFG & $28.64$ & $47.85$ & ${\bf 43.38}$ & $52.98$ & $13.46$ & $26.03$\\
         & ERG  & ${\bf 27.75}$ & ${\bf 56.40}$ & $42.86$ & ${\bf 70.61}$ & ${\bf 16.57}$ & ${\bf 26.61}$\\
        \midrule
        \multirow{2}{*}{SDXL} & CFG & $21.65$ & ${\bf 60.68}$ & $49.17$ & $80.93$ & $20.00$ & $29.54$\\
         & ERG  & ${\bf 20.94}$ & $60.66$ & ${\bf 54.39}$ & ${\bf 84.05}$ & ${\bf 20.21}$ & ${\bf 29.97}$\\
        \bottomrule
    \end{tabular}
    }
    \label{tab:sd3_comp}
\end{table}

We implement ERG on different architectures beyond the standard DiT.
Specifically, we provide additional (I-)ERG results for SDXL, PixArt-alpha and EDM2 (autoguidance). 
SDXL and EDM2 use U-Net architectures while PixArt-alpha uses local window attention. For SDXL and Pixart-alpha, we use the opensource version from Hugging Face alongside the EvalGIM evaluation framework. 
For EDM2 and DiT, we use the evaluation setups from their respective open-source implementations.

\mypar{On EDM2/Autoguidance}
As reported in \Cref{tab:edm_autoguidance}, \OURS improves over baseline EDM2 and obtains similar FID and slightly better FID\_dinov2 than the autoguidance results, without need for additional models.
\begin{table}[h]
    \caption{{\bf EDM and Autoguidance comparison}}
    \label{tab:edm_autoguidance}
    \centering
    {\scriptsize
    \begin{tabular}{lcc}
        \toprule
        Model/Metric & FID $(\downarrow)$ & FID\_dinov2 $(\downarrow)$\\
        \midrule
        EDM2-IN-XXL-512 & 2.00 & 59.34\\
        EDM2-IN-XXL-512 + ERG & 1.35 & {\bf 31.32}\\
        EDM2-IN-XXL-512  + Autoguidance & {\bf 1.33} & 31.56\\
        \bottomrule
    \end{tabular}}
\end{table}

\mypar{On open source DiT}
For completeness, we provide comparative results of our method applied to the official opensource implementation of DiT~\citep{Peebles2022DiT}.
We re-ran the experiment using the guided-diffusion library to match exactly the evaluation setup from the DiT paper for 10 different random seeds. The results of the rectified experiment are reported in \Cref{tab:dit_opensource} and \Cref{tab:dit_stats}.

\newcommand{\cellstats}[4]{%
  \begin{tabular}{@{}c@{}}%
    \num{#1}\,{\tiny\textcolor{blue}{$\pm$\,\num{#2}}}\\
    {\scriptsize\textcolor{blue}{[\num{#3}--\num{#4}]}}%
  \end{tabular}%
}

\begin{table}[h]
\centering
  \begin{minipage}[t]{0.47\textwidth}
    \centering
    \caption{{\bf Open source DiT results.} Compared with the standard DiT.}
    \vspace{1.5em}
    \resizebox{\textwidth}{!}{%
      \begin{tabular}{lccccc}
        \toprule
        Model/Metric & FID & sFID & Precision & Recall & IS\\
        \midrule
        DiT-XL/2-256 & 2.38 & 4.55 & 82.77 & 57.94 & 277.96\\
        DiT-XL/2-256 + ERG & {\bf 2.15} & {\bf 4.32} & {\bf 86.90} & {\bf 60.03} & {\bf 295.03}\\
        \bottomrule
      \end{tabular}
    }
    \label{tab:dit_opensource}
  \end{minipage}
  \hfill
  \begin{minipage}[t]{0.5\textwidth}
    \centering
    \caption{{\bf Statistical analysis on FID.} We report aggregate statistics over 10 different random seeds, minimum and maximum values are reported between brackets.}
    \resizebox{\textwidth}{!}{%
      \begin{tabular}{lccccc}
        \toprule
        & IS & FID & sFID & Precision & Recall \\
        \midrule
        Baseline &
        \cellstats{278.67}{4.93}{270.67}{273.39} &
        \cellstats{2.38}{0.05}{2.31}{2.48} &
        \cellstats{4.55}{0.12}{4.41}{4.55} &
        \cellstats{82.74}{0.16}{82.48}{82.75} &
        \cellstats{57.96}{0.41}{57.34}{58.50} \\
        I-ERG &
        \cellstats{293.79}{2.06}{289.18}{297.84} &
        \cellstats{2.15}{0.03}{2.09}{2.18} &
        \cellstats{4.28}{0.08}{4.26}{4.33} &
        \cellstats{86.94}{0.09}{86.82}{87.06} &
        \cellstats{59.98}{0.42}{59.21}{60.64} \\
        \bottomrule
      \end{tabular}
    }
    \label{tab:dit_stats}
  \end{minipage}
  \vspace{-0.75em}
\end{table}

For the baseline, we obtain an average FID of 2.38 which is higher than the one reported in the paper but is plausible given the variance of the results.
Compared to the baseline, I-ERG still achieves a better FID (2.15 vs.\ 2.38), and given the statistics, also yields smaller variance on all metrics other than recall, showing more stable performance while varying random seeds.

\section{Assets}
In \Cref{tab:assets} we provide the links to the datasets and models
used in our work and their licensing.

\begin{table}[h]
    \centering
    \caption{{\bf Reference for the different assets used in our work.}}
    {
    \begin{tabular}{lc}
        \toprule
        COCO'14 & \url{https://www.cocodataset.org}\\
        ImageNet & \url{https://www.image-net.org}\\
        CC12M & \url{https://github.com/google-research-datasets/conceptual-12m}\\
        YFCC100M & \url{https://www.multimediacommons.org}\\
        \midrule
        Llama3-8b & \url{https://huggingface.co/meta-llama/Meta-Llama-3-8B}\\
        Flan-T5-XL & \url{https://huggingface.co/google/flan-t5-xl}\\
        \midrule
        Stable Diffusion 3 & \url{https://huggingface.co/stabilityai/stable-diffusion-3-medium}\\
        \midrule
        EvalGIM & \url{https://github.com/facebookresearch/EvalGIM}\\
        \bottomrule
    \end{tabular}}
    \label{tab:assets}
\end{table}

\section{Additional qualitative results}
We provide additional qualitative examples of our model under different settings.

\begin{figure*}[t]
    \renewcommand{\arraystretch}{0.5}
    \setlength\tabcolsep{1pt}
    \setlength\fboxsep{0pt}
    \def\myim#1{\fbox{\includegraphics[width=33mm,height=33mm]{figures/examples/dog/img_#1.png}}}
    \def\myimc#1{\fbox{\includegraphics[width=33mm,height=33mm]{figures/examples/cat_sign2/img_#1.png}}}
    \def\myimh#1{\fbox{\includegraphics[width=33mm,height=33mm]{figures/examples/horse/img_#1.png}}}
\resizebox{\textwidth}{!}{
    \begin{tabular}{cccccc}
    \multicolumn{6}{c}{\shortstack{{\it``A french bulldog wearing a red sweater and a blue hat, sitting at a table with a person's hand}\\ {\it holding a doughnut in front of it. The background of the image is a wall.''}}}
    \vspace{2pt}\\
     \begin{sideways} {\it Classifier-free guidance} \end{sideways}   
     & \myim{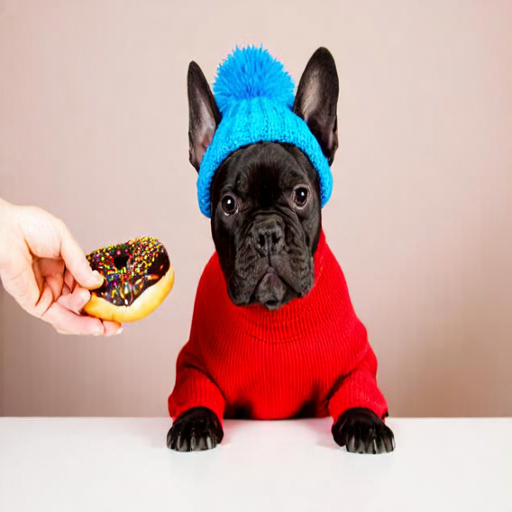}     
     & \myim{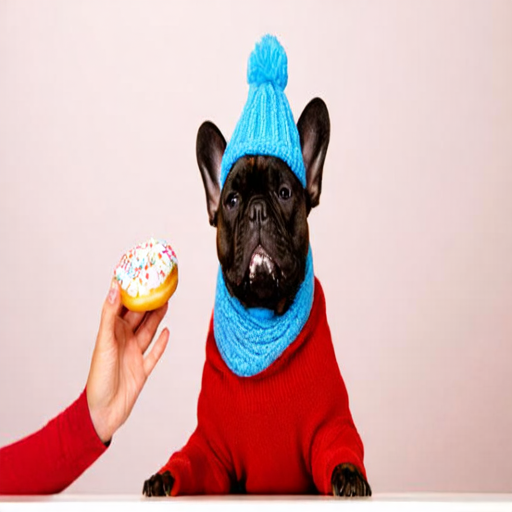}     & \myim{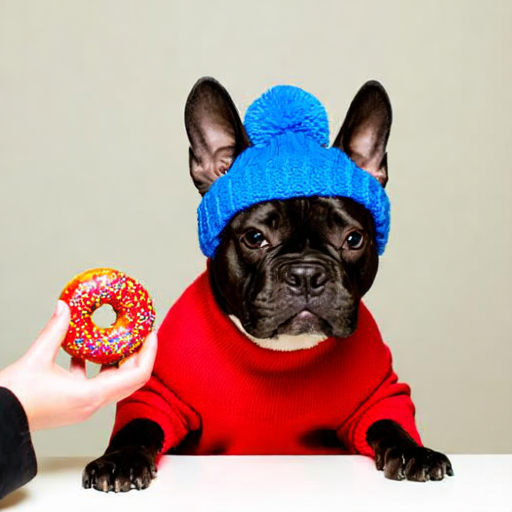}
     & \myim{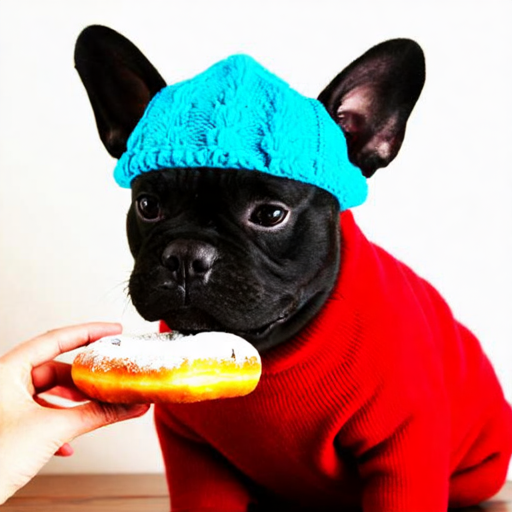}     & \myim{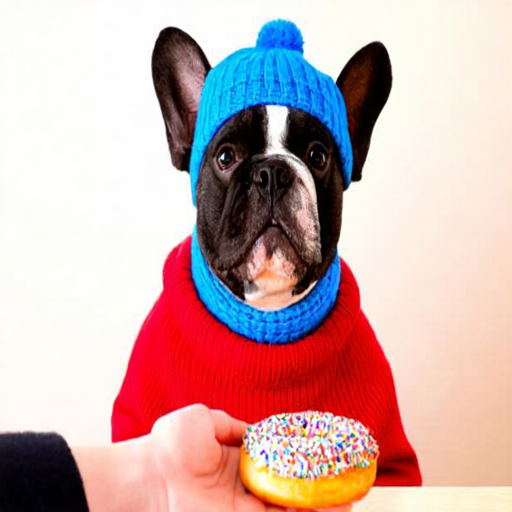}
     \\
     \begin{sideways} {\it \OURS} \end{sideways}   
     & \myim{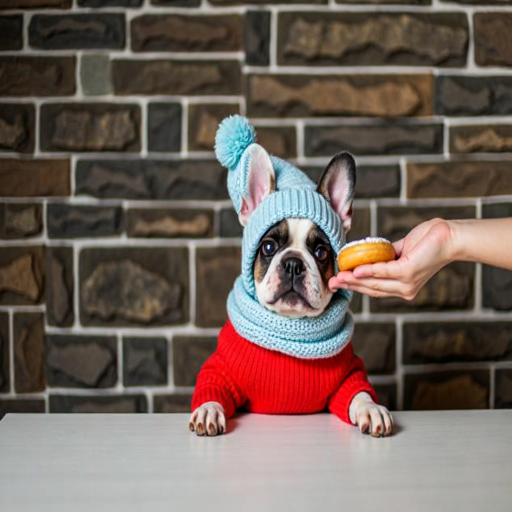}     
     & \myim{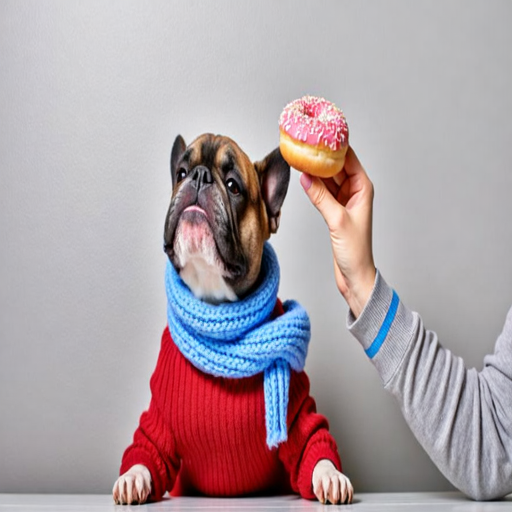}     & \myim{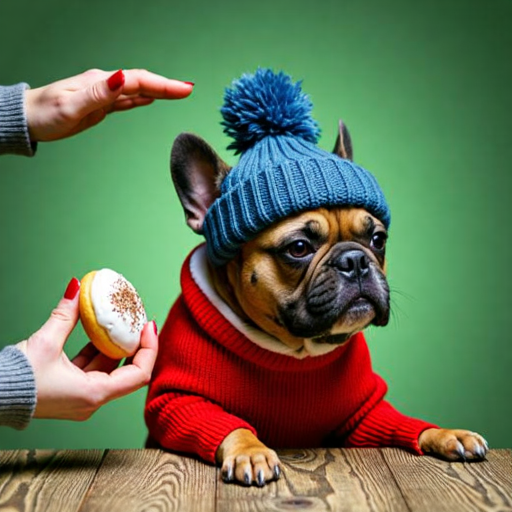}
     & \myim{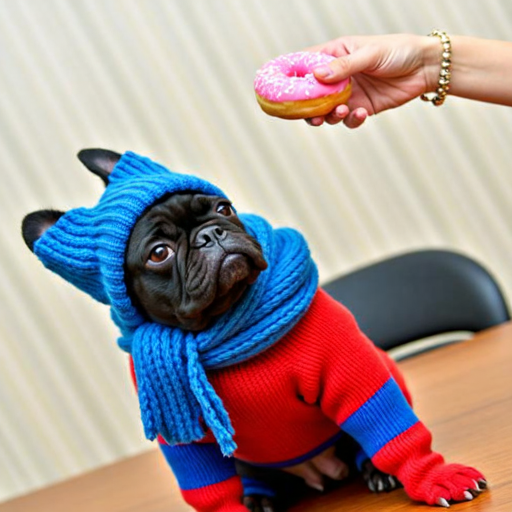}     & \myim{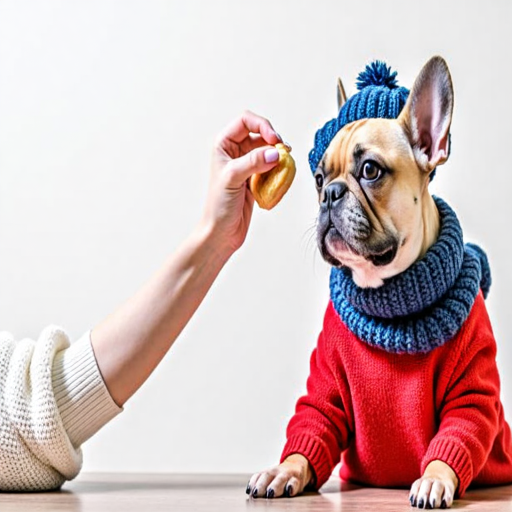}
    \\\\
     \multicolumn{6}{c}{{\it``A small kitten holding a sign. The sign shows a fish drawing.''}}\\
     \begin{sideways} {\it Classifier-free guidance} \end{sideways}   
     & \myimc{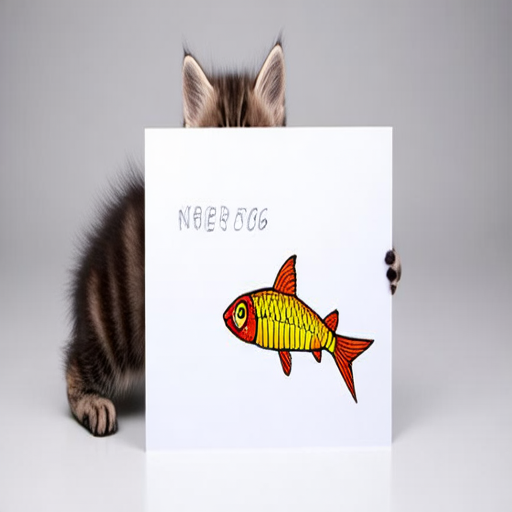}     
     & \myimc{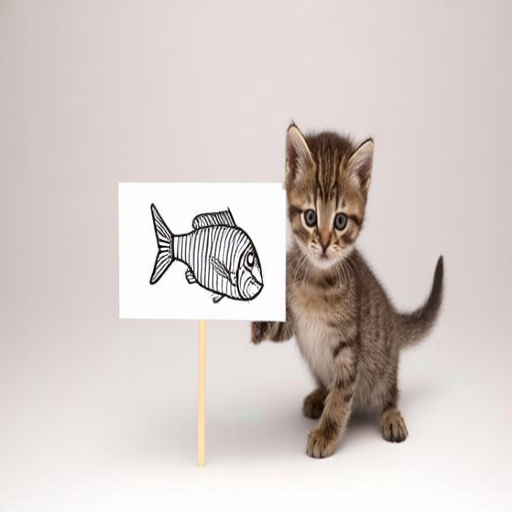}     & \myimc{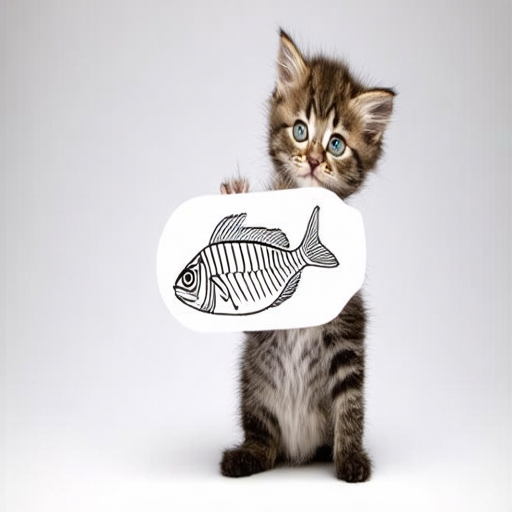}
     & \myimc{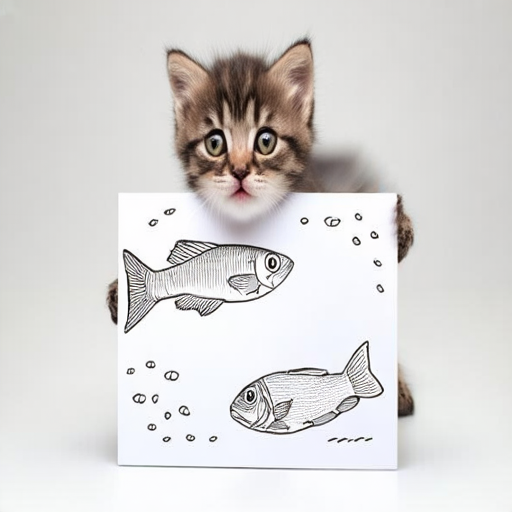}     & \myimc{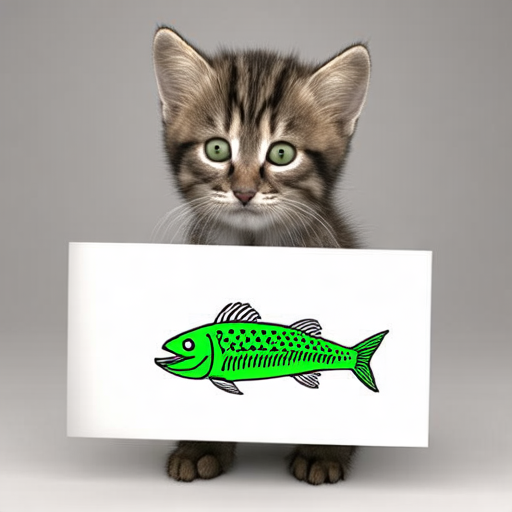}     
     \\
    \begin{sideways} {\it \OURS} \end{sideways}   
     & \myimc{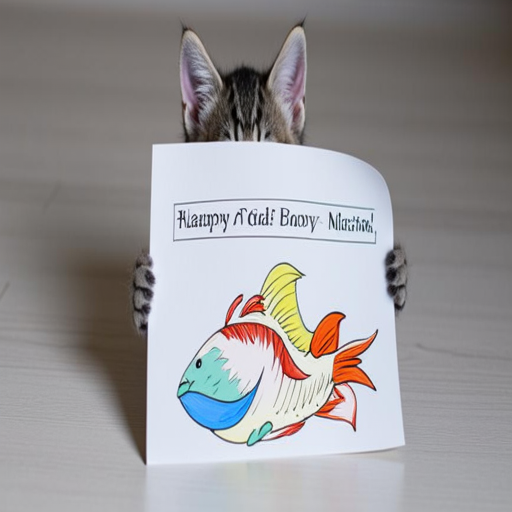}     
     & \myimc{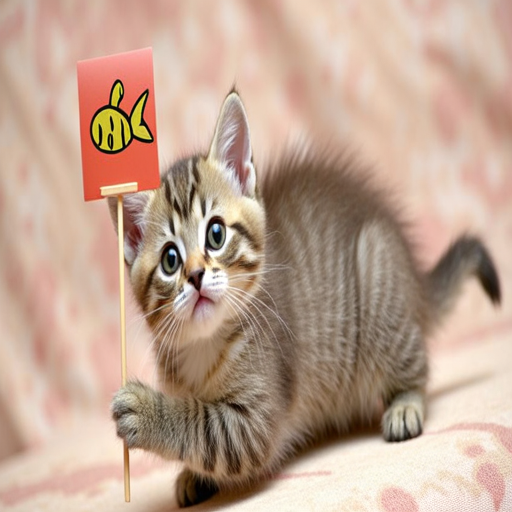}     & \myimc{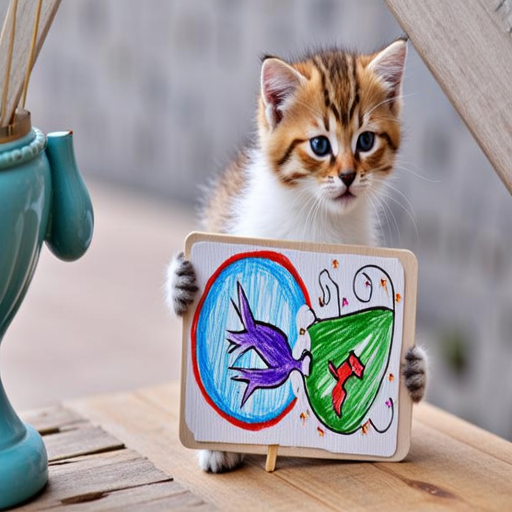}
     & \myimc{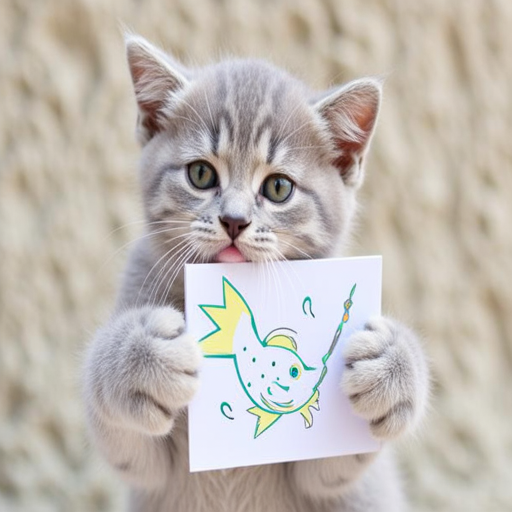}     & \myimc{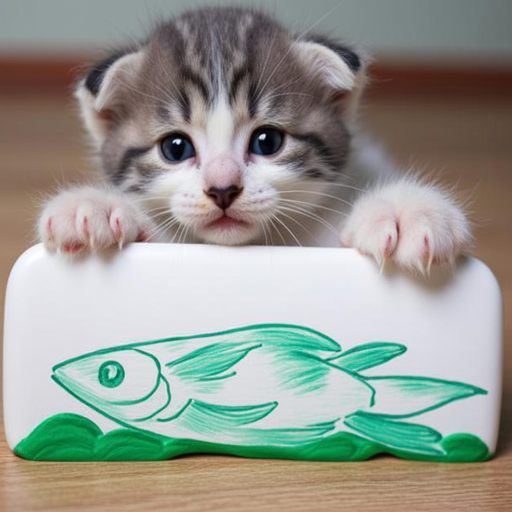}
     \\\\
    \multicolumn{6}{c}{{\it``A horse.''}}\\
     \begin{sideways} {\it Classifier-free guidance} \end{sideways}   
     & \myimh{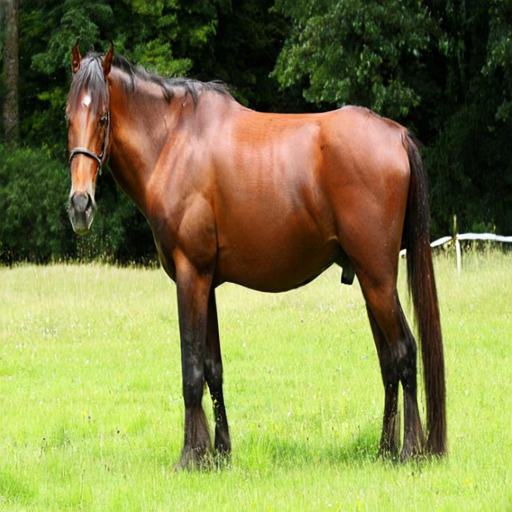}     
     & \myimh{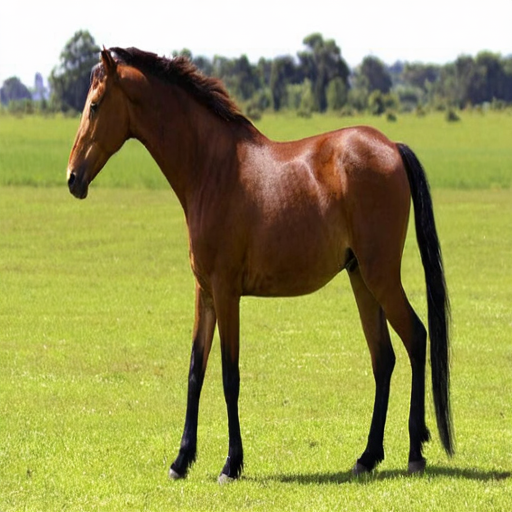}     & \myimh{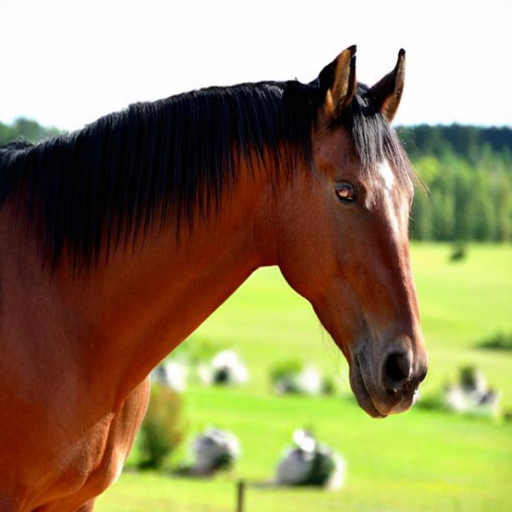}
     & \myimh{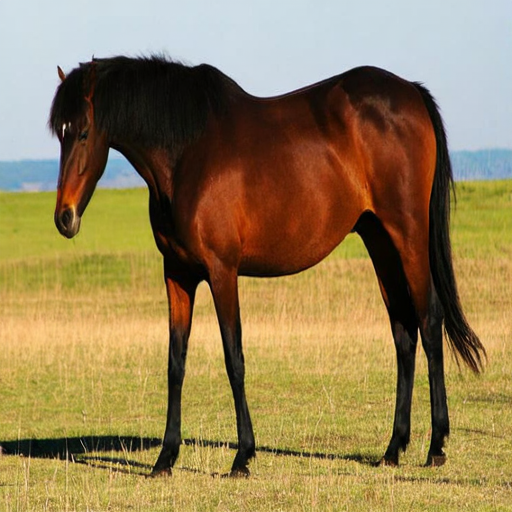}     & \myimh{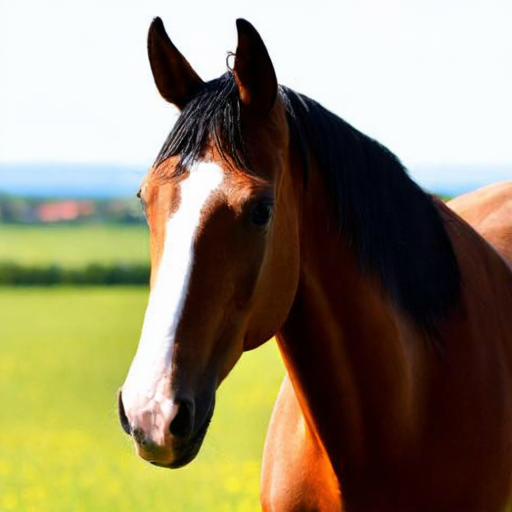}     
     \\
    \begin{sideways} {\it \OURS} \end{sideways}   
     & \myimh{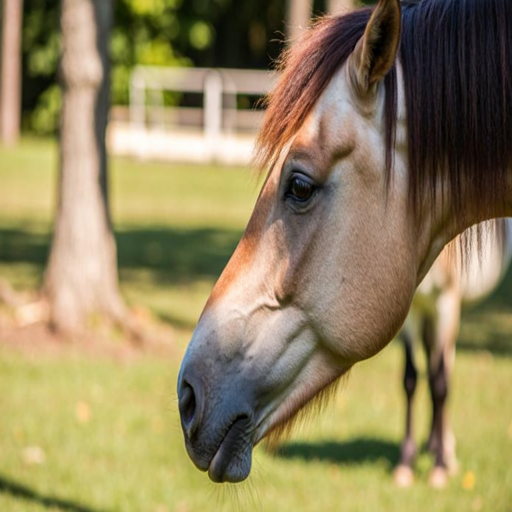}     
     & \myimh{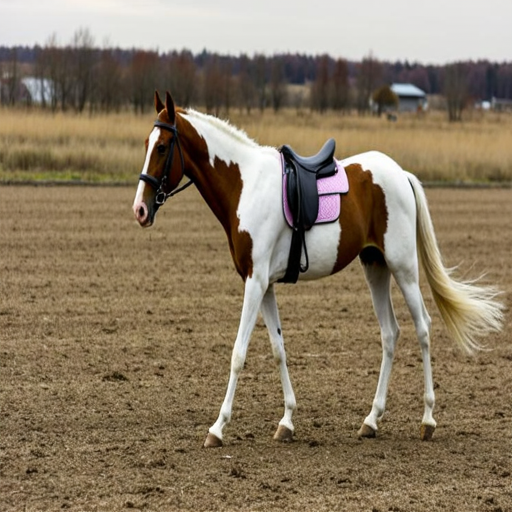}     & \myimh{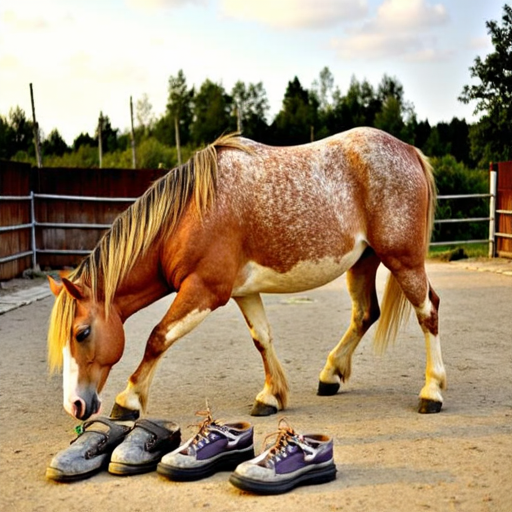}
     & \myimh{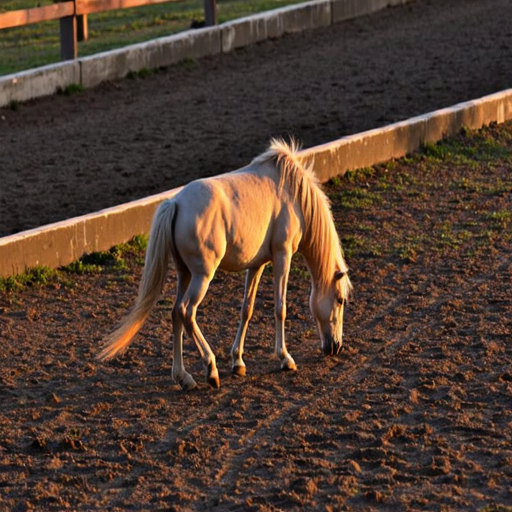}     & \myimh{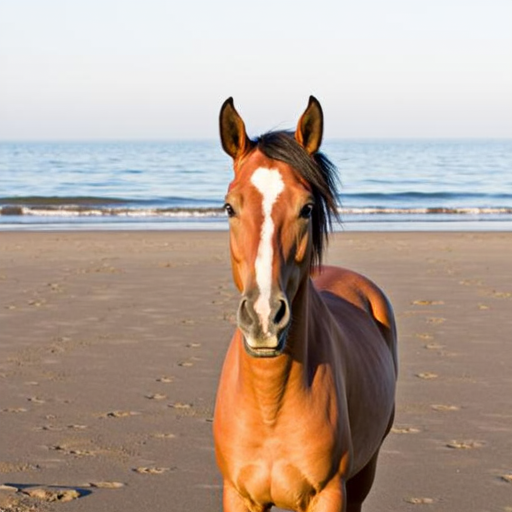}
    \end{tabular}        
}
    \caption{{\bf Qualitative comparison standard classifier-free guidance (top) and \ours (bottom).} 
     Each column represents images generated using the same random seed.
     Images generated with scaled temperature show more complex  textures and variations than the standard guidance.  
    }
    \label{fig:qual_app}
\end{figure*}

In \Cref{fig:qual_app}, we provide a comparison between classifier-free guidance and \ours on three different prompts with varying levels of detail, we showcase the conditional diversity of the methods by providing samples with $5$ different random seeds for each prompt.
For longer prompts (top row), we observe similar consistency but higher diversity in terms of background colors, the dog's physical traits etc.
Similarly, the second row shows increased diversity with ERG in terms of colors, textures and the drawing that the cat is holding.
For very short prompts (third row), we observe certain redundancies in the standard CFG generations while ERG is able to generate high quality samples showcasing higher diversity (depicting the horse against  different backgrounds, different points of view, different time of day \etc).

\begin{figure*}[t]
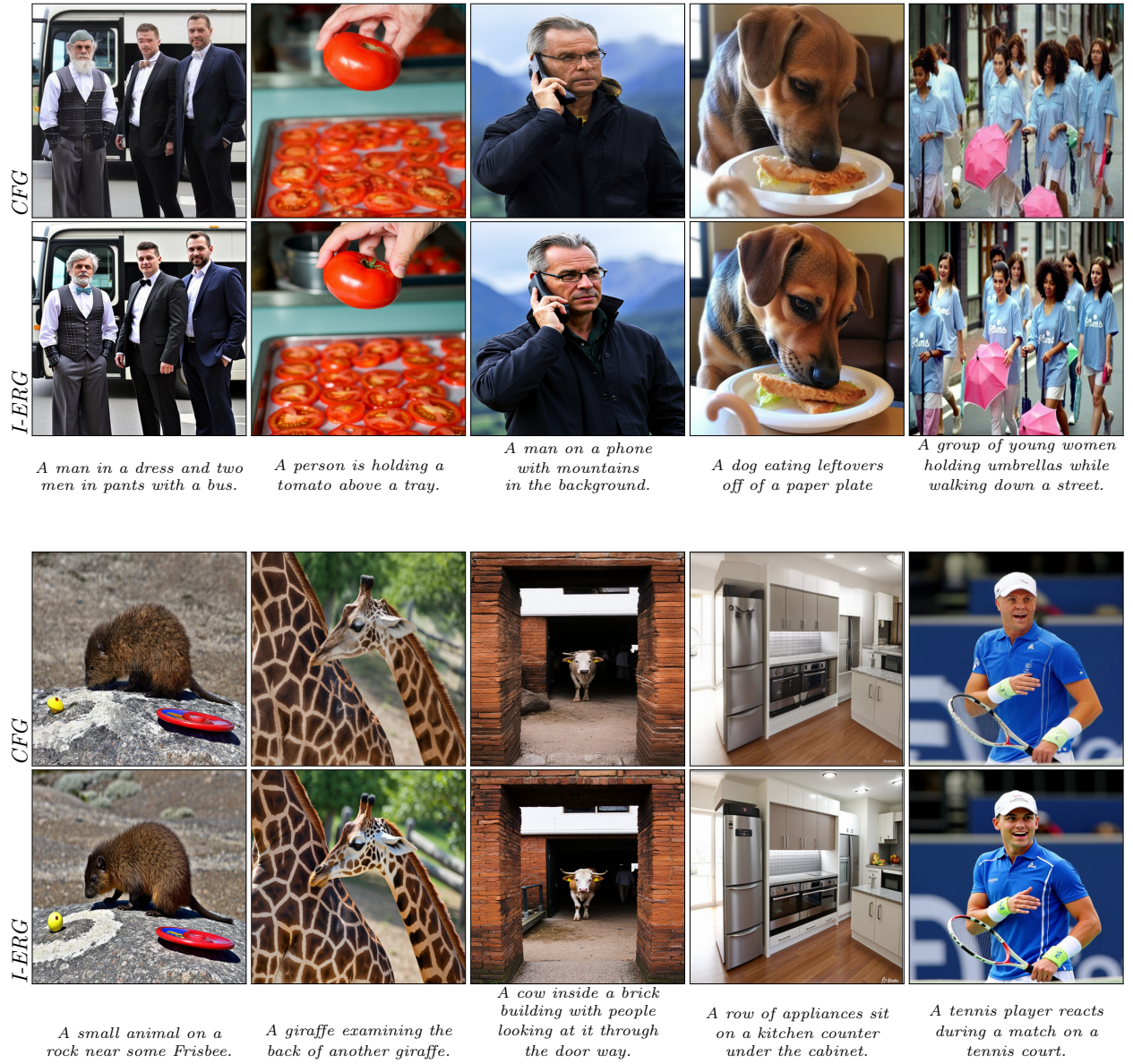


    \renewcommand{\arraystretch}{0.5}
    \setlength\tabcolsep{1pt}
    \setlength\fboxsep{0pt}
    \def\myim#1#2{\fbox{\includegraphics[width=34mm,height=34mm]{figures/examples/eatt_tau/#1/#2.png}}}
    
    \resizebox{\textwidth}{!}{
    \begin{tabular}{cccccc}
     \begin{sideways} {\it CFG} \end{sideways}   
     & \myim{standard}{000053}     
     & \myim{standard}{000159}     & \myim{standard}{000221}
     & \myim{standard}{000249}     & \myim{standard}{000135}
     \\
    \begin{sideways} {\it I-\OURS} \end{sideways}   
     & \myim{eatt}{000053}     
     & \myim{eatt}{000159}     & \myim{eatt}{000221}
     & \myim{eatt}{000249}     & \myim{eatt}{000135}
    \\
     & \scriptsize \shortstack{{\it A man in a dress and two}\\ {\it men in pants with a bus.}} & \scriptsize \shortstack{{\it A person is holding a}\\ {\it tomato above a tray.}} & \scriptsize \shortstack{{\it A man on a phone}\\ {\it with mountains}\\ {\it in the background.}} & \scriptsize \shortstack{{\it A dog eating leftovers}\\ {\it off of a paper plate}} & \scriptsize \shortstack{{\it A group of young women}\\ {\it holding umbrellas while}\\ {\it walking down a street.}}\\
    \vspace{2em}\\
     \begin{sideways} {\it CFG} \end{sideways}   
     & \myim{standard}{000195}     
     & \myim{standard}{000064}     & \myim{standard}{000078}
     & \myim{standard}{000061}     & \myim{standard}{000175}
     \\
    \begin{sideways} {\it I-\OURS} \end{sideways}   
     & \myim{eatt}{000195}     
     & \myim{eatt}{000064}     & \myim{eatt}{000078}
     & \myim{eatt}{000061}     & \myim{eatt}{000175}
    \\
     & \scriptsize \shortstack{{\it A small animal on a}\\ {\it rock near some Frisbee.}} & \scriptsize \shortstack{{\it A giraffe examining the}\\ {\it back of another giraffe.}} &\scriptsize \shortstack{{\it A cow inside a brick}\\ {\it building with people}\\ {\it looking at it through}\\ {\it the door way.}} & \scriptsize \shortstack{{\it A row of appliances sit}\\ {\it on a kitchen counter}\\ {\it under the cabinet.}} & \scriptsize \shortstack{{\it A tennis player reacts}\\ {\it during a match on a}\\ {\it tennis court.}}\\
    \end{tabular}}
    \caption{{\bf Influence of I-\OURS on boosting image quality.} 
    Top row: images generated with standard guidance. 
    Bottom row: images generated with $\alpha=0.01$, a guidance scale of $w_\text{CFG} = 3.0$ and $\tau=0.2$ order to match the image semantic in the comparison.}
    \label{fig:qualitative_comp}
\end{figure*}

In \Cref{fig:qualitative_comp}, we provide a comparison between classifier-free guidance and I-\OURS showcasing the improvements in image quality when generating samples with the same random seed.
By setting $\tau_i=0.2$, the image semantics are not steered excessively compared to the CFG generations, resulting in comparable generations.
We observe noticeable improvements in terms of high level details in the images when using I-\OURS.

\begin{figure*}[t]
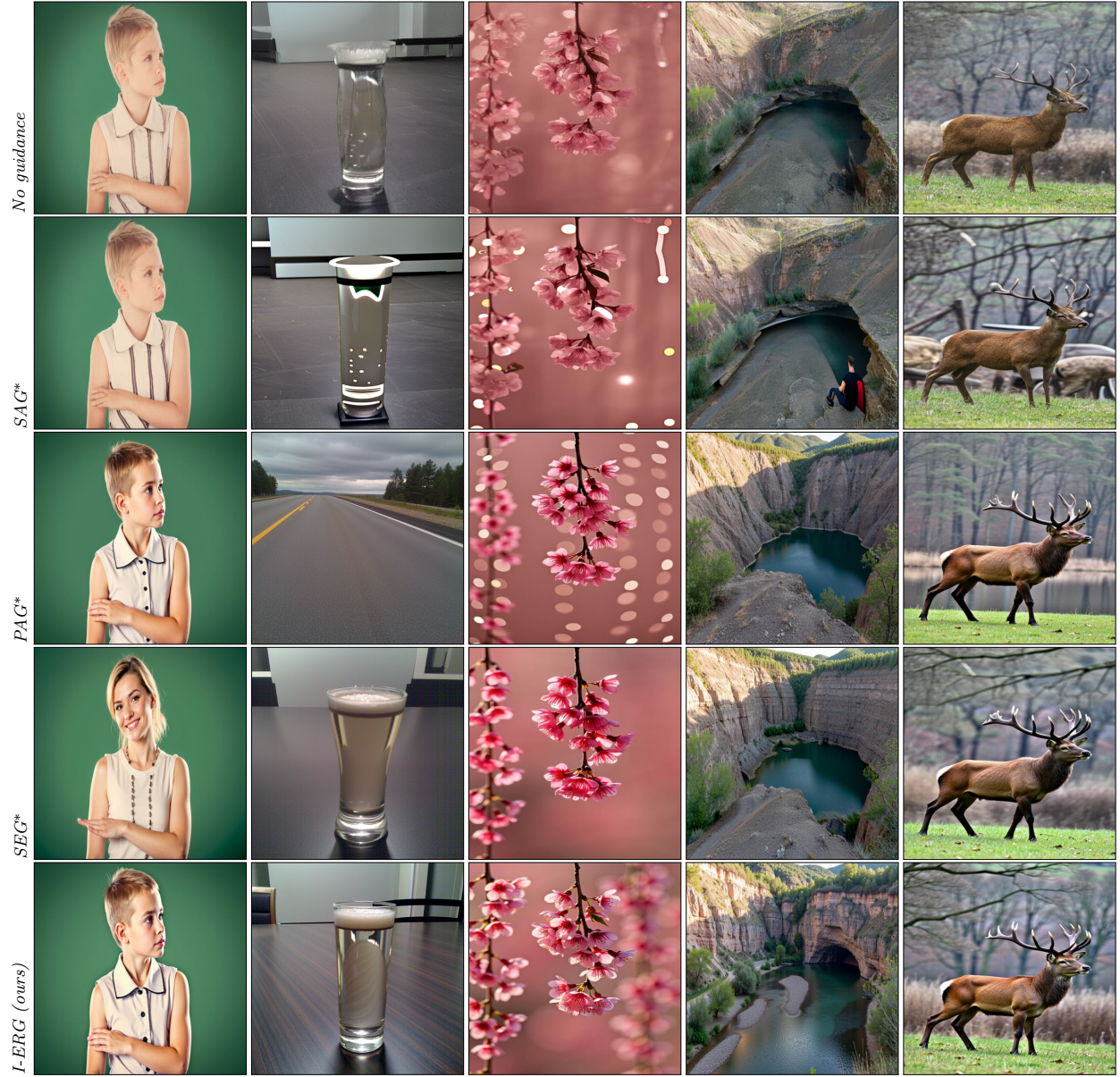

    \renewcommand{\arraystretch}{0.5}
    \setlength\tabcolsep{1pt}
    \setlength\fboxsep{0pt}
    \def\myim#1#2{\fbox{\includegraphics[width=34mm,height=34mm]{figures/uncond_v2/#1/#2.png}}}
    \footnotesize
        \resizebox{\textwidth}{!}{
    \begin{tabular}{ccccccc}
     \begin{sideways} {\it No guidance} \end{sideways}   
     & \myim{base}{000003}     
     & \myim{base}{000006}     & \myim{base}{000008}
     & \myim{base}{000023}     & \myim{base}{000029}
     \\
     \begin{sideways} {\it SAG$^*$} \end{sideways} 
     & \myim{sag}{000003}     
     & \myim{sag}{000006}     & \myim{sag}{000008}
     & \myim{sag}{000023}     & \myim{sag}{000029}
     \\
     \begin{sideways} {\it PAG$^*$} \end{sideways} 
     & \myim{pag}{000003}     
     & \myim{pag}{000006}     & \myim{pag}{000008}
     & \myim{pag}{000023}     & \myim{pag}{000029}
     \\
     \begin{sideways} {\it SEG$^*$} \end{sideways} 
     & \myim{seg}{000003}     
     & \myim{seg}{000006}     & \myim{seg}{000008}
     & \myim{seg}{000023}     & \myim{seg}{000029}
     \\
     \begin{sideways} {\it I-\OURS (ours)} \end{sideways}   
     & \myim{erg}{000003}     
     & \myim{erg}{000006}     & \myim{erg}{000008}
     & \myim{erg}{000023}     & \myim{erg}{000029}
    \end{tabular}        
}
    \caption{{\bf Comparing different methods for unconditional image generation.} Images are sampled from the text-to-image model while conditioning on an empty prompt,  using a guidance scale of $3.0$ and Euler sampler with $50$ sampling steps.
    }
    \label{fig:qual_uncond}
\end{figure*}

In \Cref{fig:qual_uncond}, we provide comparisons between vanilla generation (no guidance), our implementations of SAG,  PAG, SEG, and I-\OURS for unconditional image generation.
While PAG$^{*}$, and SEG$^{*}$ are also able to improve image  quality to satisfying levels, I-\OURS provides an additional level of detail that is not present in the other methods. See for example the table texture on second column, background in the third and fifth column, water texture in fourth column and grass texture in the fifth column.

\begin{figure*}[t]
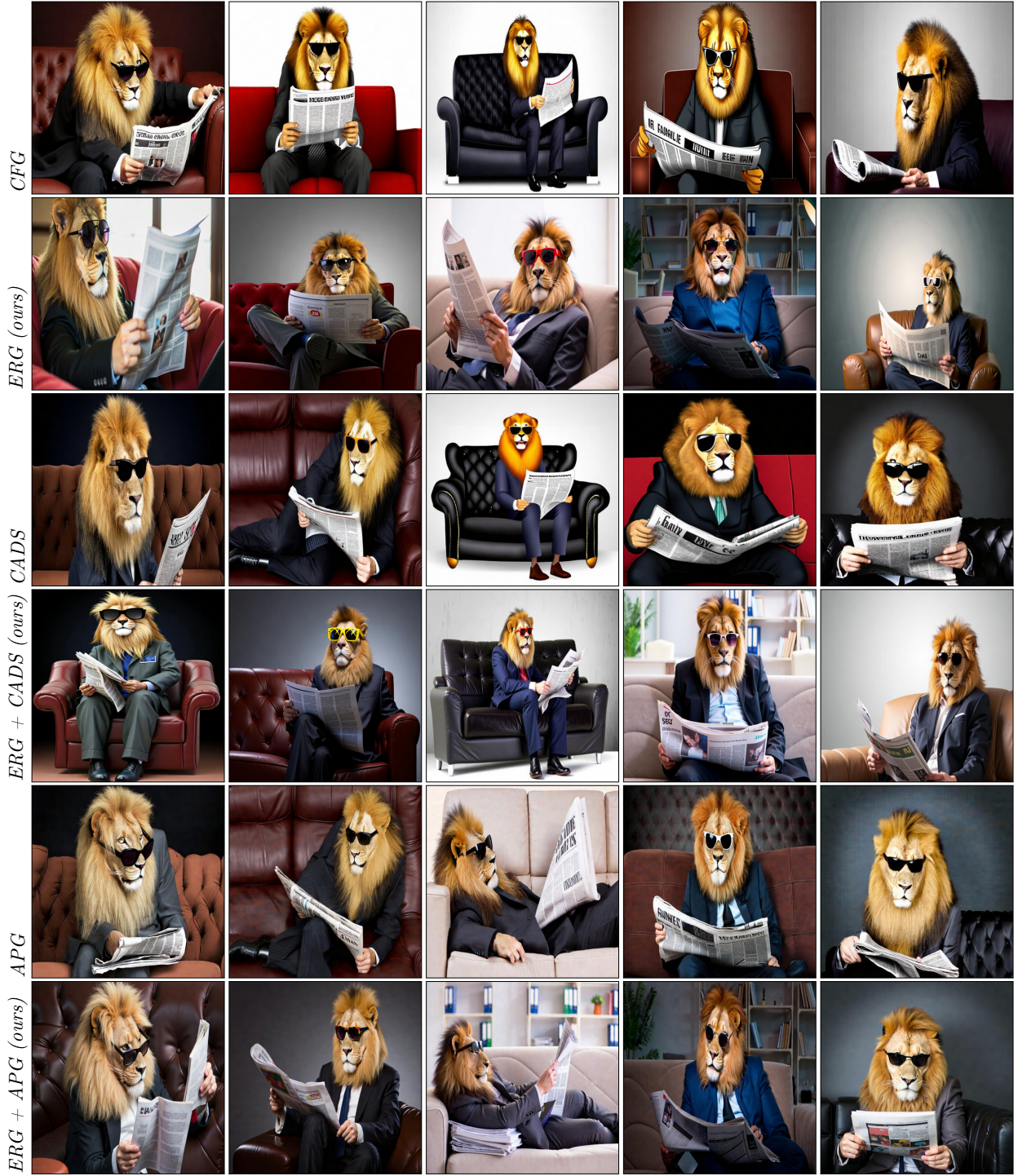

\centering
    \renewcommand{\arraystretch}{0.5}
    \setlength\tabcolsep{1pt}
    \setlength\fboxsep{0pt}
    \def\myim#1#2{\fbox{\includegraphics[width=34mm,height=34mm]{figures/t2i_methods_comp/#1/img_#2.png}}}
        \resizebox{\textwidth}{!}{
    \begin{tabular}{cccccc}
    \begin{sideways} {\it CFG} \end{sideways}   
     & \myim{cfg}{9_1_7.0_original}     
     & \myim{cfg}{2_0_7.0_original}     & \myim{cfg}{3_1_7.0_original}
     & \myim{cfg}{8_2_7.0_original}     & \myim{cfg}{0_3_7.0_original}
     \\
     \begin{sideways} {\it \OURS (ours)} \end{sideways}   
     & \myim{ours}{9_1_7.0_t}     
     & \myim{ours}{2_0_7.0_t}     & \myim{ours}{3_1_7.0_t}
     & \myim{ours}{8_2_7.0_t}     & \myim{ours}{0_3_7.0_t}
     \\
     \begin{sideways} {\it CADS} \end{sideways}   
     & \myim{cads}{9_1_5.0_original}     
     & \myim{cads}{2_0_5.0_original}     & \myim{cads}{3_1_5.0_original}
     & \myim{cads}{8_2_5.0_original}     & \myim{cads}{0_3_5.0_original}
     \\
      \begin{sideways} {\it \OURS + CADS (ours)} \end{sideways}   
     & \myim{ours_cads}{9_1_7.0_t}     
     & \myim{ours_cads}{2_0_7.0_t}     & \myim{ours_cads}{3_1_7.0_t}
     & \myim{ours_cads}{8_2_7.0_t}     & \myim{ours_cads}{0_3_7.0_t}
     \\
     \begin{sideways} {\it APG} \end{sideways}   
     & \myim{apg}{9_1_16.0_original}     
     & \myim{apg}{2_0_16.0_original}     & \myim{apg}{3_1_16.0_original}
     & \myim{apg}{8_2_16.0_original}     & \myim{apg}{0_3_16.0_original}
     \\
     \begin{sideways} {\it \OURS + APG (ours)} \end{sideways}   
     & \myim{ours_apg}{9_1_16.0_t}     
     & \myim{ours_apg}{2_0_16.0_t}     & \myim{ours_apg}{3_1_16.0_t}
     & \myim{ours_apg}{8_2_16.0_t}     & \myim{ours_apg}{0_3_16.0_t}
     \\
    \end{tabular}
    }
    \caption{{\bf Qualitative comparison between different methods.} We provide a detailed qualitative comparison between different methods, we use the ``{\it A lion with sunglasses and a suit, seated in a sofa, reading the newspaper.}'' and sample images with 5 different random seeds, each corresponding to a column in the figure.
    }
    \label{fig:methods_comp}
\end{figure*}

In \Cref{fig:methods_comp}, we provide qualitative comparisons between different methods for text-to-image generation using a single prompt and multiple seeds. 
We compare our method with CFG, CADS and APG. We also provide samples generated when combining our method with either APG or CADS.
In each column the same random seed  is used for all  generations.
We observe that both CFG and CADS tend to generate images that have a cartoonish style, APG on the other hand generates images that are more realistic.
\OURS alone significantly boosts image realism and results in less saturated images (pure white/black backgrounds for example).
Mixing, \OURS with either CADS or APG results in significant improvements in both image quality and diversity.

\begin{figure*}
    \renewcommand{\arraystretch}{0.5}
    \setlength\tabcolsep{1pt}
    \setlength\fboxsep{0pt}
    \def\myim#1{\fbox{\includegraphics[width=0.17\linewidth,height=0.17\linewidth]{figures/showcase_apg/piggybank/img_#1.png}}}
    \def\myimc#1{\fbox{\includegraphics[width=0.17\linewidth,height=0.17\linewidth]{figures/showcase_apg/astronaut/img_#1.png}}}
    \def\myima#1{\fbox{\includegraphics[width=0.17\linewidth,height=0.17\linewidth]{figures/showcase_apg/panda_cook/img_#1.png}}}
    \def\myimb#1{\fbox{\includegraphics[width=0.17\linewidth,height=0.17\linewidth]{figures/showcase_apg/ballerina/img_#1.png}}}
    \footnotesize
        \resizebox{\textwidth}{!}{
    \begin{tabular}{ccccccc}
     & \multicolumn{3}{c}{\shortstack{{\it ``An astronaut on a beach chasing a pig.}\\ {\it The pig is running on the sand while the astronaut }\\ {\it is chasing him. In the background we can see the water.''}}} & \multicolumn{3}{c}{{\it ``hand putting a coin into piggy bank.''}} \\
     \begin{sideways} {\it APG} \end{sideways}   
     & \myimc{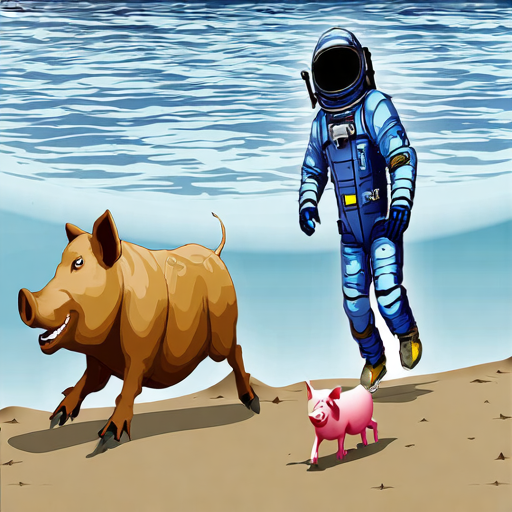}     
     & \myimc{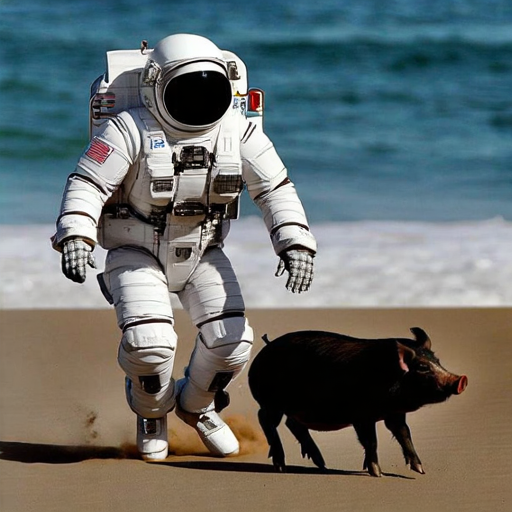}     & \myimc{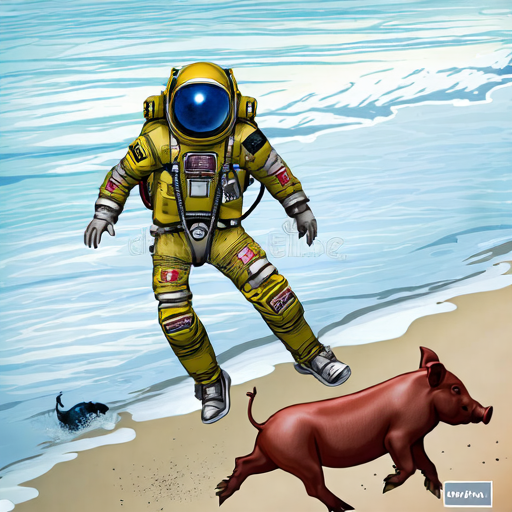}
     & \myim{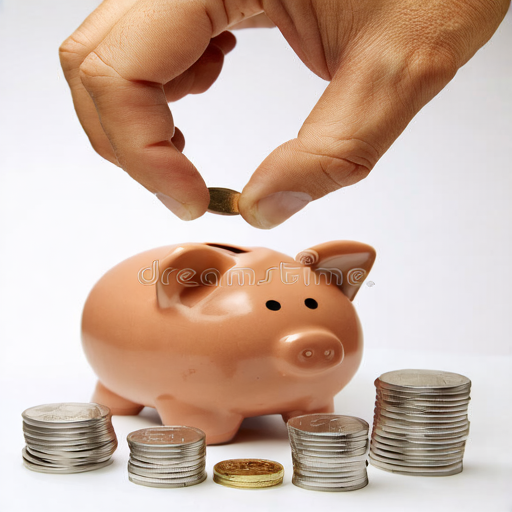}
     & \myim{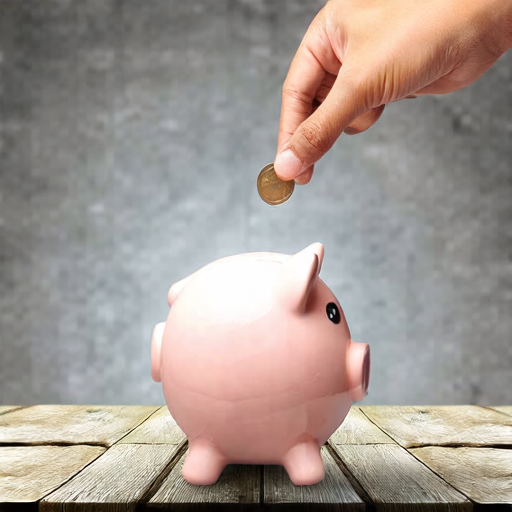}     & \myim{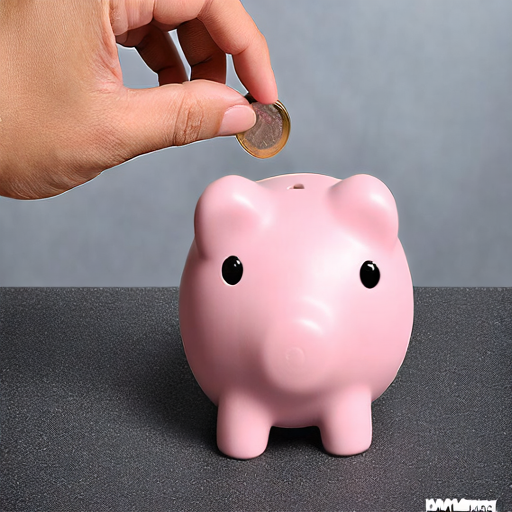}
     \\
    \begin{sideways} {\it \OURS + APG (ours)} \end{sideways}   
     & \myimc{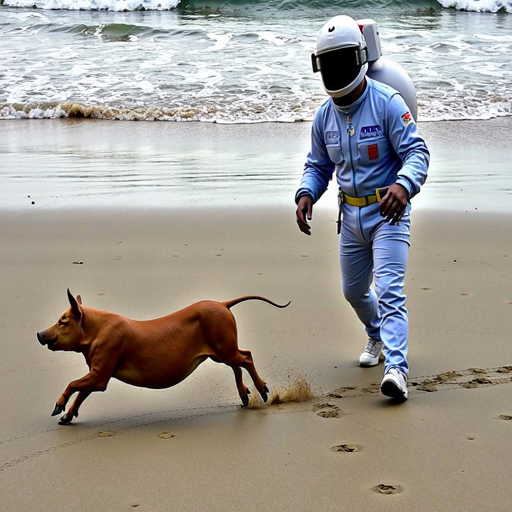}     
     & \myimc{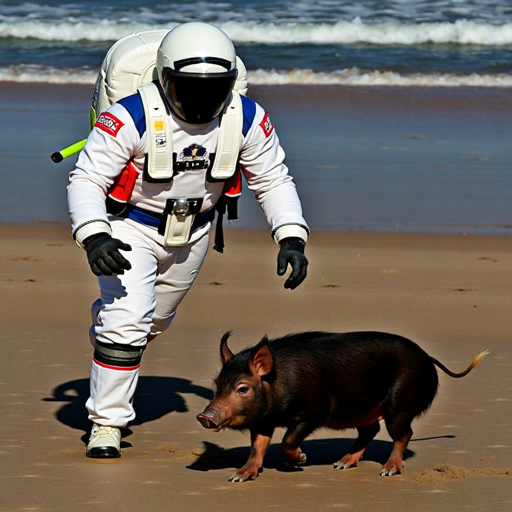}     & \myimc{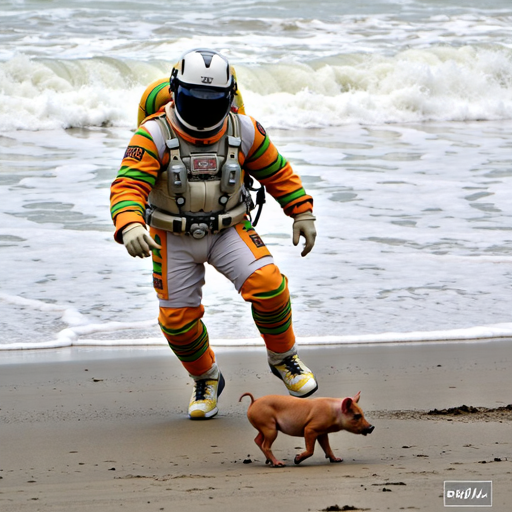}
     & \myim{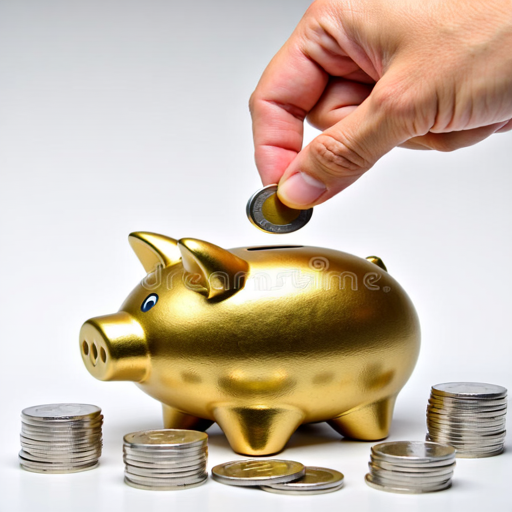}
     & \myim{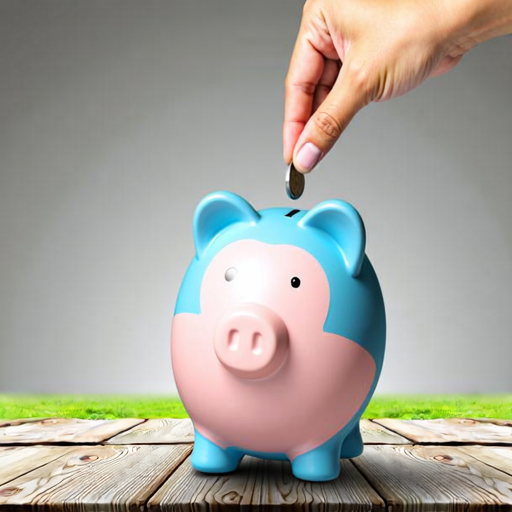}     & \myim{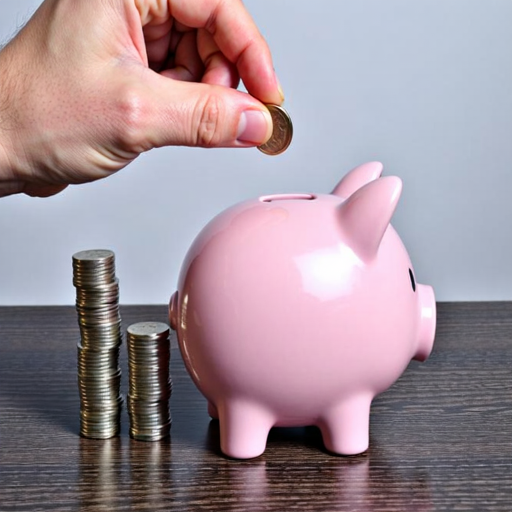}
     \\
     \vspace{2em}\\
     & \multicolumn{3}{c}{{\it ``A ballet dancer next to a waterfall.''}} & \multicolumn{3}{c}{\shortstack{``{ \it A very happy fuzzy panda dressed as a chef}\\ {\it in a high end kitchen making dough. The panda is wearing}\\ {\it an apron,  a toque blanche and baking gloves. }\\ {\it There is a painting of flowers on the wall behind him.''}}}\\
     \begin{sideways} {\it APG} \end{sideways}   
     & \myimb{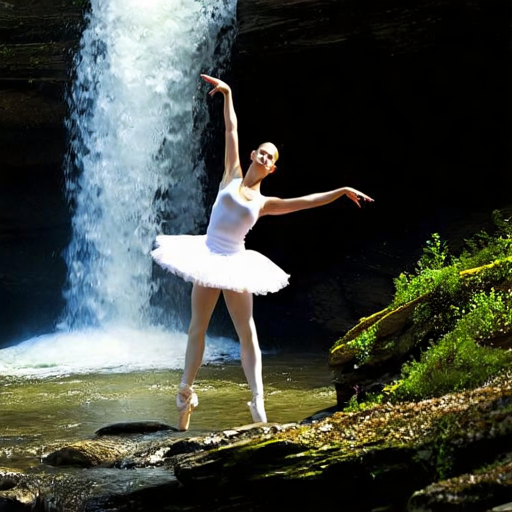}
     & \myimb{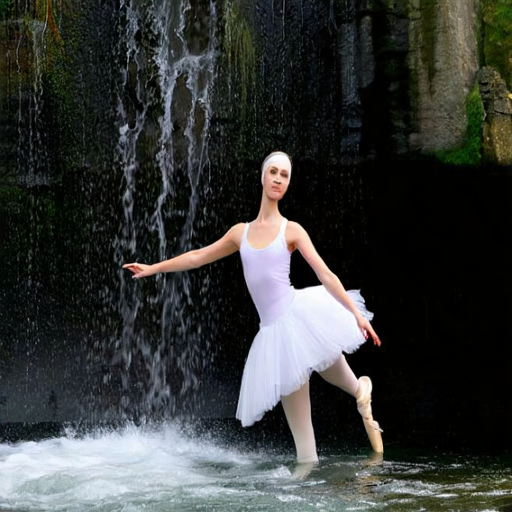}     & \myimb{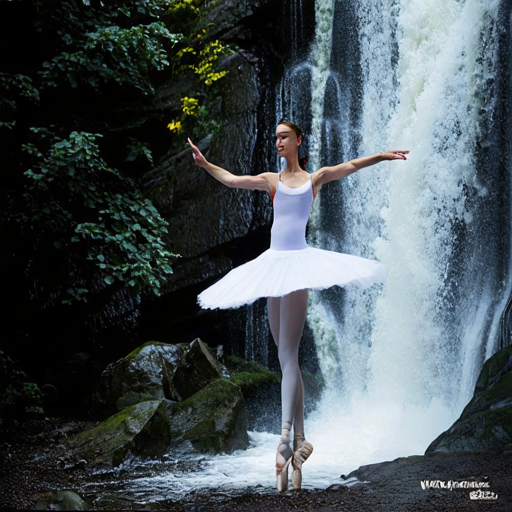} 
     & \myima{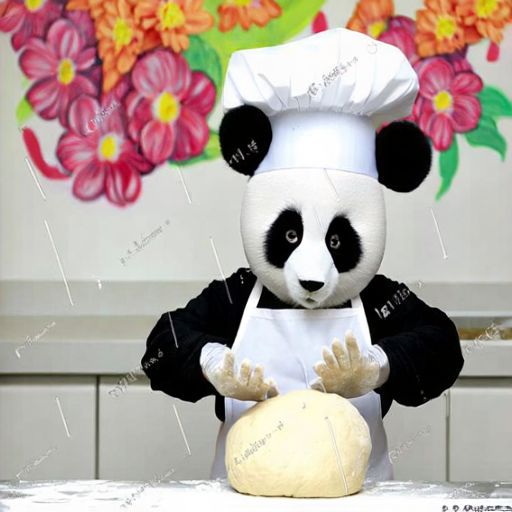}     
     & \myima{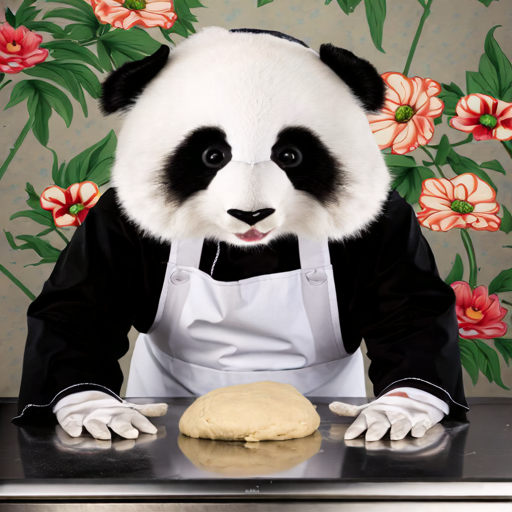}     
     & \myima{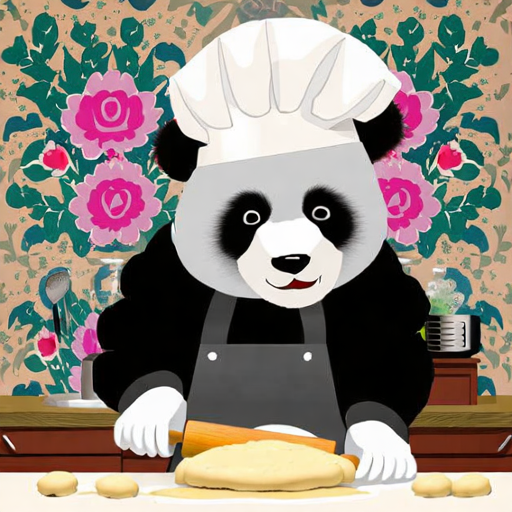}
     \\
    \begin{sideways} {\it \OURS + APG (ours)} \end{sideways}   
     & \myimb{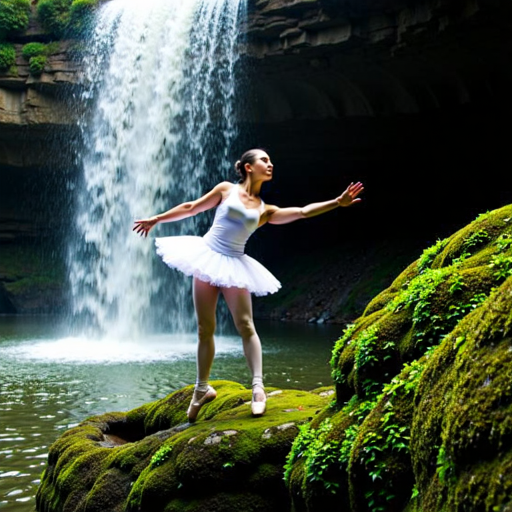}  
     & \myimb{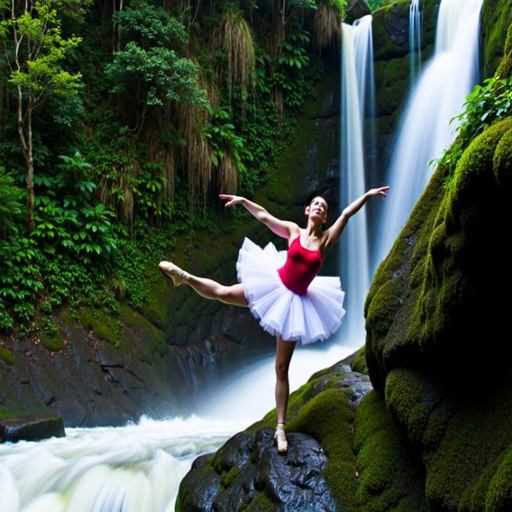}        & \myimb{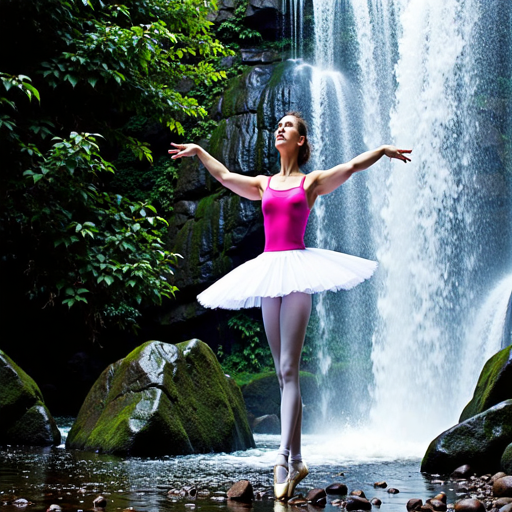} 
     & \myima{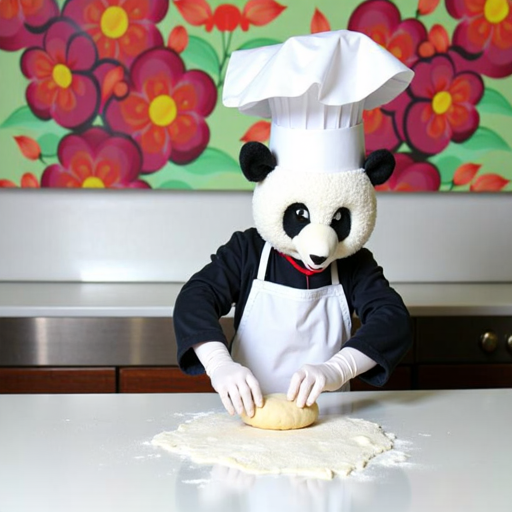}     
     & \myima{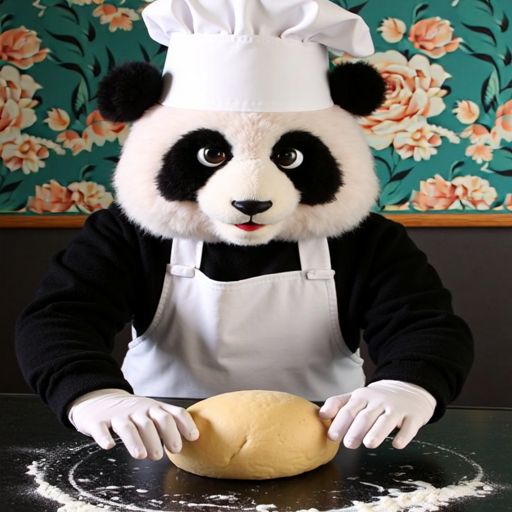}     & \myima{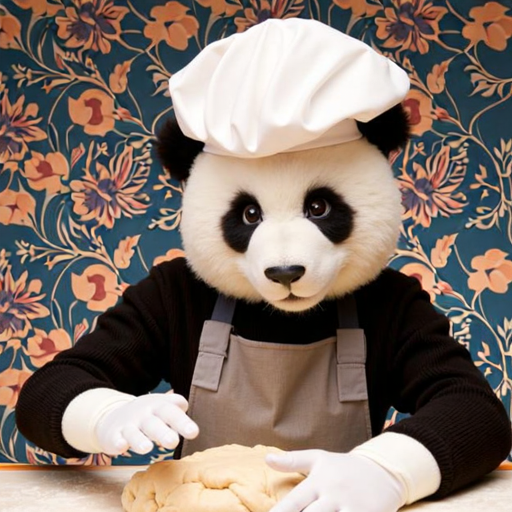}
     \\
    \end{tabular}        
    }
    \caption{{\bf Qualitative comparison showcasing the effect of \ours (\OURS) on generation quality when using APG.}}
    \label{fig:showcase_method_apg}
\end{figure*}

In \Cref{fig:showcase_method_apg}, we provide a qualitative comparison between APG and APG + \OURS, similar to \Cref{fig:showcase_method}, we observe improvements in image quality and diversity.
Also noticeable is a drift from unrealistic image styles (cartoonish) towards more realistic as can be seen in the astronaut and panda examples.

\begin{figure*}[t]
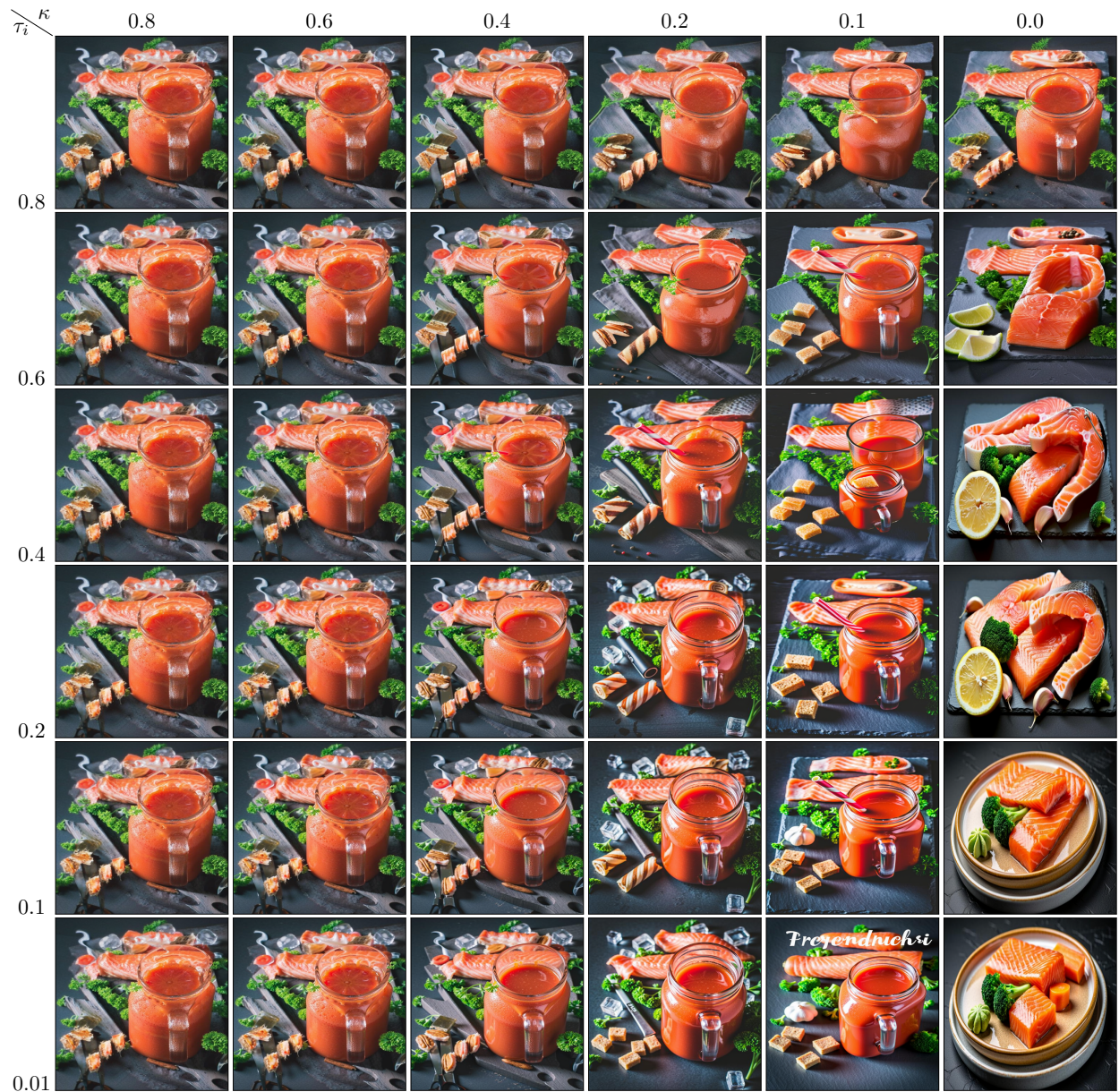

    \renewcommand{\arraystretch}{0.5}
    \setlength\tabcolsep{1pt}
    \setlength\fboxsep{0pt}
    \def\myim#1#2{\fbox{\includegraphics[width=0.17\linewidth,height=0.17\linewidth]{figures/tau_i_tau_interp/img_1_2_#1_#2.png}}}
    \resizebox{\textwidth}{!}{
        \begin{tabular}{cccccccc}
            \diagbox{$\tau_i$}{$\kappa$}& $ 0.8$ & $0.6$ & $0.4$ & $0.2$ & $0.1$ & $0.0$ \\
            $0.8$ & \myim{0.8}{0.8}     
            & \myim{0.8}{0.6} & \myim{0.8}{0.4}
            & \myim{0.8}{0.2} & \myim{0.8}{0.1}
            & \myim{0.8}{0.0}
            \\
            $0.6$ & \myim{0.6}{0.8}     
            & \myim{0.6}{0.6} & \myim{0.6}{0.4}
            & \myim{0.6}{0.2} & \myim{0.6}{0.1}
            & \myim{0.6}{0.0}
            \\
            $0.4$ & \myim{0.4}{0.8}     
            & \myim{0.4}{0.6} & \myim{0.4}{0.4}
            & \myim{0.4}{0.2} & \myim{0.4}{0.1}
            & \myim{0.4}{0.0}
            \\
            $0.2$ & \myim{0.2}{0.8}     
            & \myim{0.2}{0.6} & \myim{0.2}{0.4}
            & \myim{0.2}{0.2} & \myim{0.2}{0.1}
            & \myim{0.2}{0.0}
            \\
            $0.1$ & \myim{0.1}{0.8}     
            & \myim{0.1}{0.6} & \myim{0.1}{0.4}
            & \myim{0.1}{0.2} & \myim{0.1}{0.1}
            & \myim{0.1}{0.0}
            \\
            $0.01$ & \myim{0.01}{0.8}     
            & \myim{0.01}{0.6} & \myim{0.01}{0.4}
            & \myim{0.01}{0.2} & \myim{0.01}{0.1}
            & \myim{0.01}{0.0}
        \end{tabular}
    }
    \caption{{\bf Effect of I-\OURS main parameters.} Interpolating the image kickoff threshold $\kappa$ on the vertical axis and denoiser attention temperature $\tau_i$ on the horizontal axis showcases the effect on image quality and global semantics of the generated image.
    }
    \label{fig:tau_i_interp}
\end{figure*}

In \Cref{fig:tau_i_interp}, we showcase the effect of the image kickoff threshold $\kappa$ and denoiser attention temperature $\tau_i$ by interpolating both parameters in an unconditional generation experiment.
We observe that lowering the attention temperature $\tau_i \rightarrow 0$ results in improved image quality.
Similarly, lowering the image kickoff threshold $\kappa$ further improves image quality but steers the semantics of the image away from the vanilla generation ($\kappa=1.0, \tau_i=1$).
However, as the image level threshold gets smaller $\kappa \rightarrow 0$, the semantics of the original image diverge from the vanilla generations, resulting in images with largely different semantics.
We find $\kappa \in [0.2, 0.4]$ and $\tau_i = 0.01$ to work well in most cases.

\begin{figure*}[t]
    \renewcommand{\arraystretch}{0.5}
    \setlength\tabcolsep{0pt}
    \setlength\fboxsep{0pt}
    \def\myim#1{\fbox{\includegraphics[width=41mm,height=41mm]{figures/lmin_lmax/img_0_2_#1.png}}}
        \resizebox{\textwidth}{!}{
        \begin{tabular}{cccc}
           none & $3-6$ & $6-9$ & $9-12$\\
            \myim{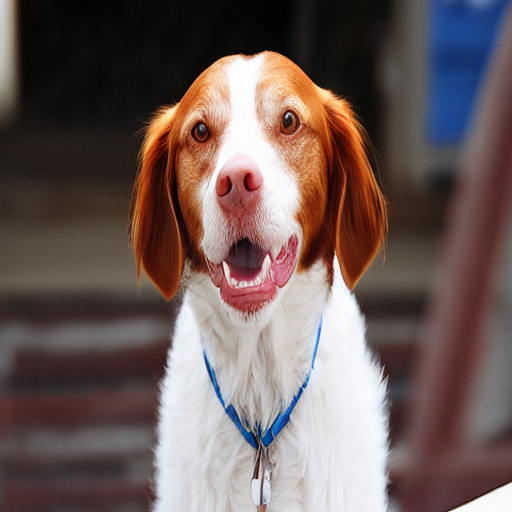}     
            & \myim{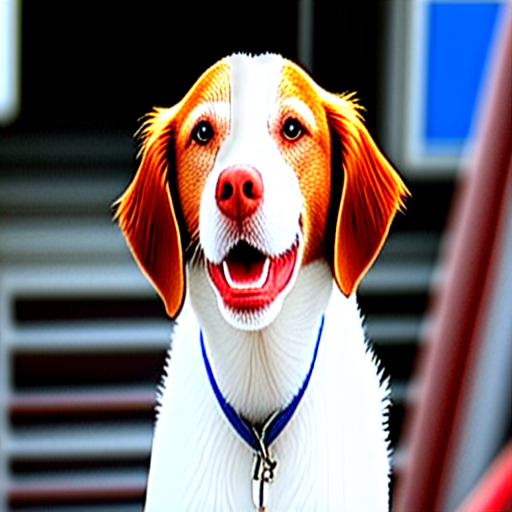} & \myim{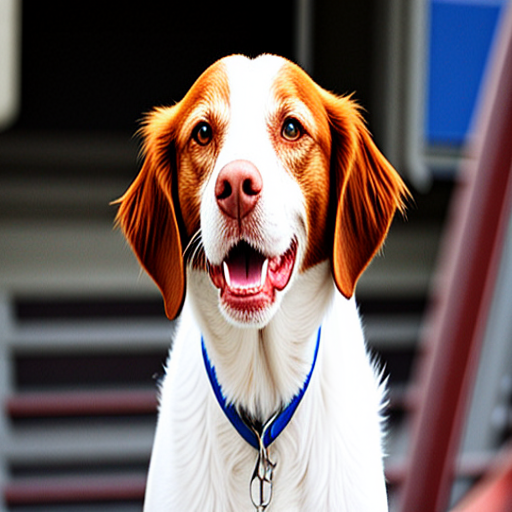}
            & \myim{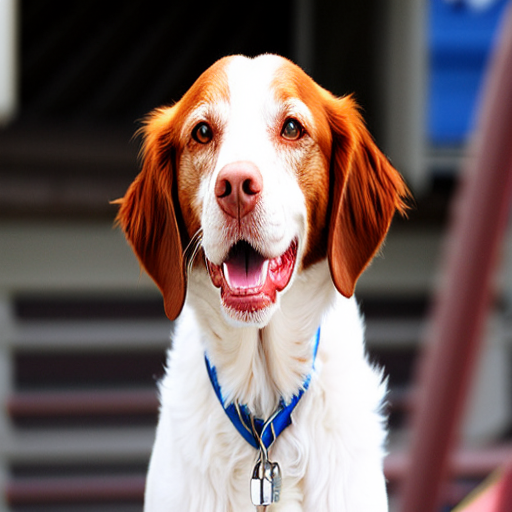}\\
            $12-15$ & $15-18$ & $18-21$ & $21-24$\\
            \myim{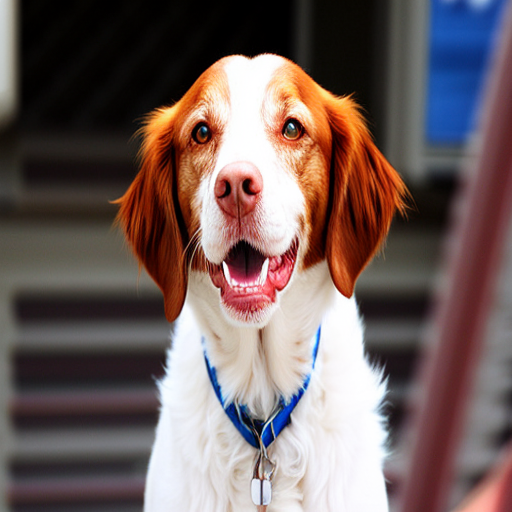}
            & \myim{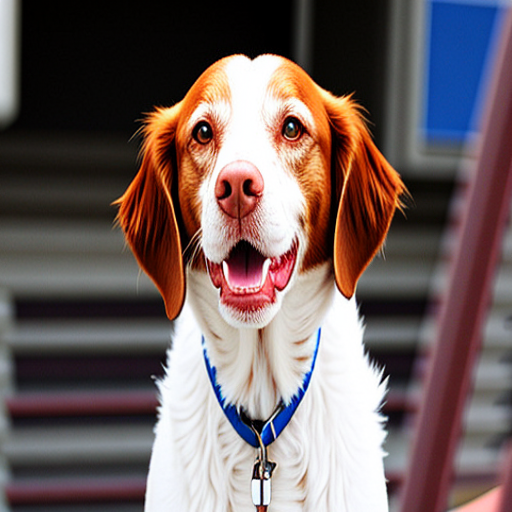} & \myim{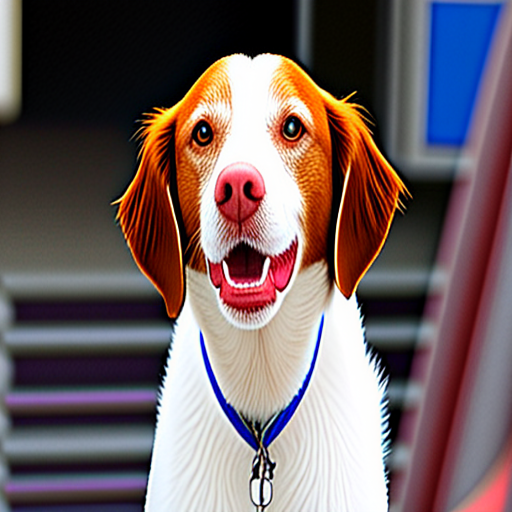} 
            & \myim{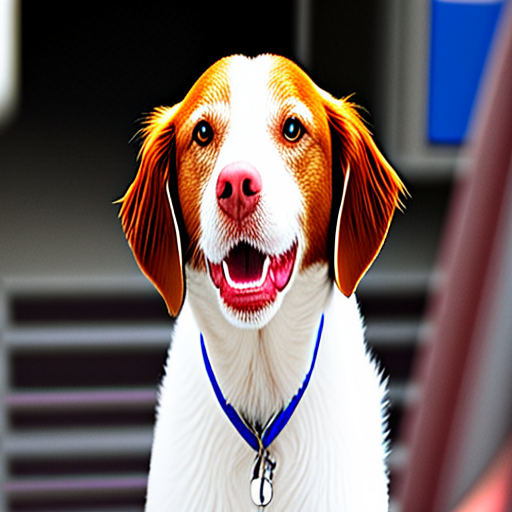}
            \\
        \end{tabular}
        }
    \caption{{\bf Effect of I-\OURS main parameters.} We show the effect of applying I-\OURS on different layers of the transformer $l_\text{min}^i, l_\text{max}^i$.
    Experiment conducted on DiT-XL/2 model, using a guidance scale of $5.0$.
    }
    \label{fig:lmin_lmax}
\end{figure*}

\begin{figure*} 
    \renewcommand{\arraystretch}{0}
    \setlength\tabcolsep{0.5pt}
    \setlength\fboxsep{0pt}
    \def\myim#1{\fbox{\includegraphics[width=40mm,height=40mm]{figures/room_mechanisms/img_#1.png}}}
    \footnotesize
    \resizebox{\textwidth}{!}{
    \begin{tabular}{cccc}
    CFG & Energy Smoothing & Identity Mapping & Temperature Reduction (ours)\\
     \myim{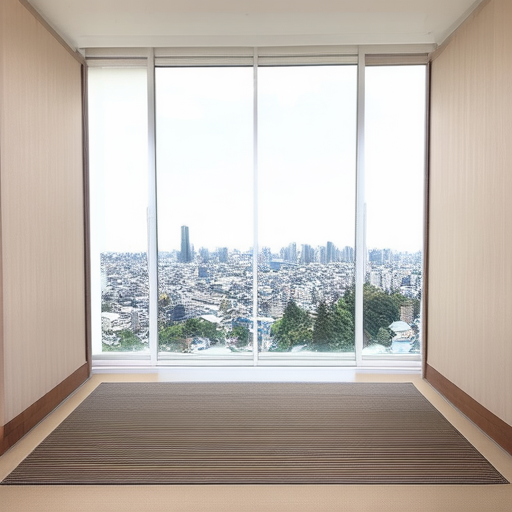}     & \myim{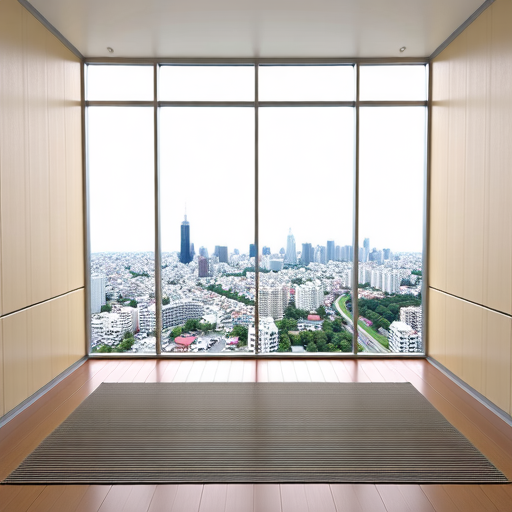}     
     & \myim{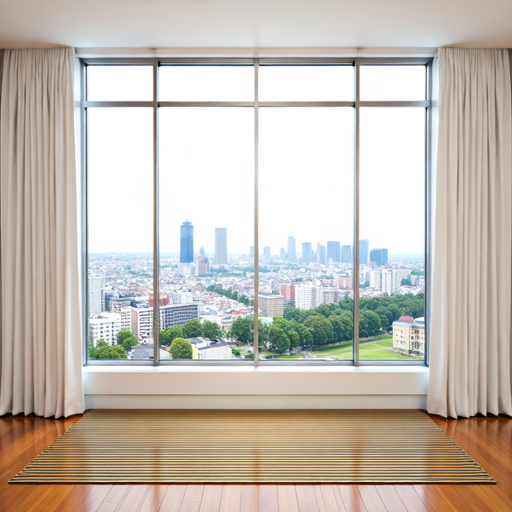} & \myim{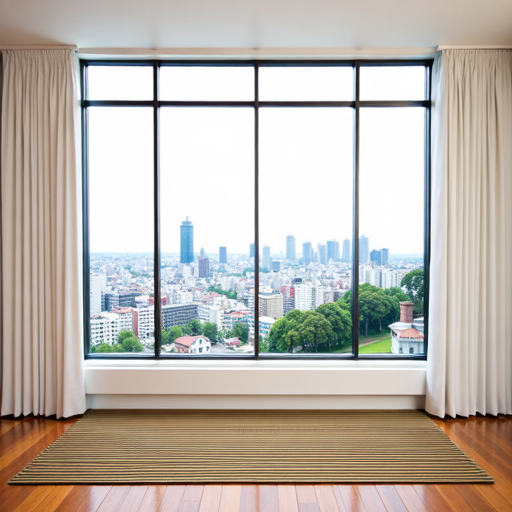}
     \\
     \multicolumn{4}{c}{\shortstack{{\it A spacious, serene room influenced by modern Japanese aesthetics}\\ {\it with a view of a cityscape outside of the window.}}}\\
     \\\\
     \myim{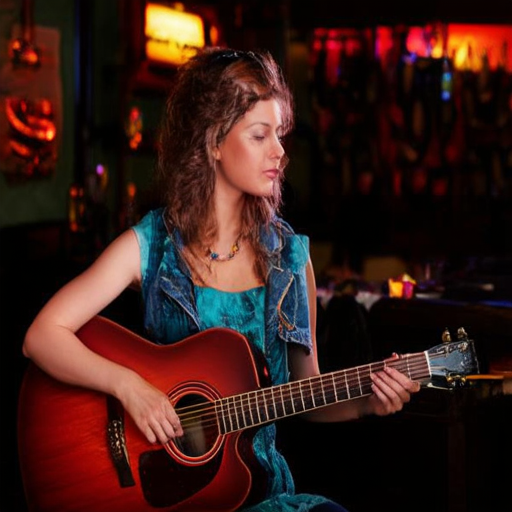}     & \myim{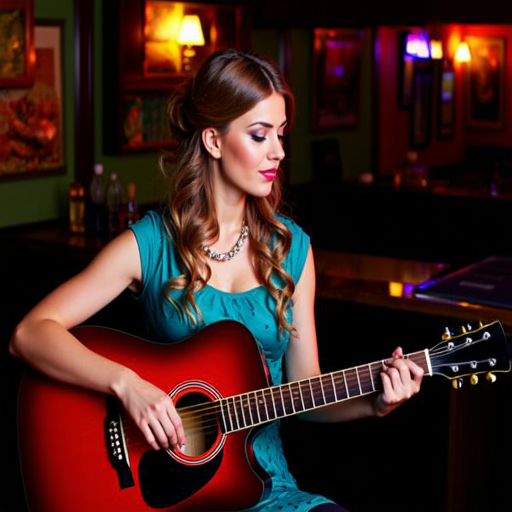}     
     & \myim{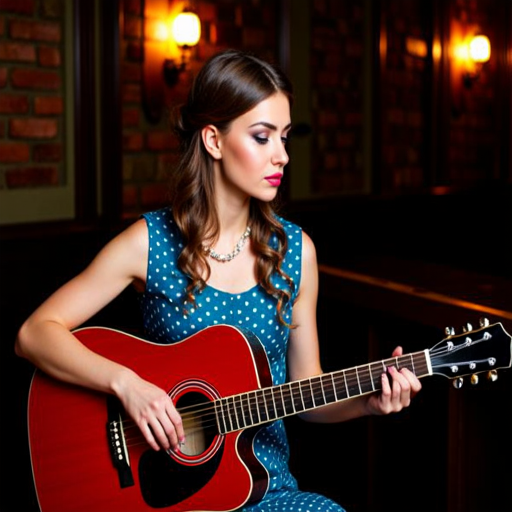} & \myim{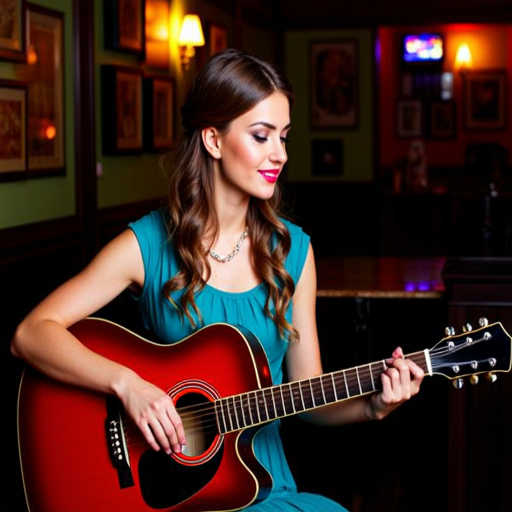}
     \\
     \multicolumn{4}{c}{{\it A woman playing guitar in a jazz bar.}}\\
    \end{tabular}        
}
    \caption{{\bf Comparing attention rectification mechanisms.} Compared to attention smoothing and identity mapping, temperature rescaling better models high-level details in the image such as textures of the tiling and small buildings in the distance. Temperature rescaling also showcases better structural coherence when compared with different methods.
    }
    \label{fig:qual_rectif_mech}
\end{figure*}

In \Cref{fig:lmin_lmax}, we illustrate the effect of applying I-\OURS on different layers of the denoiser.
This experiment is conducted for class-conditional ImageNet-1k generations with our DiT-XL/2 model with 28 layers.
We apply I-\OURS to a different range of layers where three layers are involved each time.
Applying I-\OURS to either very early or late layers results in similar effects, with noticeable over-saturation and unnatural patterns that are overly simplistic as can be seen in the dog's fur.
On the other hand, applying I-\OURS to middle layers results in improved image quality and sharper details.
We find $l_\text{min}=12,l_\text{max}=15$ to work well and use it for the rest of the experiments  in this setup.

\begin{figure*}[t]
\centering
    \renewcommand{\arraystretch}{0.5}
    \setlength\tabcolsep{1pt}
    \setlength\fboxsep{0pt}
    \def\myim#1{\fbox{\includegraphics[width=34mm,height=34mm]{figures/in_512_xl2/img_#1.png}}}
    \resizebox{\textwidth}{!}{
    \begin{tabular}{cccccc}
     \begin{sideways} {\it CFG} \end{sideways}   
     & \myim{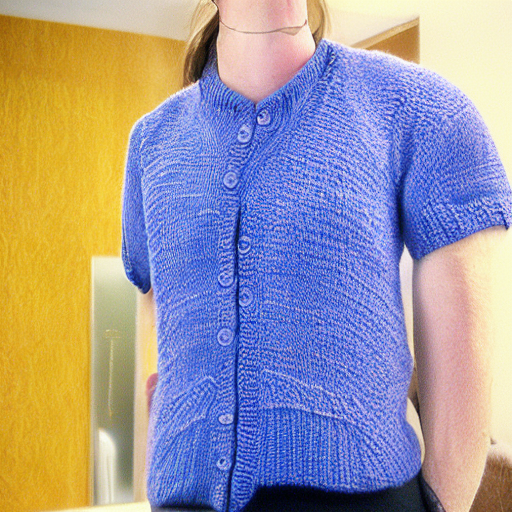}     
     & \myim{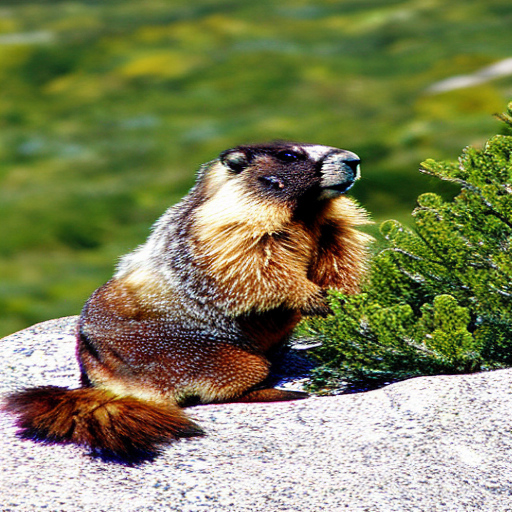}     & \myim{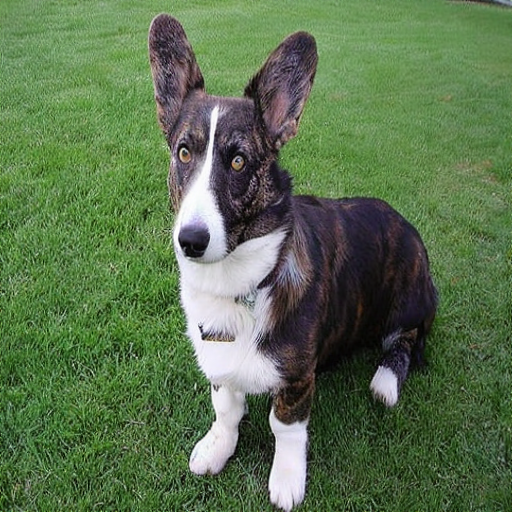}
     & \myim{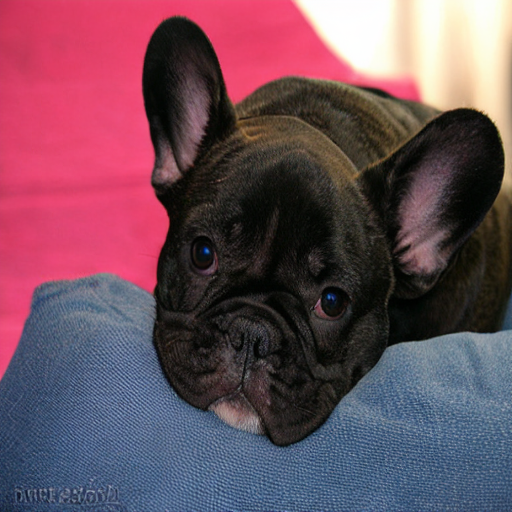}     & \myim{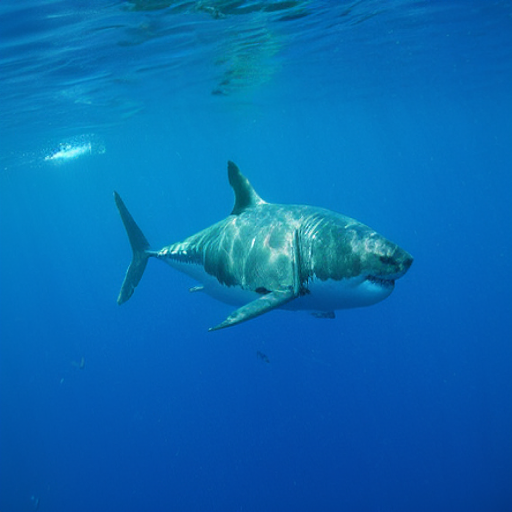}
     \\
     \begin{sideways} {\it SEG$^*$} \end{sideways}   
     & \myim{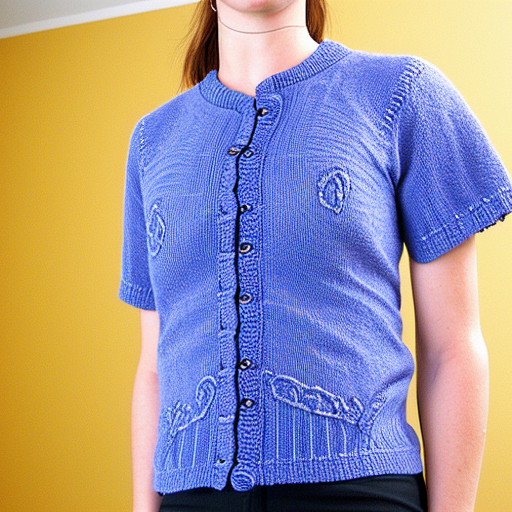}     
     & \myim{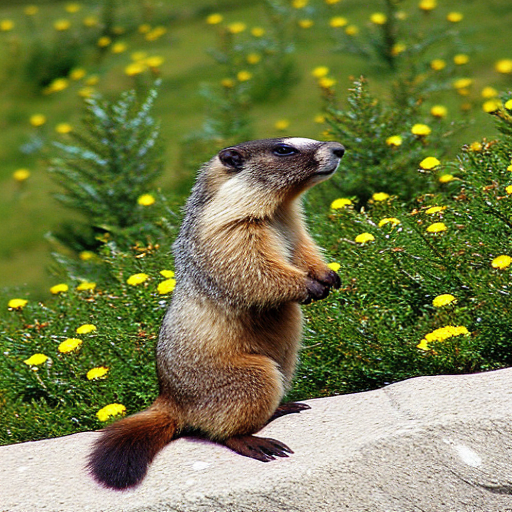}     & \myim{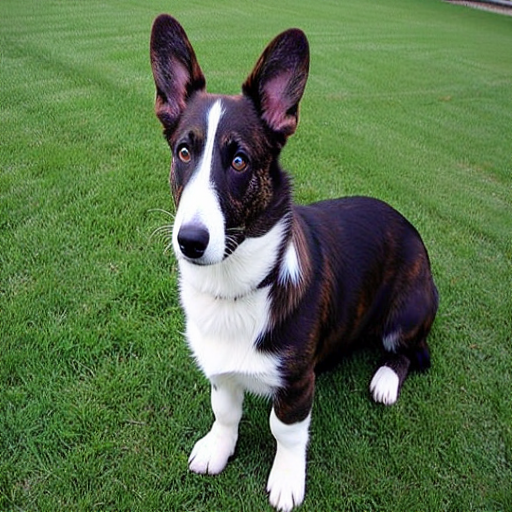}
     & \myim{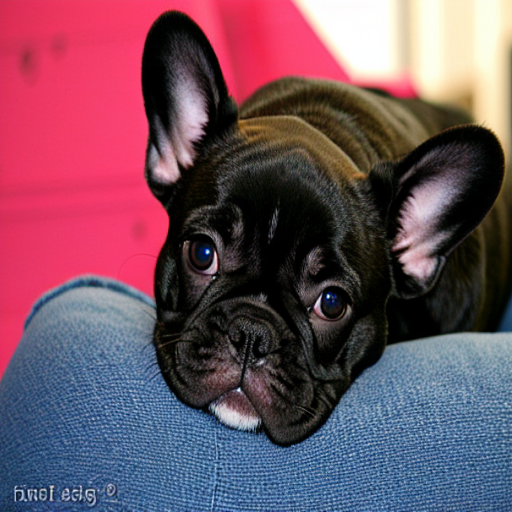}     & \myim{20_0_0.0_9999.0_False_False}
     \\
     \begin{sideways} {\it PAG$^*$} \end{sideways}   
     & \myim{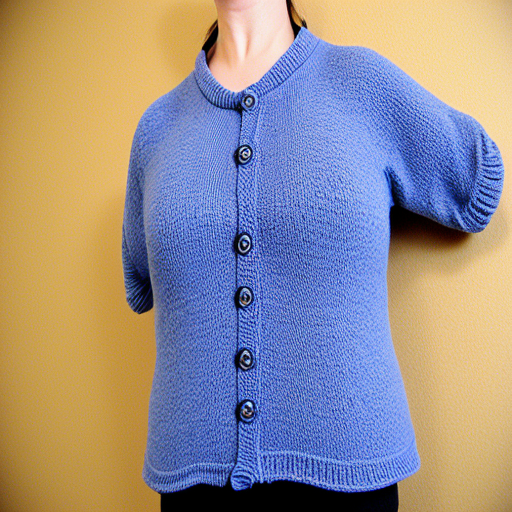}     
     & \myim{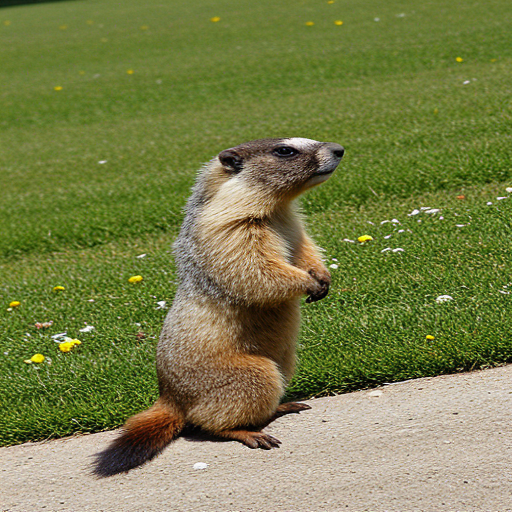}     & \myim{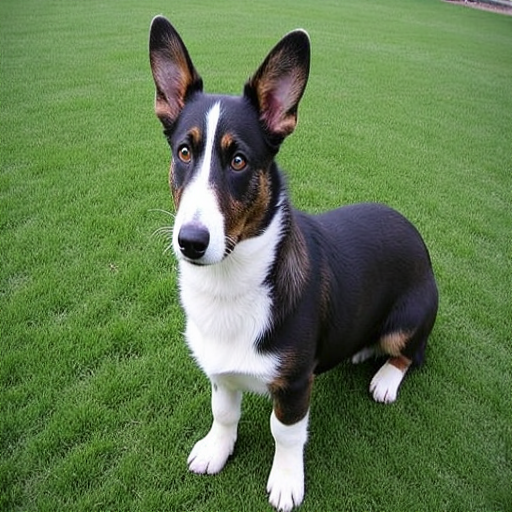}
     & \myim{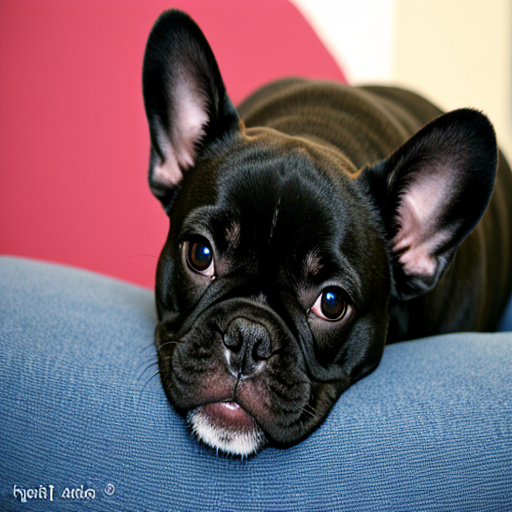}     & \myim{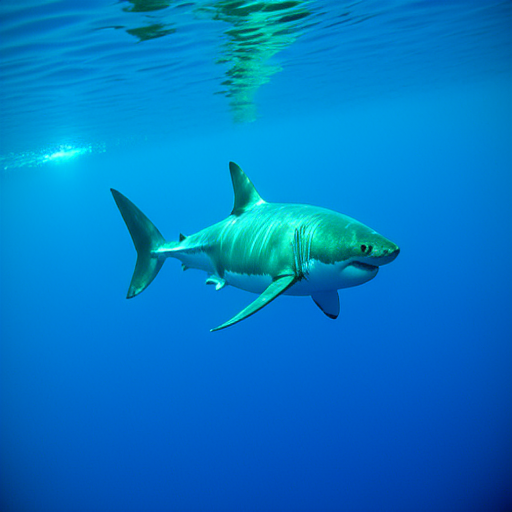}
     \\
    \begin{sideways} {\it SAG$^*$} \end{sideways}   
     & \myim{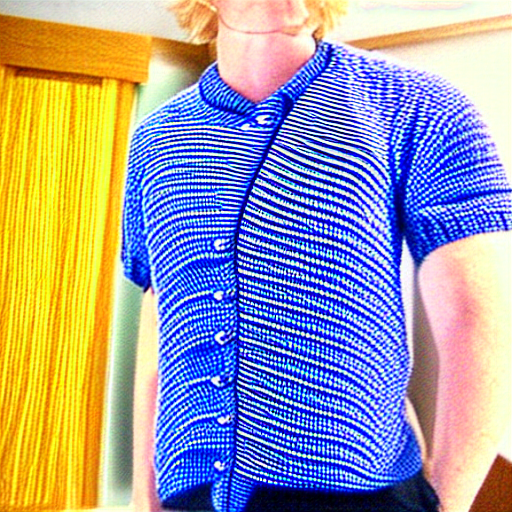}     
     & \myim{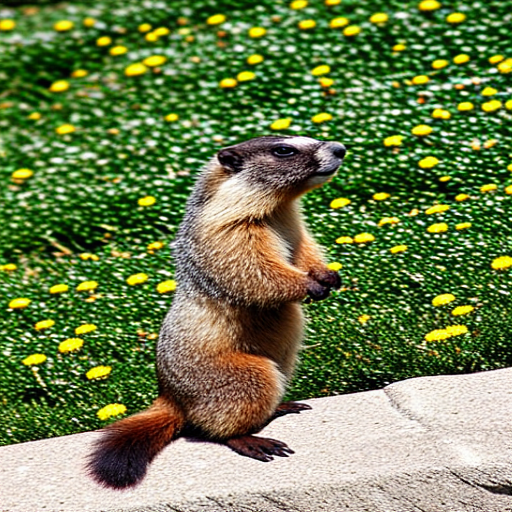}     & \myim{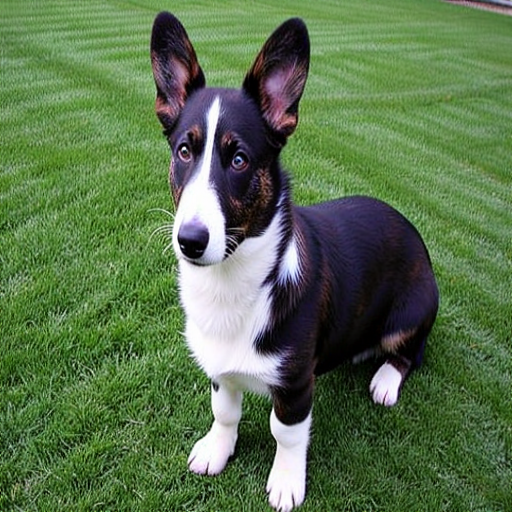}
     & \myim{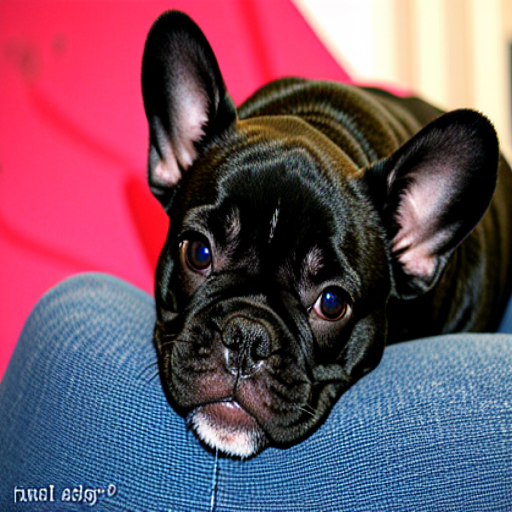}     & \myim{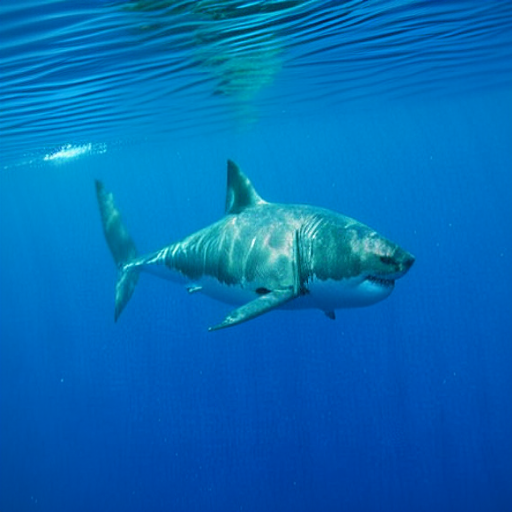}
     \\
     \begin{sideways} {\it I-\OURS} \end{sideways}   
     & \myim{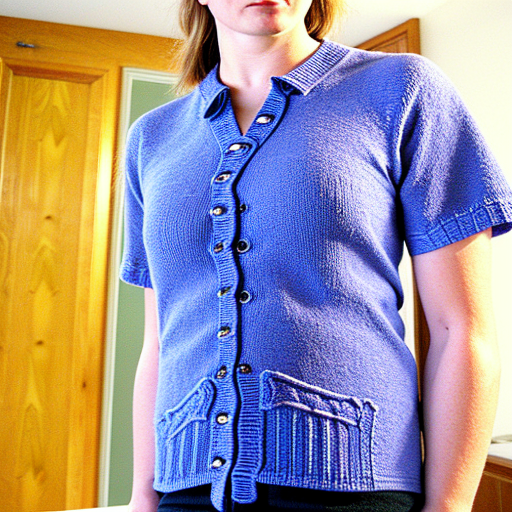}     
     & \myim{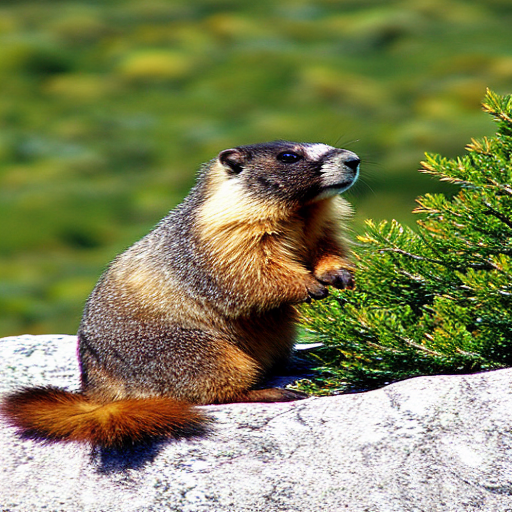}     & \myim{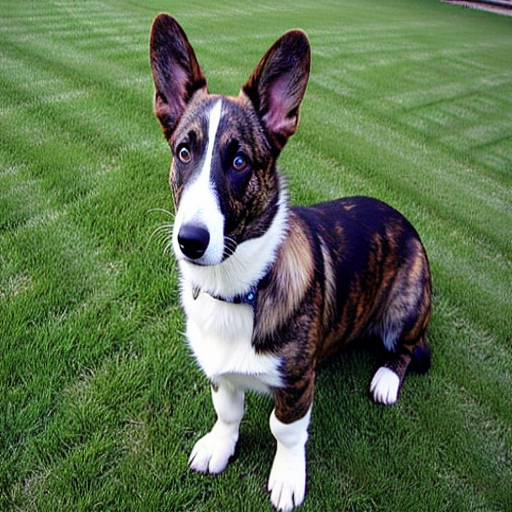}
     & \myim{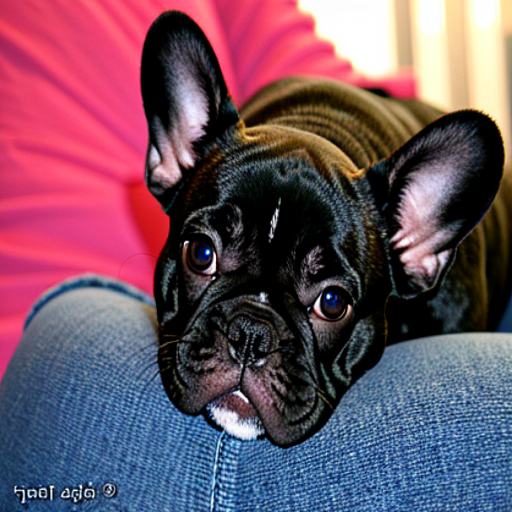}     & \myim{20_0_0.2_0.01_False_False}
     \\
    \end{tabular}
    }
    \caption{{\bf Qualitative comparison on ImageNet-1k@512.} 
    Samples from the same column are generated using the same random seed, we compare \OURS with CFG and SEG. Our method produces images of better quality.
    Images generated using Euler sampler with 50 steps.
    }
    \label{fig:qualitative_comp_in1k_512_xl2}
\end{figure*}

In \Cref{fig:qualitative_comp_in1k_512_xl2}, we provide qualitative results comparing I-\OURS with SEG and CFG for class-conditional generation on ImageNet at $512$ resolution, using a DiT-XL/2 model.
Similarly to the unconditional and text-to-image cases, we find I-\OURS to achieve higher image quality when compared CFG and our reimplementation of other guidance methods.

\newcommand{\promptside}[1]{%
  \multirow{2}{*}[2.5cm]{%
    \rotatebox{90}{%
      \parbox{5cm}{\centering\itshape #1}%
    }%
  }%
}

\begin{figure*}[t]
    \renewcommand{\arraystretch}{0.5}
    \setlength\tabcolsep{10pt}
    \setlength\fboxsep{0pt}
    \def\myim#1{\fbox{\includegraphics[width=50mm,height=50mm]{figures/sd3/img_#1.png}}}
    \def\myiml#1{\fbox{\includegraphics[width=50mm,height=50mm]{figures/sd3p5/img_#1.png}}}

    \resizebox{\textwidth}{!}{
        \begin{tabular}{cccc}
            \promptside{An elephant crossing a river.} & \myim{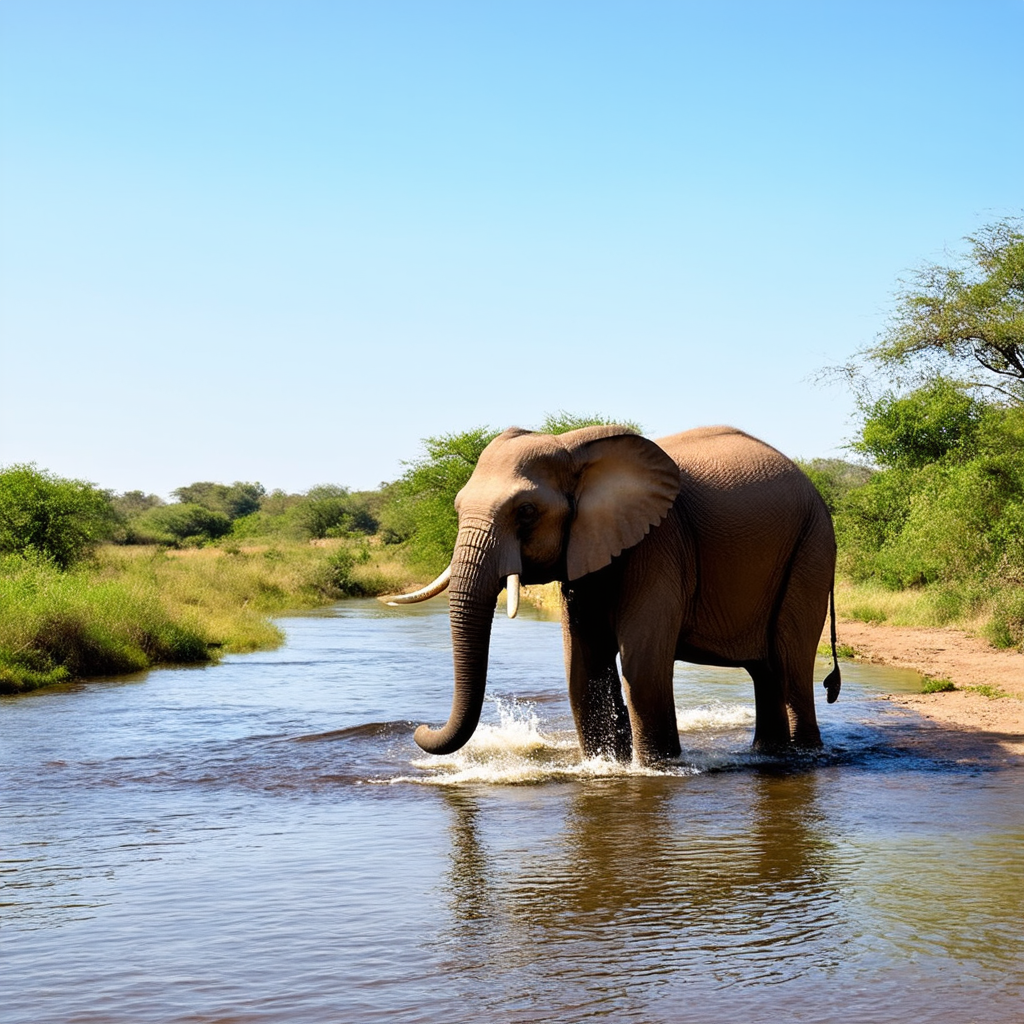} &  \promptside{\shortstack{{\it A child wearing an apron and sunglasses}\\ {\it posing in front of a blackboard with heart}\\ {\it drawings and 'ERG' written in it.}}} & \myim{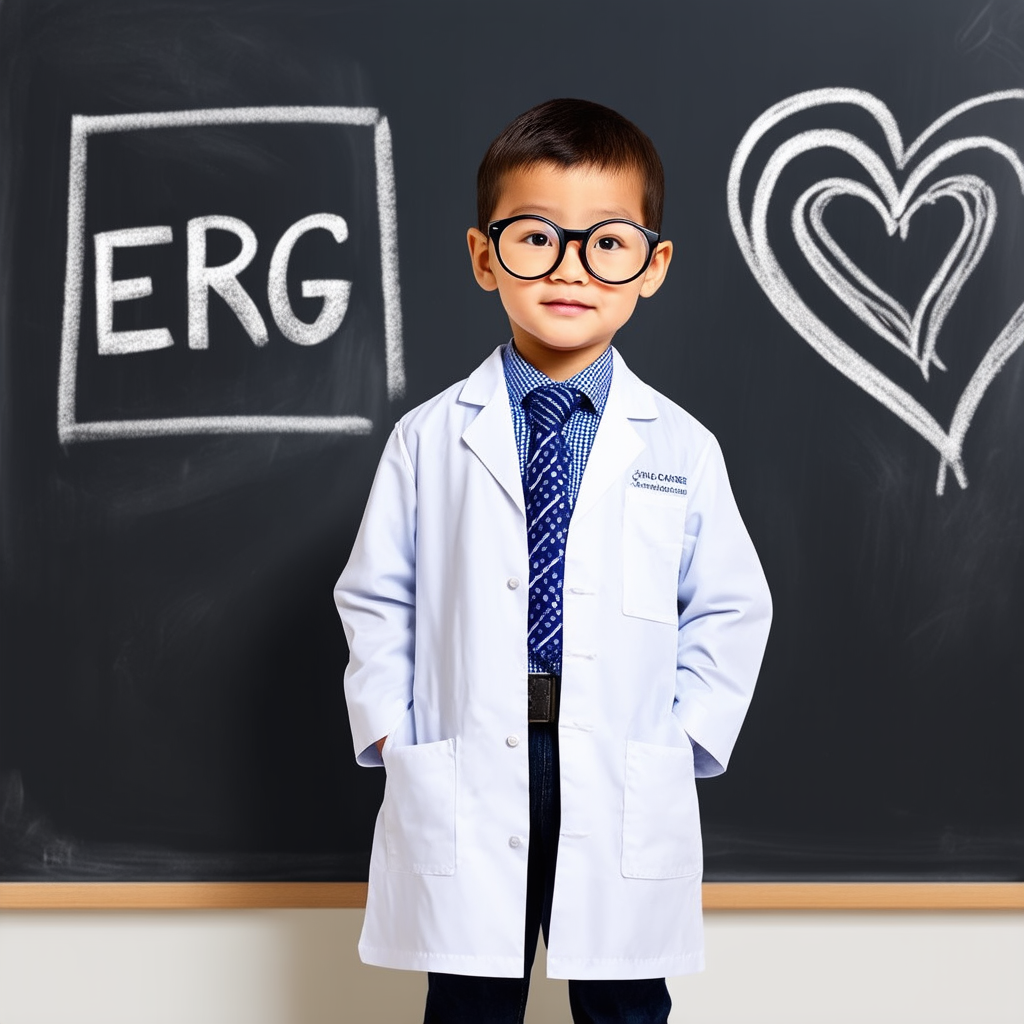}\\
             & \myim{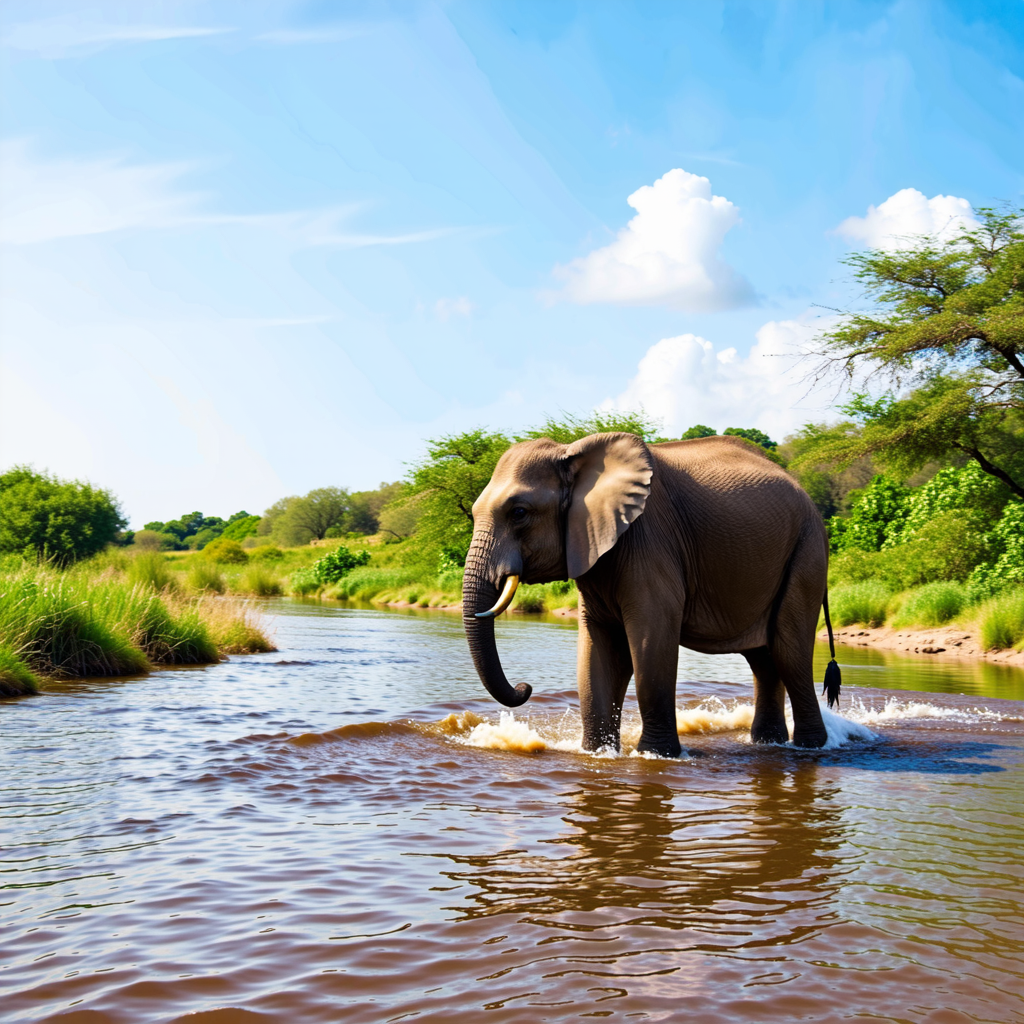} & &
            \myim{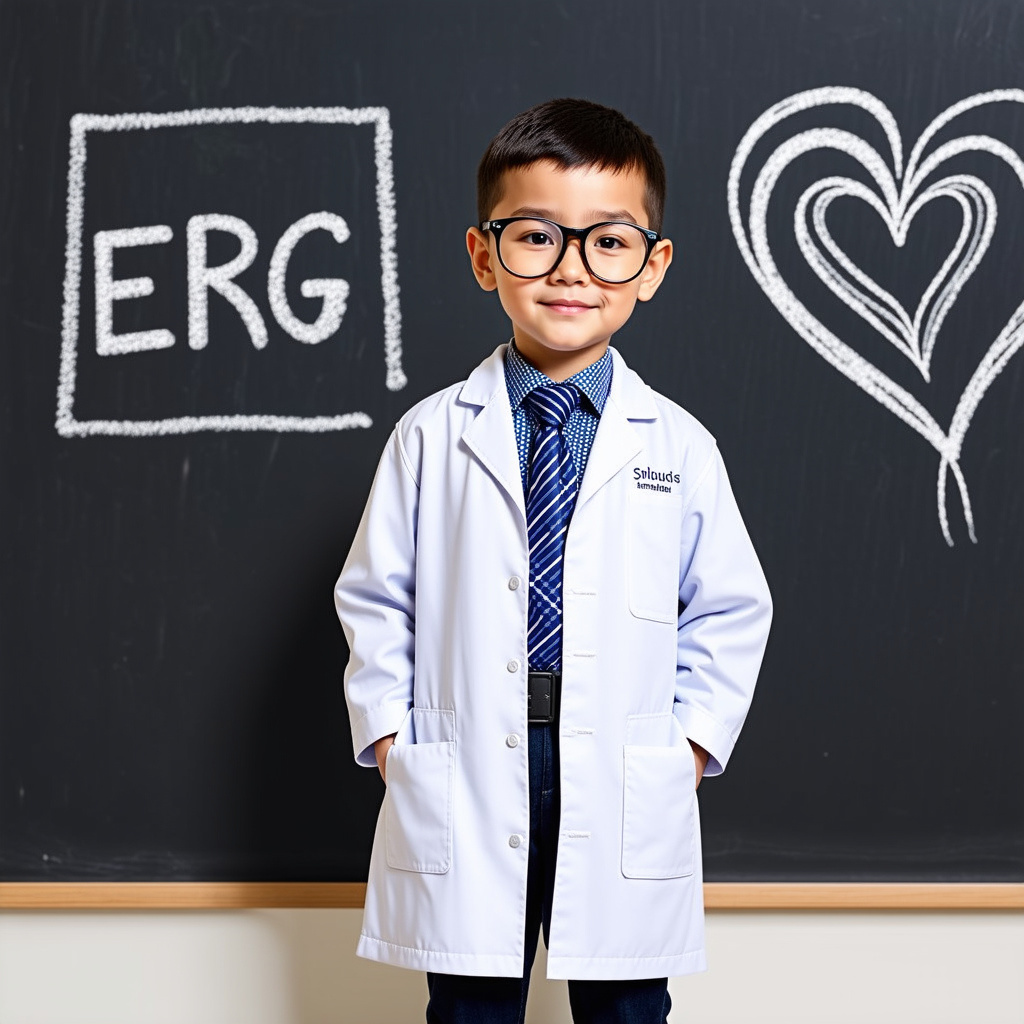}\\
            \promptside{\shortstack{{\it A group of supporters wearing football jerseys}\\ {\it cheering for their team after a victory.}}} & \myiml{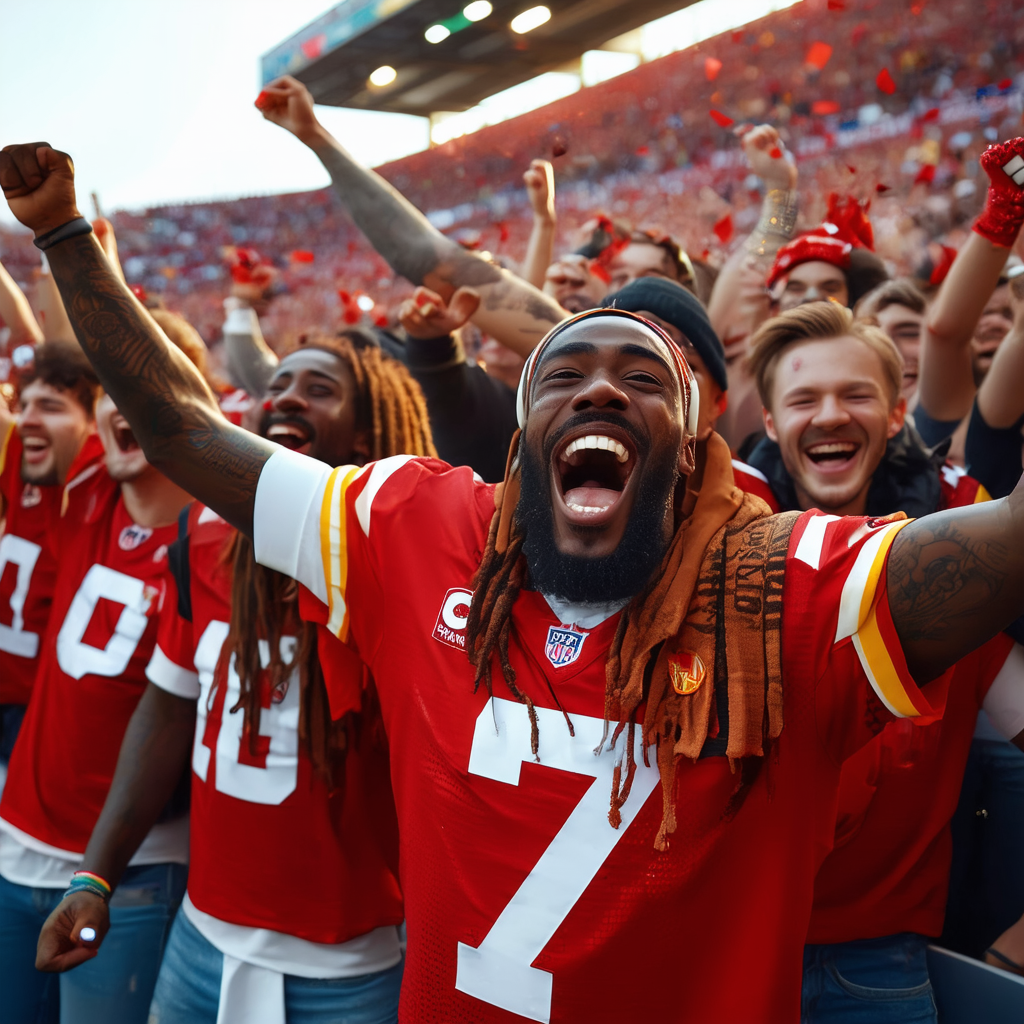} & \promptside{A maltese dog getting a haircut by a hairdresser.}  & \myiml{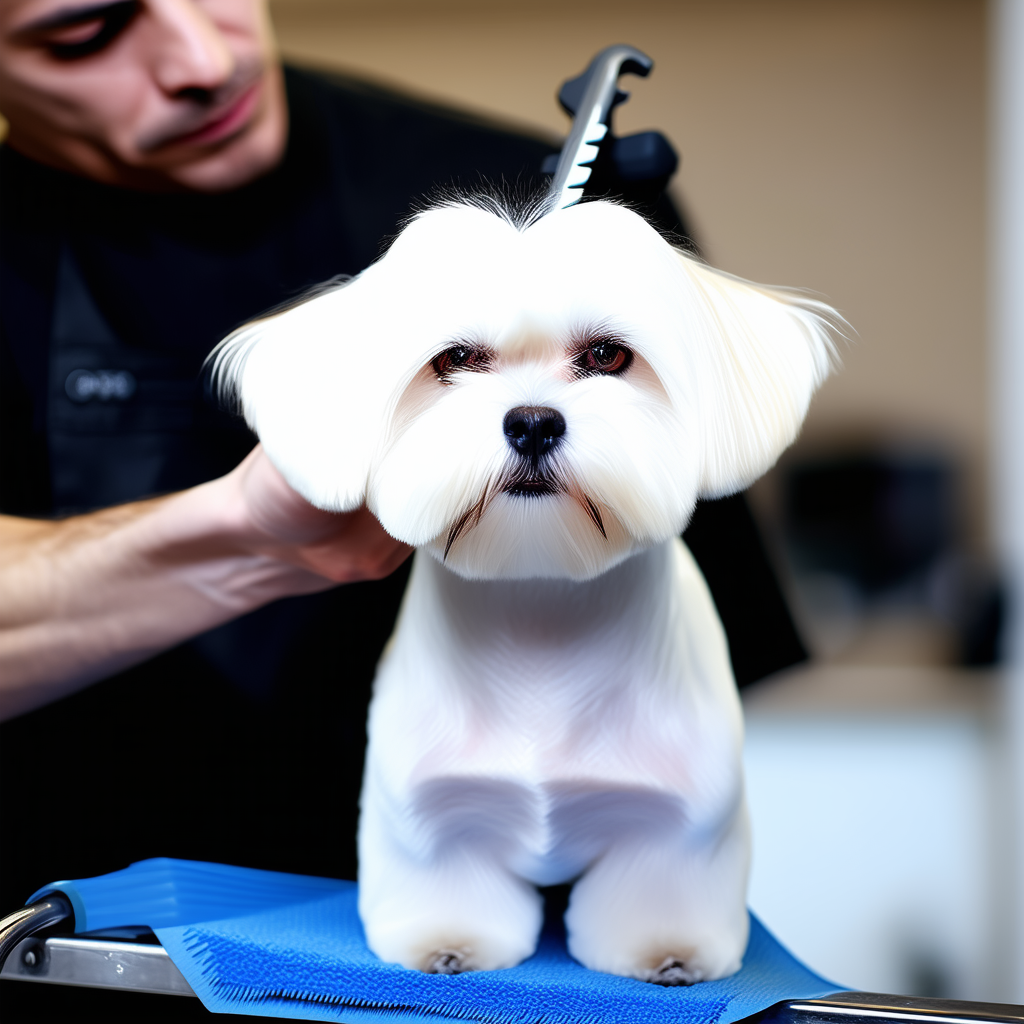}\\
             & \myiml{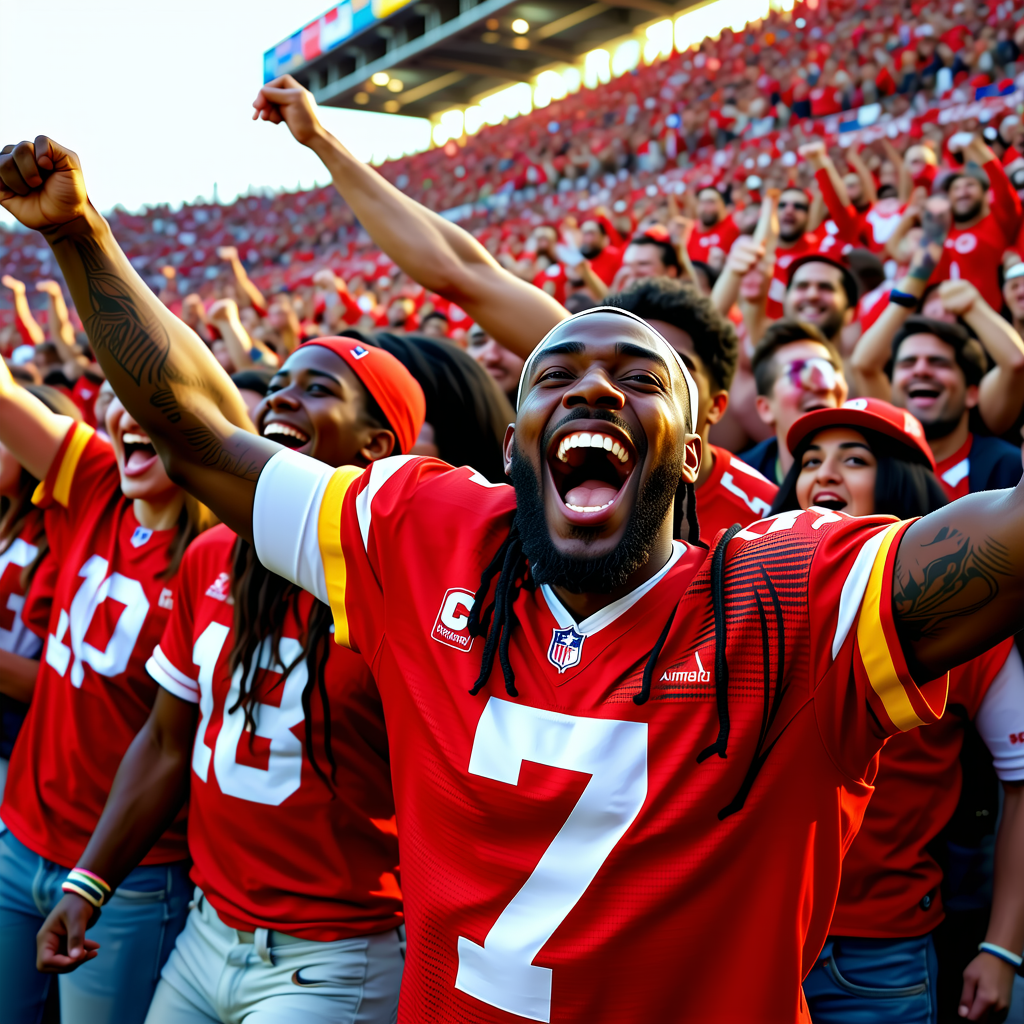}&  &
            \myiml{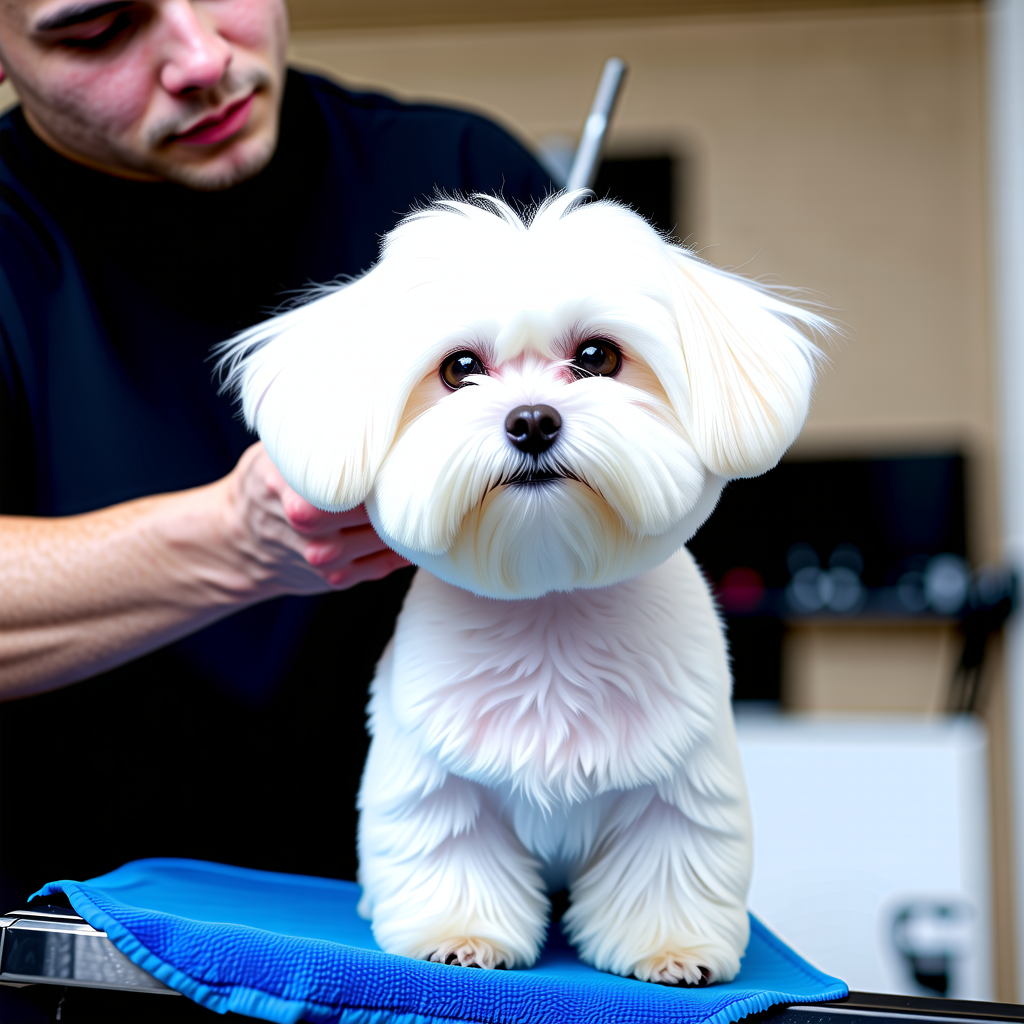}\\
            
        \end{tabular}
    }
    \caption{{\bf Qualitative comparison using  Stable Diffusion 3.} 
    For each prompt we use the same seed to sample images from \texttt{sd3-medium} (top rows) and \texttt{sd3.5-large} (bottom rows) with  standard CFG (top) and   I-\OURS (bottom).
    We observe better image details using I-\OURS.
    }
    \label{fig:sd3_example}
\end{figure*}

\end{document}